\documentclass[10pt,twocolumn,letterpaper]{article}

\usepackage{iccv}
\usepackage{times}
\usepackage{epsfig}
\usepackage{graphicx}
\usepackage{amsmath}
\usepackage{amssymb}

\usepackage{float}
\usepackage{amsmath}
\usepackage{comment}
\usepackage{array}
\usepackage{multirow}
\usepackage{hhline}
\usepackage{threeparttable}
\usepackage{dsfont}
\usepackage [english]{babel}
\usepackage[switch]{lineno}
\usepackage [autostyle, english = american]{csquotes}

\MakeOuterQuote{"}
\newtheorem{definition}{Definition}
\usepackage[pagebackref=true,breaklinks=true,letterpaper=true,colorlinks,bookmarks=false]{hyperref}

\iccvfinalcopy 

\def\iccvPaperID{7255} 

\newcommand{\revision}[1]{{#1}}

\begin{document}

\title{Understanding and Mitigating Annotation Bias \\  in Facial Expression Recognition}

\author{Yunliang Chen \qquad Jungseock Joo \\
University of California, Los Angeles\\
{\tt\small chenyunliang@ucla.edu, jjoo@comm.ucla.edu}
\thanks{To appear in ICCV 2021.}
}

\maketitle

\begin{abstract}
The performance of a computer vision model depends on the size and quality of its training data. 
Recent studies have unveiled previously-unknown \textit{composition} biases in common image datasets which then lead to skewed model outputs, and have proposed methods to mitigate these biases. 
However, most existing works assume that human-generated annotations can be considered gold-standard and unbiased. In this paper, we reveal that this assumption can be problematic, and that special care should be taken to prevent models from learning such \textit{annotation} biases. We focus on facial expression recognition and compare the label biases between lab-controlled and in-the-wild datasets. We demonstrate that many expression datasets contain significant annotation biases between genders, especially when it comes to the happy and angry expressions, and that traditional methods cannot fully mitigate such biases in trained models. To remove expression annotation bias, we propose an AU-Calibrated Facial Expression Recognition (AUC-FER) framework that utilizes facial action units (AUs) and incorporates the triplet loss into the objective function. Experimental results suggest that the proposed method is more effective in removing expression annotation bias than existing techniques. 
\end{abstract}

\section{Introduction}
Computer vision models rely heavily on large sets of training images. Unfortunately, most datasets are "biased" in one way or another \cite{mehrabi2019survey}. Traditional (\ie, lab-controlled) datasets are often too small and not diverse enough to train a robust model. Recently, many large-scale image datasets have been created through web-scraping and crowdsourced annotations \cite{deng2009imagenet,zhou2014learning}. 
While this practice helps researchers collect millions of diverse ``in-the-wild'' images rapidly at low cost, it also introduces an undesired problem of dataset bias \cite{torralba2011unbiased,tommasi2017deeper,pan2009survey}. 
To mitigate the problem of biases effectively, we need to know (1) what causes biases (source), (2) which specific problems, datasets, or models suffer from biases, and (3) which methods are effective in each situation.
Machine learning models, unless explicitly modified, have been shown to be capable of learning bias from data \cite{el2016face} and, consequently, to produce biased outcomes against certain groups of people, undermining fairness and social trust of AI systems \cite{wang2019balanced,hendricks2018women,zhao2017men,bolukbasi2016man,buolamwini2018gender,drozdowski2020demographic,grother2019face,koenecke2020racial}. 



We consider the scenario of supervised learning. Let $X = \{X_i\}_{i=1}^N$ denote the collection of input images, and $Y = \{Y_i\}_{i=1}^N$ be the set of labels. A dataset is unbiased if the joint distribution $P(X, Y)$ matches reality. 
In particular, this requires the annotated labels, $Y|X$, to be unbiased.

For tabulated data, label bias is a classical focus in the fairness literature, where machine learning models are applied to some historically discriminatory data whose labels are unfair to certain racial or gender groups \cite{pedreshi2008discrimination}, such as recidivism prediction \cite{chouldechova2017fair},
loan approval, and employment decisions \cite{zafar2017fairness,olfat2018spectral}. For crowdsourced image annotations, however, it is often assumed that the annotations are not \textit{systematically} biased. Each annotator may have their personal biases, and there may be labeling mistakes, but given the diversity and large size of the data, they are generally assumed to be just another component of random noises \cite{beigman2009learning,zhuang2015leveraging}.

In reality, however, it is unlikely that people's biases are all idiosyncratic. In fact, annotators may possess \textit{systematic} cultural or societal biases, and if not specifically trained, they may incorporate such biases into their annotations. As a result, models trained on such data will become unfair. In this paper, we investigate the presence of systematic annotation bias in large in-the-wild datasets. 
We focus on the task of facial expression recognition. Fairness in expression recognition has not received wide attention \cite{xu2020investigating,rhue2018racial}, yet it has a profound impact: more and more companies nowadays conduct video job interviews in which algorithms are used to evaluate applicants' facial expressions, voice, and word selection to predict their skills, behaviors, and personality traits \cite{zetlin_2018,perry_2019,harwell_2019}; in addition, automated emotion analysis is already ubiquitous and used in consumer analysis, content recommendation, clinical psychology, lie detection, pain assessment, and many other human computer interfaces (\eg ``smile'' shutters) \cite{tian2011facial,pitaloka2017enhancing}.

In the context of facial expression recognition, studies in psychology have shown that human observers are more likely to perceive women's faces as happier than men's faces even when their smiles have the same intensity \cite{steephen2018we}, and it is believed that raters hold cultural stereotypes and that these stereotypes influence the judgment of emotions \cite{hess1997intensity,kilbride1983ethnic}. We hypothesize that such bias is present in many in-the-wild expression datasets whose labels are annotated by non-experts. In particular, we seek to answer the question: "Are annotators equally likely to assign different expression labels between males and females?" As we will show, for subjective tasks such as facial expression recognition, image annotations can be systematically biased, and special efforts need to be taken to address such bias.

We note that, currently, most debiasing techniques in the deep learning literature focus on biases that come from the images themselves (\ie, the bias in the distribution $P(X)$). This is often known as "dataset bias" \cite{torralba2011unbiased,tommasi2017deeper} or "sample selection bias" \cite{pan2009survey}. It happens when the dataset is biased in its \textit{composition} of images. As a result, models trained on one dataset do not generalize well to the real world due to the domain shift between the source and target. The trained model can also have undesirable accuracy differences across different groups or classes \cite{buolamwini2018gender}. 
Additionally, the data may contain spurious or undesirable correlations. 
When such undesirable correlation involves protected attributes (\eg, gender, race, or age), the model is considered "unfair." Numerous methods have been proposed to decorrelate these attributes and ensure that models trained on such data do not discriminate people based on their protected attributes \cite{robinson2020face,amini2019uncovering,iosifidis2018dealing,lu2020gender,ngxande2020bias,calmon2017optimized,morales2020sensitivenets,joo2020gender}.

However, debiasing $P(X)$ does not solve all problems since the joint distribution $P(X,Y)$ will still be biased if the annotated labels, $P(Y|X)$, are biased. As we will demonstrate in this paper, existing techniques that are designed to mitigate data \textit{composition} bias fail to fully mitigate the bias that comes from \textit{annotations}. On the other hand, classical methods that address label bias are intrusive in that they often involve changing the labels prior to training \cite{luong2011k,kamiran2009classifying}. In this paper, we address annotation bias that arises in facial expression recognition tasks. We propose an AU-Calibrated Facial Expression Recognition (AUC-FER) framework that uses the facial action units (AUs) and incorporates the triplet loss into the objective function to learn an embedding space in which the expressions are classified similarly for people with similar AUs. We demonstrate that the presented method more effectively mitigates annotation biases than existing methods. 
We note that although our framework is designed for facial expression recognition, it can be adapted to many other applications that require subjective human labeling such as activity recognition or image captioning but some fair or objective measures are available (such as the AUs and body keypoints).

The contribution of this paper is threefold:
\begin{itemize}
\item We compare the existence of annotation bias between lab-controlled datasets and in-the-wild datasets for facial expression recognition and observe that in-the-wild datasets often contain significant systematic bias in their annotations. To the best of our knowledge, this is the first work to demonstrate the effect of systematic annotation bias associated with image data.
\item We further demonstrate that such systematic annotation bias will be learned by trained models and thus cannot be ignored as is often assumed in the literature. 
\item We propose a novel AU-Calibrated Facial Expression Recognition (AUC-FER) framework that 
utilizes facial action units to remove expression annotation bias. Experiments suggest that it outperforms existing debiasing techniques for removing annotation bias. 
\end{itemize}


\section{Related Work}
As discussed in the previous section, the focus of this paper is the bias of $P(Y|X)$. We briefly review the literature on fairness and bias specific to this type, as well as research on facial expression recognition.

\textbf{Fairness.}
Fairness generally means that the model is not discriminatory with respect to some protected attribute, such as race, color, religion, sex, or national origin \cite{hutchinson201950}. Many formal definitions of fairness exist, and they generally can be divided into two types: \textit{group fairness}, which requires different demographic groups to receive the same treatment on average \cite{hardt2016equality}, and \textit{individual fairness}, which requires individuals who are similar to have similar probability distributions on classification outcomes \cite{dwork2018fairness}. As is common notation, we will denote the \textit{protected variable} by $Z$ and the model prediction by $\hat{Y}$.
A major barrier to achieving individual fairness is the selection of a similarity measure between individuals. A closely related concept is \textit{counterfactual fairness}, which requires the decision to be unchanged had the person belonged to a different demographic group while keeping everything else the same \cite{kusner2017counterfactual}. 
Denton \etal \cite{denton2019detecting} build on this idea and use a generative model that can manipulate specific attributes of faces (\eg, from \textit{young} to \textit{old}) to reveal the biases of a smile classifier.

\textbf{Debiasing techniques.}
Common techniques to address dataset bias include transfer learning \cite{pan2009survey}, domain adaptation \cite{tzeng2015simultaneous,tzeng2017adversarial,ganin2016domain,wang2019racial}, and adversarial mitigation~\cite{zhang2018mitigating,wang2019balanced}. Many methods have also been proposed to remove or prevent models from learning spurious or undesirable correlations. 
Hardt \etal \cite{hardt2016equality} propose a post-hoc correction technique that enforces equality of odds on a learned predictor. Other group fairness definitions have also been transformed into constrained optimization problems \cite{olfat2018spectral,zhao2017men,zafar2017fairnessbeyond,zafar2017fairness}. 
Robinson \etal \cite{robinson2020face} propose learning subgroup-specific thresholds.
In the realm of deep learning, modifying the loss functions to penalize unfairness \cite{alvi2018turning} and adversarial learning \cite{raff2018gradient,zhang2018mitigating,jia2018right,morales2020sensitivenets} 
are two common techniques, with the goal of learning a "fair" representation that does not contain information of the protected attribute $Z$.

In the case where the data labels are historically biased, data massaging is the most commonly used technique. This includes directly correcting the labels by changing them prior to training \cite{luong2011k,kamiran2009classifying}, or use some weights or sampling techniques during training \cite{kamiran2010classification,kamiran2012data}.

\textbf{Annotation bias.}
For tabulated data, historical label bias is a well-known issue \cite{pedreshi2008discrimination}. 
Jiang and Nachum \cite{jiang2020identifying} propose a re-weighting scheme that can correct label bias under certain assumptions about the relationship between the biased labels and the true labels. 
In the case of large-scale in-the-wild datasets prepared for deep learning, however, annotation bias has received little attention compared to the more salient data composition bias. 
Regardless of the exact methods through which the images are labeled (manual, semi-automatic, or automatic), the general assumption is that they add random noise to the labels but are unbiased on average \cite{beigman2009learning,zhuang2015leveraging}. In the case where each image is annotated by multiple workers, the focus has been on improving the compilation step of the dataset creation process to increase the accuracy of the labels. 
Methods have also been developed to fix errors in the case of multi-label supervised learning \cite{cui2020label}.
Zhuang and Young \cite{zhuang2015leveraging} note that presenting data items in batches to annotators can lead to in-batch annotation bias. In general, crowd annotators have lower accuracy when labeling difficult cases, but researchers have found that this is relatively unproblematic under certain conditions \cite{beigman2009learning}. In this paper, we examine the bias of labels in the case of facial expression recognition, and we will show that, unlike what previous studies assumed, \textit{systematic} bias exists and needs to be actively managed.

\textbf{Facial expression recognition and facial action units.}
Facial expression recognition, which analyzes people's \textit{expressed} emotions from visual data \cite{tian2011facial}, is one of the central tasks in facial analysis and widely used in many domains such as media analytics~\cite{joo2019automated,xi2020understanding,peng2018same}, HCI~\cite{bartlett2003real,cowie2001emotion}, education~\cite{khalfallah2015facial}, and psychology~\cite{lewinski2014automated,cohn1999automated}. A seminal study conducted by Ekman and Friesen \cite{ekman1971constants} 
identified six prototypical emotions: anger, disgust, fear, happiness, sadness, and surprise. They noticed that the association between certain facial muscular patterns and discrete emotions is universal and independent of gender and race, and adopted a Facial Action Coding System (FACS) consisting of facial action units (AUs) \cite{ekman1978action} that objectively code the fundamental muscle actions typically seen for various facial expressions of emotion \cite{du2014compound}. 
Early works in facial expression recognition are often rule-based methods using FACS \cite{tian2000recognizing}.

With deep learning, the average performance of expression recognition has significantly improved, and many works recently started to focus on model bias and dataset (composition) bias. A common observation is that disgust, anger, fear, and surprise are minority classes in datasets and harder to learn compared to happiness and sadness \cite{li2020deep}, and classical methods for addressing data composition bias such as weighting, re-sampling, data augmentation, hierarchical modeling \cite{howard2017addressing}, and confusion loss \cite{xu2020investigating} have been proposed.

In another line of research, studies have shown that women look happier than men \cite{steephen2018we} and that people are faster and more accurate at detecting angry expressions on male faces and happy expressions on female faces \cite{becker2007confounded}. As a result, correction of the annotations is necessary \cite{steephen2018we}. On a similar note, 
Denton \etal \cite{denton2019detecting} find that a smiling classifier trained on CelebA is more likely to predict "smiling" when they remove the person's beard or apply makeup or lipstick to the image but keep everything else the same. Based on these psychological studies as well as the observed model bias, we hypothesize that systematic annotation bias exists in many large in-the-wild expression datasets and it (in addition to the data composition bias) contributes to the gender bias in trained models.



\section{Annotation Bias in Expression Datasets}

In this section, we illustrate the existence of systematic annotation bias in facial expression datasets. As previously noted, psychological studies have shown that raters tend to hold stereotypical biases that women are happier than men \cite{hess1997intensity} and that they detect angry expressions on male faces more quickly 
\cite{becker2007confounded}, we hypothesize that these biases will manifest themselves in annotated datasets. In particular, we examine the ``happiness'' and ``anger'' annotations and ask the question: Are annotators equally likely to assign happiness/anger labels to male and female images -- if they indeed show the same expression? In order to quantify the ``same'' expression, we use the AUs since they were specifically designed to measure facial expressions objectively, and past studies have used them to assess the accuracy in the imitation of facial expressions \cite{kilbride1983ethnic}.
\revision{We mainly focus on gender due to its well-studied psychological connection to expression perception. We also conduct analysis on age and race. However, unlike gender, most public datasets used in our experiments are not well-balanced between different age and racial groups but are instead heavily dominated by younger and white people. The full analysis on age and race is included in the Supplementary Material.}

\subsection{Facial Action Units (AUs) Recognition}\label{subsec:OpenFace_fairness}

In the framework of FACS, happiness is defined as the combination of AU6 (cheeks raised and eyes narrowed) and AU12 (lip corners pulled up and laterally), and anger is defined as the combination of AU4 (brow lowerer), AU5 (upper lid raiser), AU7 (lid tightener), and AU23 (lip tightener) \cite{matsumoto2008facial,kanade2000comprehensive}. Therefore, we will use them as objective benchmarks to evaluate the classification of emotions. Due to limited space, we include in the paper the numerical results for the happiness expression only; 
detailed analysis for anger is presented in the Supplementary Material.

We use OpenFace, a state-of-the-art facial behavior analysis toolkit \cite{baltruvsaitis2015cross}, for our facial action unit recognition purpose. In order for it to serve as a benchmark for evaluating the bias of emotion annotations, we first check that its AU recognition is not biased between males and females itself. 

We use EmotioNet \cite{emotionet}, which includes 24,600 images with AUs manually annotated by experienced coders, to evaluate the performance of AU presence and intensity recognition by OpenFace. 
Since OpenFace and EmotioNet use different thresholds when binarizing the AU variables, we use OpenFace's AU intensity output to re-classify AU presence by choosing the threshold that optimizes the overall classification accuracy for each AU based on EmotioNet annotations. 

We use the FairFace dataset \cite{karkkainen2021fairface} to train a simple gender classifier that achieves a test accuracy of 94.5\%. We then use it to classify the 24,600 EmotioNet images and this enables us to test whether the performance of OpenFace differs by gender. Table \ref{tab:AU_accuracy} summarizes the accuracies and F1-scores for the calibrated OpenFace AU6 and AU12 output between males and females. We can see from the p-values of the t-tests for the accuracies that the differences are insignificant for both AU6 and AU12. Therefore, we conclude that even though OpenFace's AU6 and AU12 recognition is imperfect, it is unbiased between males and females and thus can be used as a proxy for the true AUs and as an objective benchmark for evaluating happiness annotations. A similar evaluation is conducted for AUs associated with the angry expression; see the Supplementary Material for details). 

\begin{table}
\begin{center}
\small
\begin{tabular}{|l||cc|cc|} 
\hline
      & \multicolumn{2}{c|}{AU6} & \multicolumn{2}{c|}{AU12}  \\ 
\hline
  & Accuracy & F1      & Accuracy & F1       \\ 
\hline\hline
Male    & 0.859    & 0.613         & 0.887    & 0.830          \\
Female  & 0.860    & 0.598         & 0.885    & 0.866          \\ 
\hline
p-value & 0.835    &               & 0.715    &                \\
\hline
\end{tabular}
\end{center}
\vspace{-5pt}
\caption{Accuracies and F1-scores of OpenFace AU Recognition, evaluated on 24,600 EmotioNet images with expert-coded AUs.}
\label{tab:AU_accuracy}
\vspace{-10pt}
\end{table}

\subsection{Expression Annotation Bias}

As we mentioned previously, there are two potential sources of bias that in-the-wild datasets may contain: \textit{data composition bias} (\eg, the data contains significantly more happy women and unhappy men) and \textit{annotation bias} (\eg, even when two images are the same otherwise, a woman is more likely to be annotated as "happy" than a man). Since expressions are objectively defined as combinations of AUs, those respective AUs can help make the important distinction between these two biases.

\begin{definition}\textsc{Annotation Bias.} Let $Y \in \{0,1\}$ denote the emotion label. Let $Z \in \{M,F\}$ denote the gender (or some other protected attributes) of the person. We say that the expression annotations are unbiased if
\begin{equation}
Y \perp \!\!\! \perp  Z |AU
\end{equation}
\end{definition}
For happiness annotations, this means
\begin{equation}
\begin{split}
P(Y=1|AU6,AU12,Z=\text{M})
\\ = P(Y=1|AU6,AU12,Z=\text{F})
\end{split}
\end{equation}
where the AUs can be discrete (\ie, $(AU6, AU12) \in \{(0,0), (0,1), (1,0), (1,1)\}$) or continuous (\ie, intensity scores). The case for anger annotations is similar.

\textbf{Remark.} This definition is similar to equality of odds ($\hat{Y} \perp \!\!\! \perp  Z |Y$) except that each image is conditioned on the AUs and the requirement is to the labels $Y$ instead of model predictions $\hat{Y}$. Note that the conditioning on AU is crucial because otherwise, we would not have been able to separate annotation bias from data composition bias (that is, it is possible that the female images in the dataset are less happy than males on average, but they are annotated with a larger probability to be "happy" and so it looks as if the data does not contain any bias). 

\subsection{Evaluation on Various Datasets}

We evaluate the expression annotations on various popular expression datasets. They can be categorized into two types: those whose images were collected in a laboratory-controlled condition and those whose images were scraped from the web (\ie, "in-the-wild"). For the first type, we select the Karolinska Directed Emotional Faces database (KDEF) \cite{lundqvist1998karolinska} and the Chicago Face Database (CFD) \cite{ma2015chicago}. For the second type, we select the Expression in-the-Wild Database (ExpW) \cite{SOCIALRELATION_ICCV2015,SOCIALRELATION_2017}, the Real-world Affective Face Database (RAF-DB) \cite{li2017reliable}, and AffectNet \cite{mollahosseini2017affectNet}.

\textbf{KDEF} \cite{lundqvist1998karolinska}: KDEF contains 70 individuals displaying the 6 basic expressions plus neutral. Each expression is viewed from 5 angles and shot twice. However, for comparability with other databases, we will only use the 980 front-view photos among the 4,900 images.

\textbf{CFD} \cite{ma2015chicago}: CFD contains photos of 597 individuals with a neutral expression. For a subset of 158 targets, it also includes happy, angry, and fearful expressions.

\textbf{ExpW} \cite{SOCIALRELATION_ICCV2015,SOCIALRELATION_2017}: ExpW is an in-the-wild dataset consisting of 91,793 faces. Each face is manually annotated as one of the 6 basic expressions plus neutral. 

\textbf{RAF-DB} \cite{li2017reliable}: RAF-DB contains 29,672 facial images downloaded from the web. Using crowdsourcing, each image is independently labeled by about 40 annotators. In particular, 15,339 of them are classified into one of the 6 basic expressions plus neutral. Gender, age, and race annotations are also provided.

\textbf{AffectNet} \cite{mollahosseini2017affectNet}: AffectNet contains about 1M facial images collected from the web. About half (420K) of the images (denoted as AffectNet-Manual) are manually annotated as one of the 6 basic expressions plus contempt and neutral. The rest (550K) (denoted as AffectNet-Automatic) are automatically annotated using ResNext Neural Network trained on all manually annotated training set samples with average accuracy of 65\%. For the purpose of our evaluation, we will use random samples of size 38,889 and 45,369 for AffectNet-Manual and AffectNet-Automatic respectively instead of the entire datasets.

\begin{figure}
\centering
\includegraphics[width=0.37\textwidth]{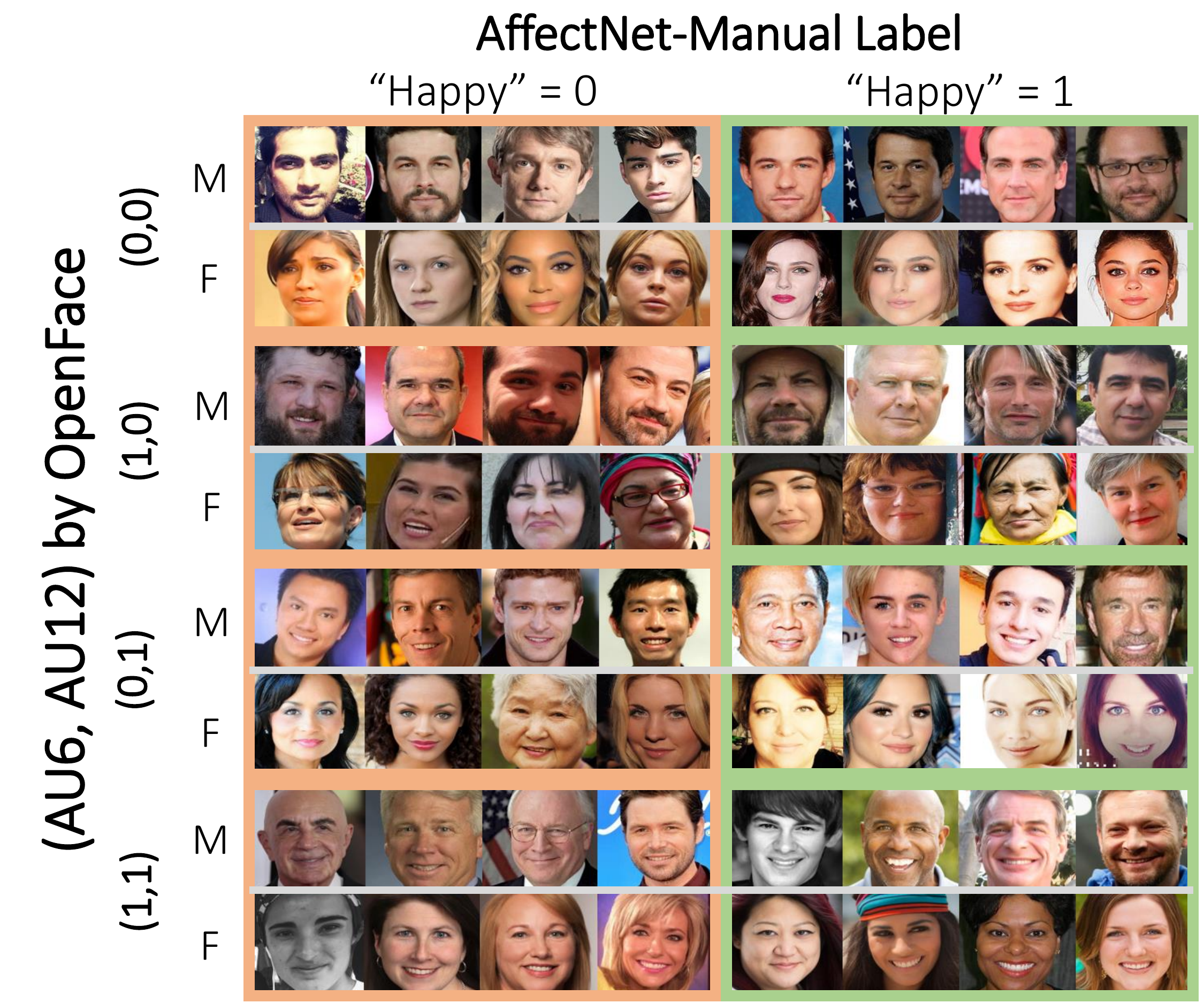}
\caption{Examples of "happy" and "not happy" faces from AffectNet-Manual for each (AU6, AU12) combination and for each gender. The emotion labels in AffectNet-Manual come from manual annotation but possibly contain errors, and the AUs are detected using OpenFace.}
\label{fig:Sample_faces}
\vspace{-10pt}
\end{figure}

For each of the above datasets, we apply the OpenFace AU detector and obtain the AU6 and AU12 intensities for each image. They are then binarized into AU presence variables using the optimal thresholds found in Section \ref{subsec:OpenFace_fairness}. We also apply our gender classifier when gender information is not available (\ie, for ExpW and AffectNet). Note that even though "happiness" is formally defined to be the presence of AU6 and AU12, the fact that both the expression and the AUs are inherently continuous-valued means that discretization may result in a few cases that violate the rule. In practice, AU detection and expression annotation are imperfect and will introduce additional noises. Nevertheless, the pattern should be similar between males and females if the errors are random. See Figure \ref{fig:Sample_faces} for some examples of "happy" and "not happy" faces from AffectNet-Manual for each (AU6, AU12) combination.

\begin{figure}
\centering
\includegraphics[width=0.18\textwidth]{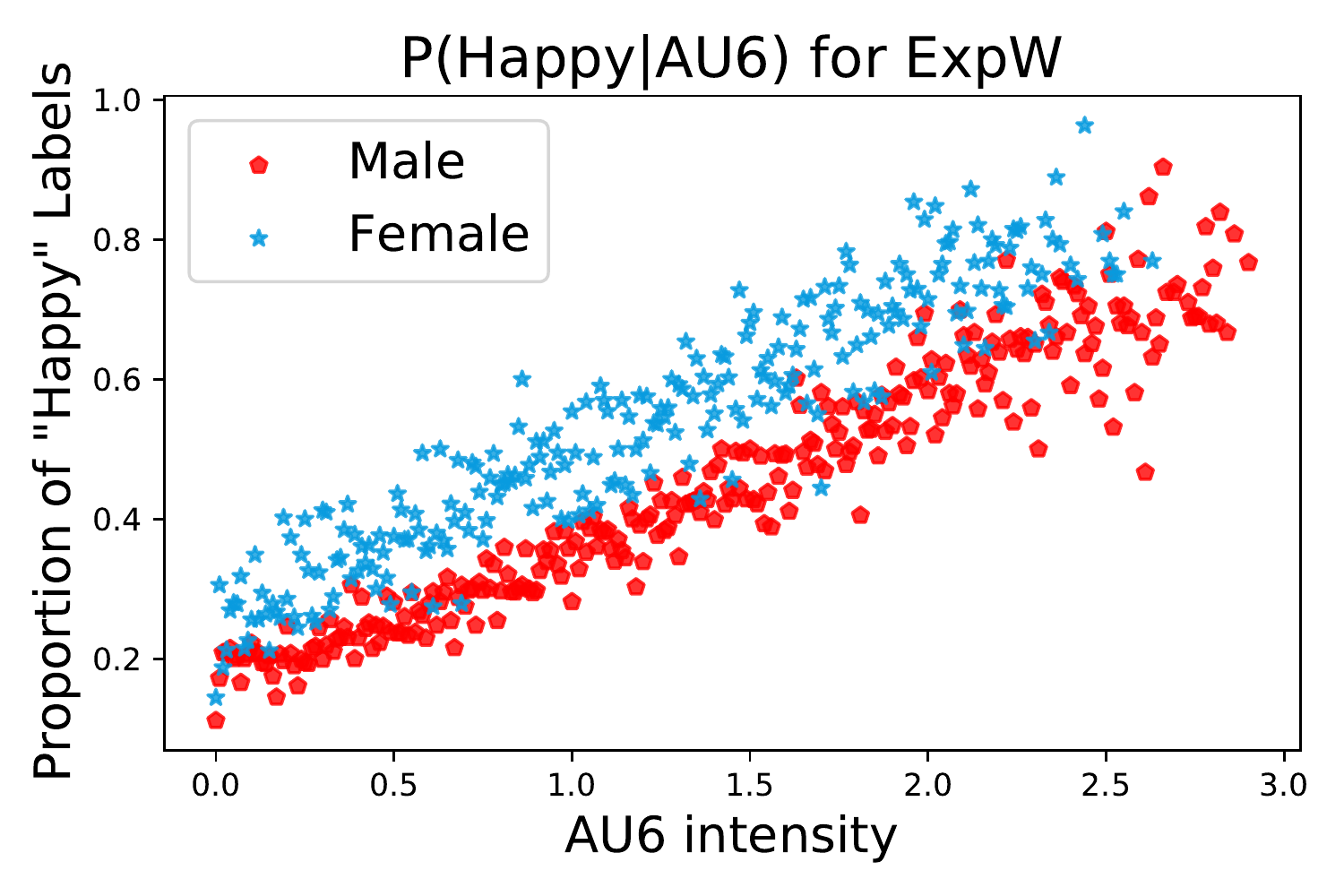}
\includegraphics[width=0.18\textwidth]{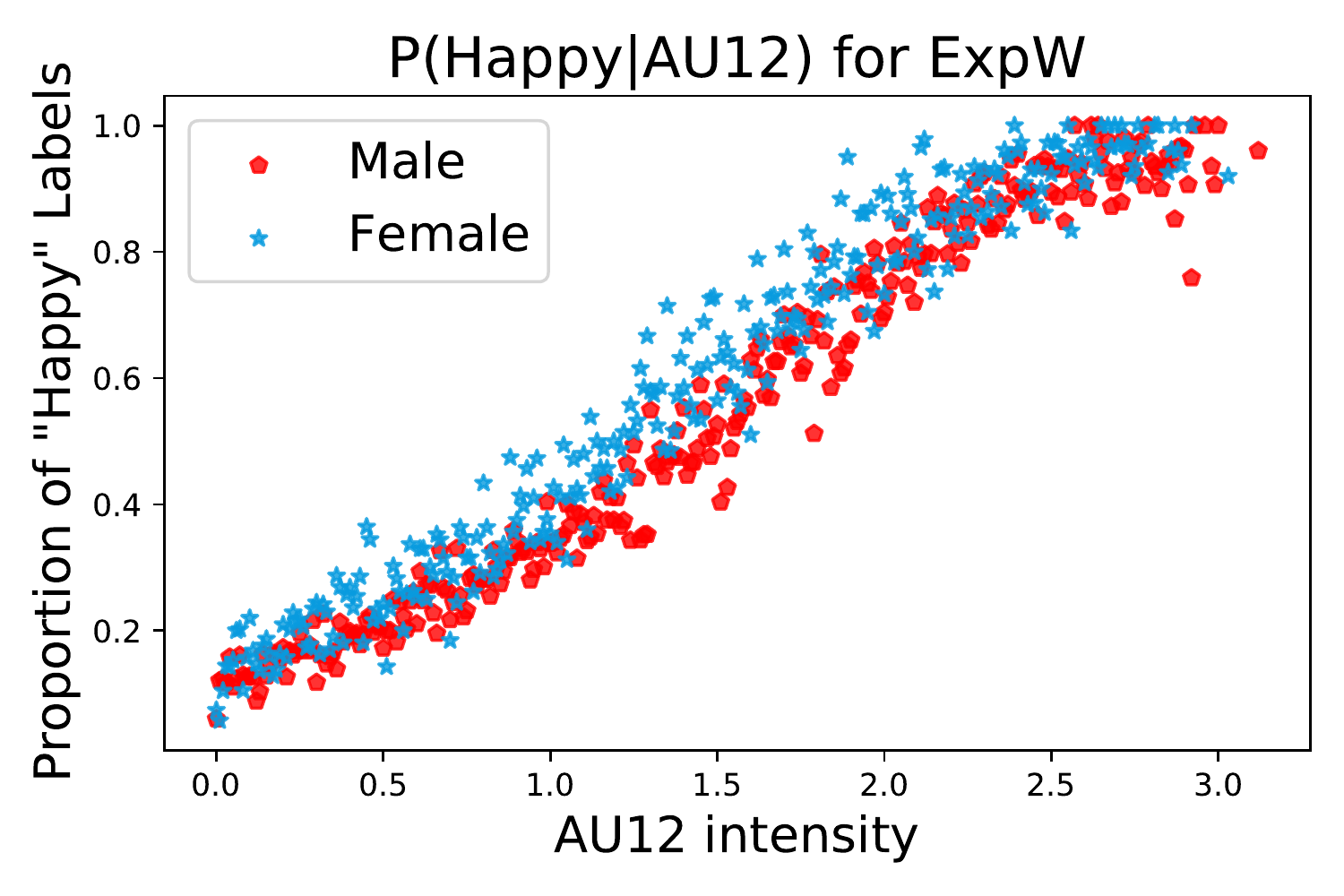} 
\includegraphics[width=0.18\textwidth]{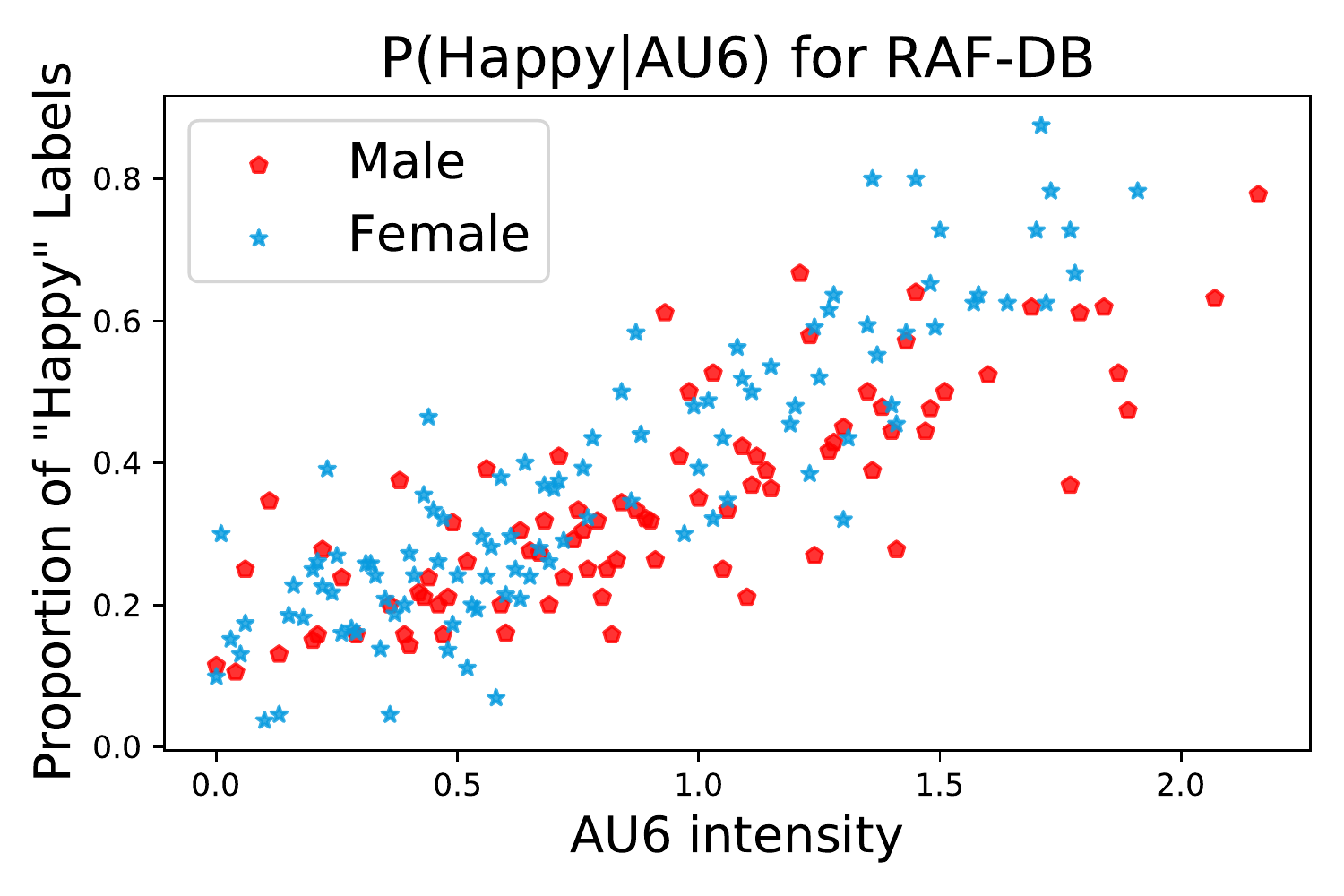}
\includegraphics[width=0.18\textwidth]{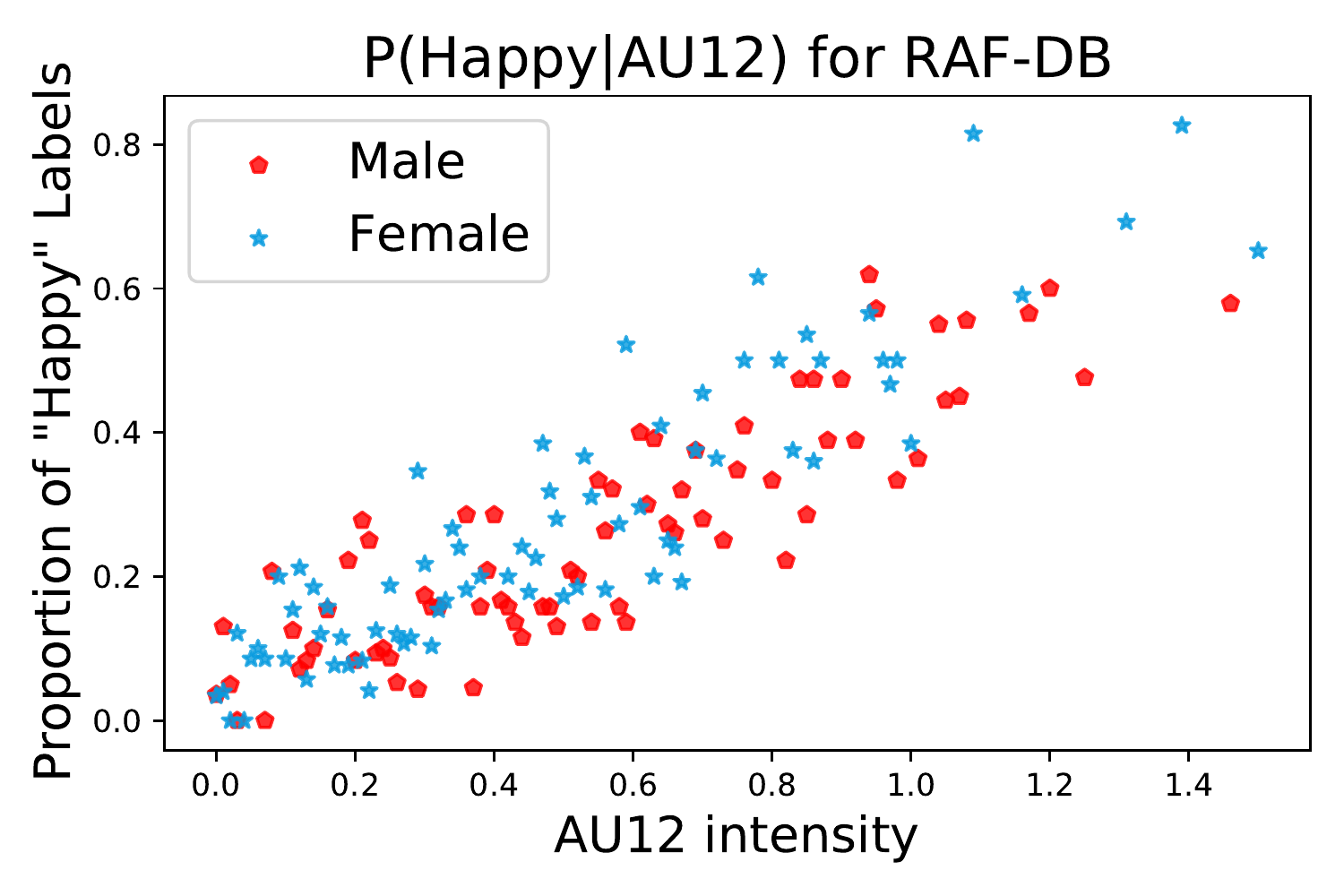}
\includegraphics[width=0.18\textwidth]{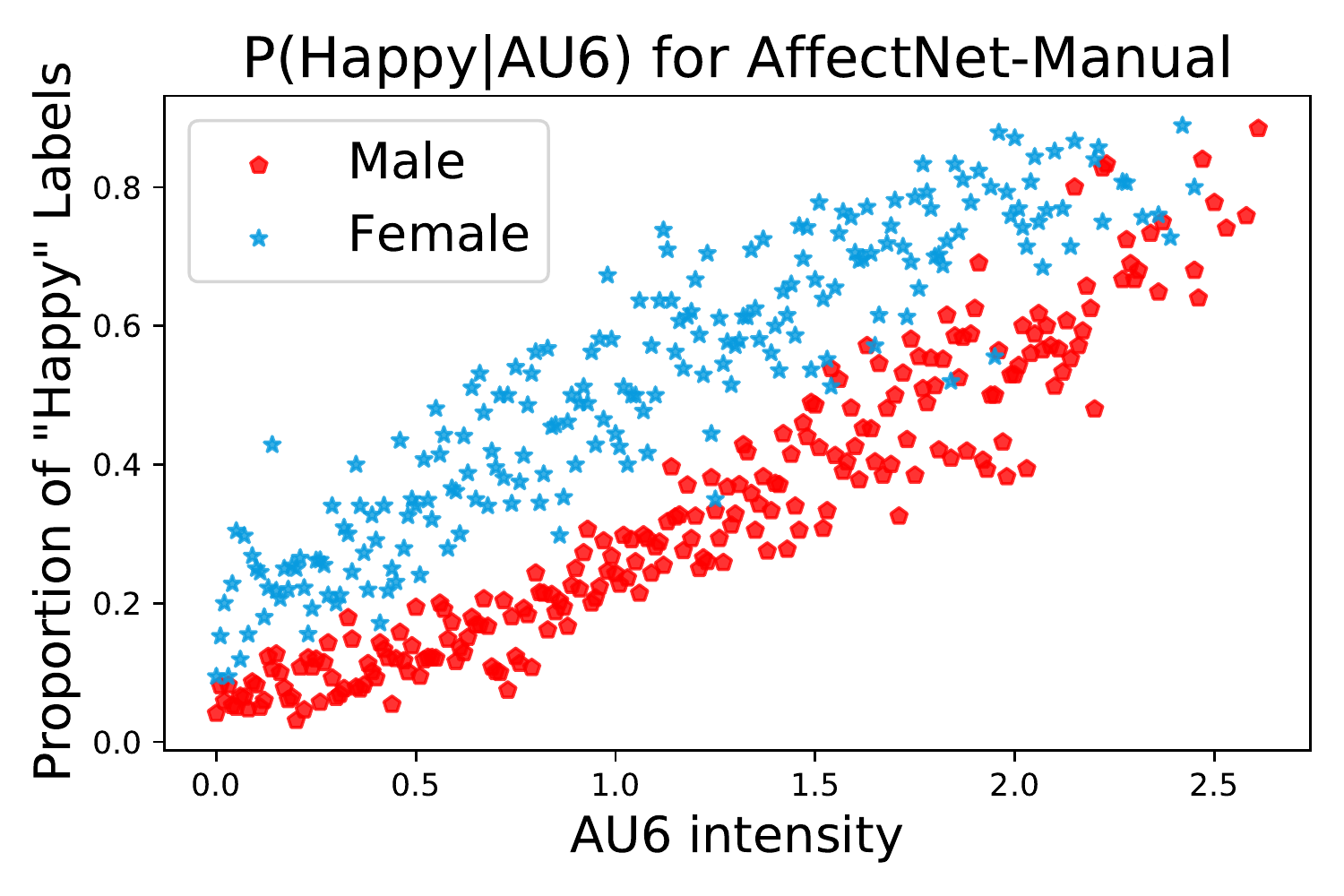}
\includegraphics[width=0.18\textwidth]{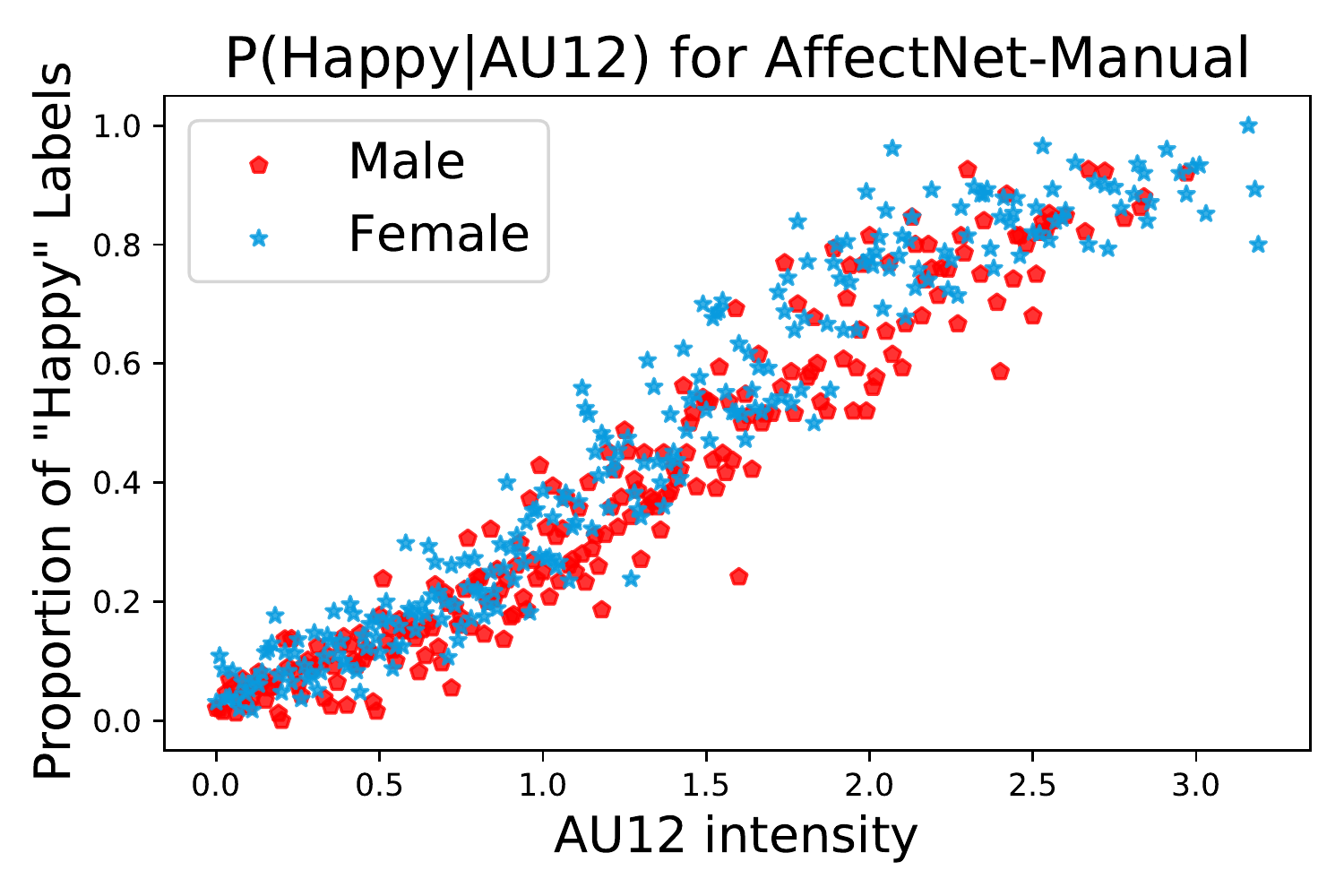}
\includegraphics[width=0.18\textwidth]{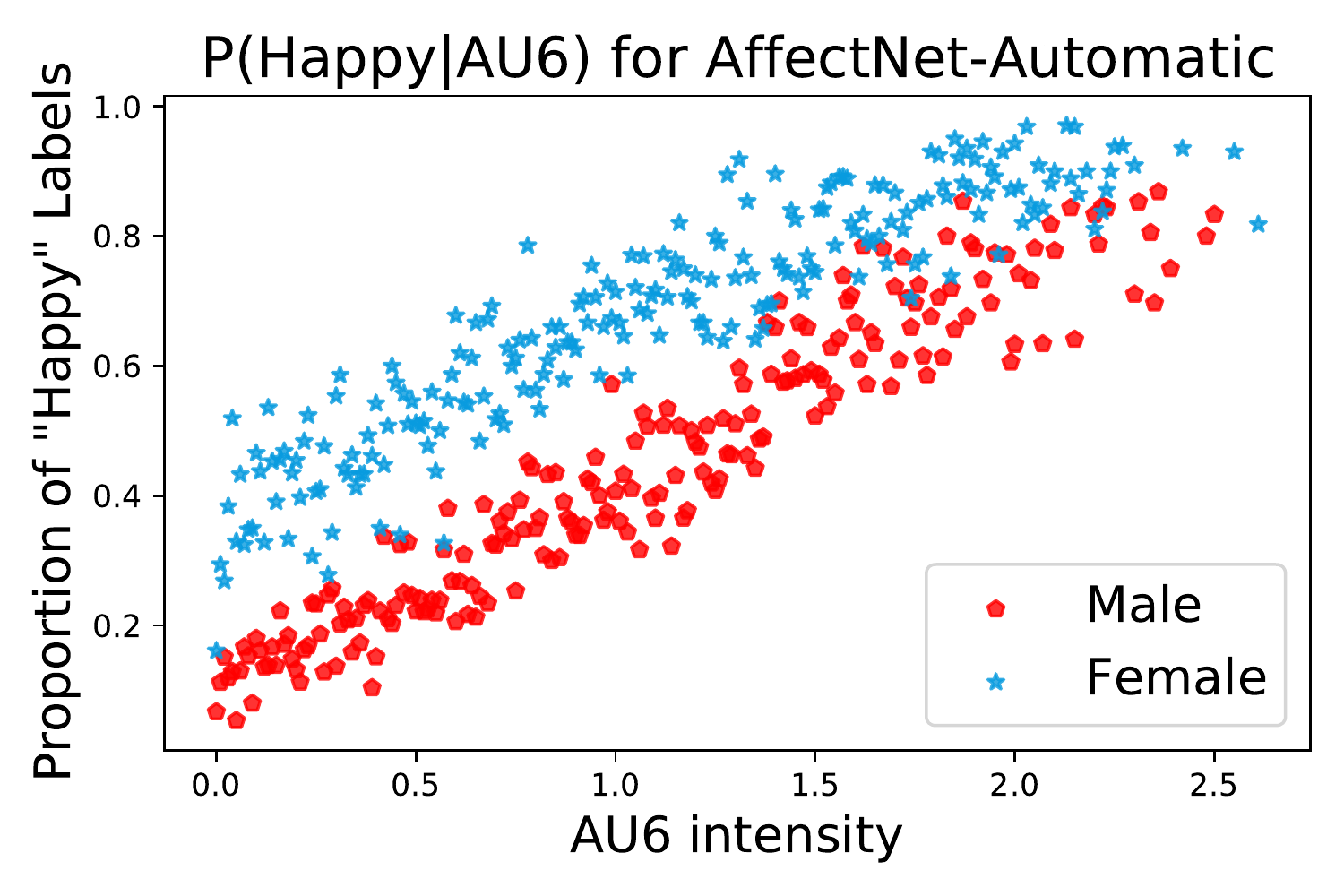}
\includegraphics[width=0.18\textwidth]{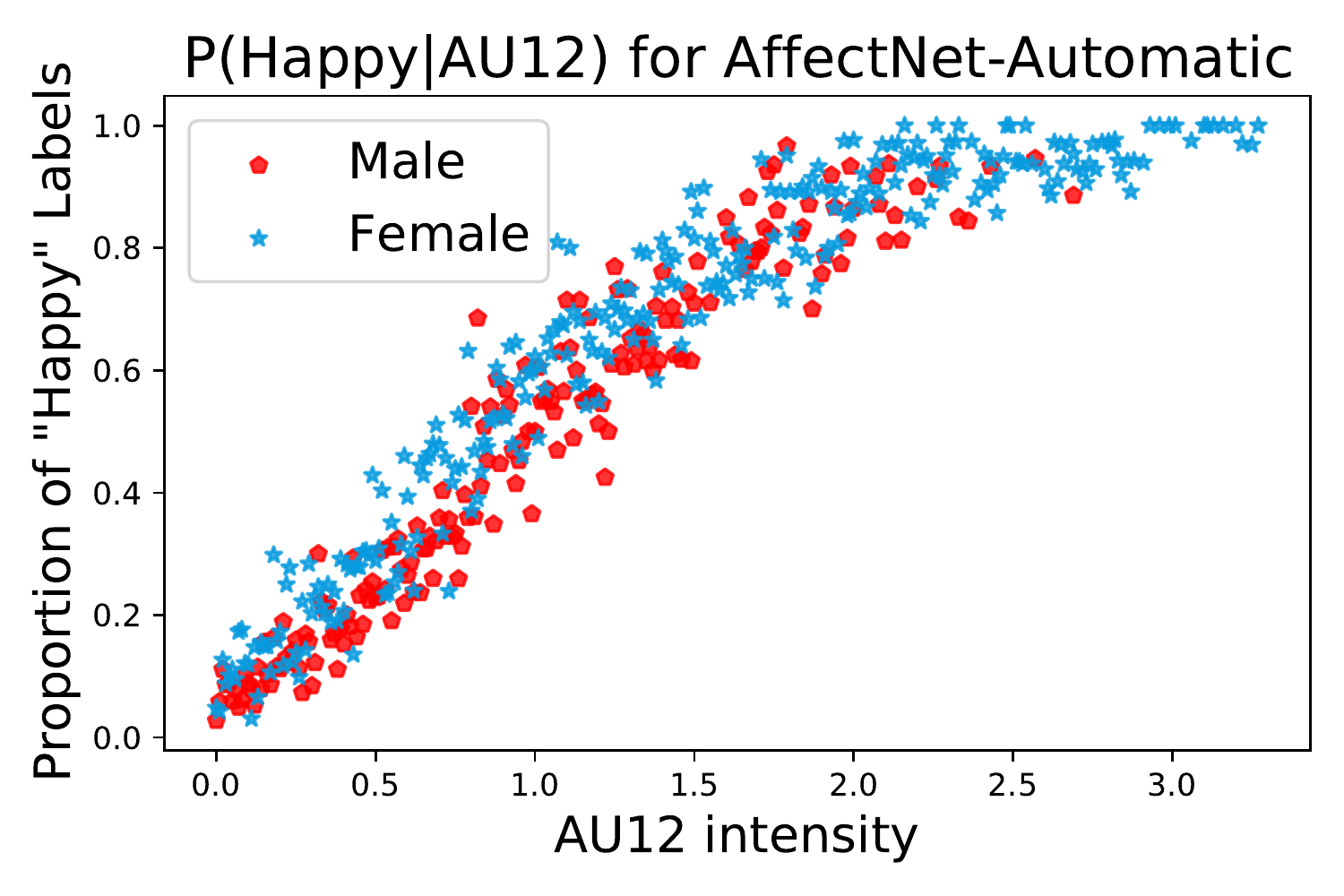}
\caption{Proportion of "happy" labels among males and females conditioned on AU6 and AU12 intensity for each "in-the-wild" expression dataset. Significant differences can be seen between males and females, suggesting the presence of annotation bias.}
\label{fig:Anno_bias_scatterplot}
\vspace{-10pt}
\end{figure}

\begin{table*}[t]
\small
\begin{center}
\begin{threeparttable}
\scalebox{0.85}{
\begin{tabular*}{1\textwidth}{|m{0.1\linewidth}||m{0.03\linewidth}|m{0.04\linewidth}|m{0.04\linewidth}|m{0.04\linewidth}|m{0.08\linewidth}||m{0.045\linewidth}|m{0.04\linewidth}|m{0.04\linewidth}|m{0.045\linewidth}|m{0.08\linewidth}|}
\cline{1-11}
\multirow{2}{\linewidth}{\parbox{1\linewidth}{\vspace{0.1cm} \centering Data \\ (Collecting Condition, Size)}} & \multicolumn{5}{c||}{Conditioned on Joint AU} & \multicolumn{5}{c|}{Conditioned on Marginal AU} \\ 
\cline{2-11}
 & \multicolumn{1}{>{\centering\arraybackslash}m{0.05\linewidth}|}{(AU6, AU12)} & \multicolumn{1}{>{\centering\arraybackslash}m{0.065\linewidth}|}{P(Happy \textbar{}AU, M)} & \multicolumn{1}{>{\centering\arraybackslash}m{0.06\linewidth}|}{P(Happy \textbar{}AU, F)} & \centering $\Delta$ & \multicolumn{1}{>{\centering\arraybackslash}m{0.08\linewidth}||}{p-value of $\chi^2$ test for $Y \perp \!\!\! \perp  Z$} & \multicolumn{1}{>{\centering\arraybackslash}m{0.06\linewidth}|}{AU} & \multicolumn{1}{>{\centering\arraybackslash}m{0.065\linewidth}|}{P(Happy \textbar{}AU, M)} & \multicolumn{1}{>{\centering\arraybackslash}m{0.06\linewidth}|}{P(Happy \textbar{}AU, F)} & \centering $\Delta$ & \multicolumn{1}{>{\centering\arraybackslash}m{0.08\linewidth}|}{p-value of $\chi^2$ test for $Y \perp \!\!\! \perp  Z$} \\ 
\hhline{|======::=====|}
\multirow{4}{\linewidth}{\centering KDEF (Lab, 980) \cite{lundqvist1998karolinska}} &  &  &  &  &  & \multicolumn{1}{l|}{AU6=0} & \multicolumn{1}{c|}{0.036} & \multicolumn{1}{c|}{0.016} & \multicolumn{1}{c|}{-0.019} & 0.095 .\tnote{1} \\
 &  &  &  &  &  & \multicolumn{1}{l|}{AU6=1} & \multicolumn{1}{c|}{0.475} & \multicolumn{1}{c|}{0.547} & \multicolumn{1}{c|}{0.072} & 0.268 \\
 &  &  &  &  &  & \multicolumn{1}{l|}{AU12=0} & \multicolumn{1}{c|}{0.000} & \multicolumn{1}{c|}{0.005} & \multicolumn{1}{c|}{0.005} & - \\
 & \multicolumn{1}{c|}{(1,1)} & \multicolumn{1}{c|}{0.838} & \multicolumn{1}{c|}{0.771} & \multicolumn{1}{c|}{-0.067} & 0.304 & \multicolumn{1}{l|}{AU12=1} & \multicolumn{1}{c|}{0.769} & \multicolumn{1}{c|}{0.673} & \multicolumn{1}{c|}{-0.096} & 0.140 \\ 
\cline{1-11}
\multirow{4}{\linewidth}{\centering CFD \\ (Lab, 1,207) \cite{ma2015chicago}} &  &  &  &  &  & \multicolumn{1}{l|}{AU6=0} & \multicolumn{1}{c|}{0.059} & \multicolumn{1}{c|}{0.079} & \multicolumn{1}{c|}{0.021} & 0.222 \\
 &  &  &  &  &  & \multicolumn{1}{l|}{AU6=1} & \multicolumn{1}{c|}{0.838} & \multicolumn{1}{c|}{0.854} & \multicolumn{1}{c|}{0.016} & 0.706 \\
 & \multicolumn{1}{c|}{(0,1)} & \multicolumn{1}{c|}{0.383} & \multicolumn{1}{c|}{0.487} & \multicolumn{1}{c|}{0.104} & 0.228 & \multicolumn{1}{l|}{AU12=0} & \multicolumn{1}{c|}{0.005} & \multicolumn{1}{c|}{0.005} & \multicolumn{1}{c|}{-0.001} & - \\
 & \multicolumn{1}{c|}{(1,1)} & \multicolumn{1}{c|}{0.884} & \multicolumn{1}{c|}{0.890} & \multicolumn{1}{c|}{0.006} & 0.877 & \multicolumn{1}{l|}{AU12=1} & \multicolumn{1}{c|}{0.725} & \multicolumn{1}{c|}{0.751} & \multicolumn{1}{c|}{0.026} & 0.546 \\ 
\cline{1-11}
\multirow{4}{\linewidth}{\centering ExpW (Web, 91,793) \cite{SOCIALRELATION_ICCV2015,SOCIALRELATION_2017}} & \multicolumn{1}{c|}{(0,0)} & \multicolumn{1}{c|}{0.176} & \multicolumn{1}{c|}{0.215} & \multicolumn{1}{c|}{0.039} & 0.000 *** & \multicolumn{1}{l|}{AU6=0} & \multicolumn{1}{c|}{0.255} & \multicolumn{1}{c|}{0.336} & \multicolumn{1}{c|}{0.081} & 0.000 *** \\
 & \multicolumn{1}{c|}{(1,0)} & \multicolumn{1}{c|}{0.246} & \multicolumn{1}{c|}{0.285} & \multicolumn{1}{c|}{0.040} & 0.0488 * & \multicolumn{1}{l|}{AU6=1} & \multicolumn{1}{c|}{0.646} & \multicolumn{1}{c|}{0.770} & \multicolumn{1}{c|}{0.124} & 0.000 *** \\
 & \multicolumn{1}{c|}{(0,1)} & \multicolumn{1}{c|}{0.663} & \multicolumn{1}{c|}{0.770} & \multicolumn{1}{c|}{0.107} & 0.000 *** & \multicolumn{1}{l|}{AU12=0} & \multicolumn{1}{c|}{0.179} & \multicolumn{1}{c|}{0.217} & \multicolumn{1}{c|}{0.039} & 0.000 *** \\
 & \multicolumn{1}{c|}{(1,1)} & \multicolumn{1}{c|}{0.801} & \multicolumn{1}{c|}{0.870} & \multicolumn{1}{c|}{0.069} & 0.000 *** & \multicolumn{1}{l|}{AU12=1} & \multicolumn{1}{c|}{0.716} & \multicolumn{1}{c|}{0.806} & \multicolumn{1}{c|}{0.091} & 0.000 *** \\ 
\cline{1-11}
\multirow{4}{\linewidth}{\centering RAF-DB (Web, 15,339) \cite{li2017reliable}} & \multicolumn{1}{c|}{(0,0)} & \multicolumn{1}{c|}{0.197} & \multicolumn{1}{c|}{0.192} & \multicolumn{1}{c|}{-0.005} & 0.570 & \multicolumn{1}{l|}{AU6=0} & \multicolumn{1}{c|}{0.289} & \multicolumn{1}{c|}{0.305} & \multicolumn{1}{c|}{0.016} & 0.089 . \\
 & \multicolumn{1}{c|}{(1,0)} & \multicolumn{1}{c|}{0.232} & \multicolumn{1}{c|}{0.254} & \multicolumn{1}{c|}{0.022} & 0.537 & \multicolumn{1}{l|}{AU6=1} & \multicolumn{1}{c|}{0.632} & \multicolumn{1}{c|}{0.785} & \multicolumn{1}{c|}{0.153} & 0.000 *** \\
 & \multicolumn{1}{c|}{(0,1)} & \multicolumn{1}{c|}{0.822} & \multicolumn{1}{c|}{0.868} & \multicolumn{1}{c|}{0.047} & 0.013 * & \multicolumn{1}{l|}{AU12=0} & \multicolumn{1}{c|}{0.200} & \multicolumn{1}{c|}{0.195} & \multicolumn{1}{c|}{-0.005} & 0.572 \\
 & \multicolumn{1}{c|}{(1,1)} & \multicolumn{1}{c|}{0.808} & \multicolumn{1}{c|}{0.905} & \multicolumn{1}{c|}{0.097} & 0.000 *** & \multicolumn{1}{l|}{AU12=1} & \multicolumn{1}{c|}{0.814} & \multicolumn{1}{c|}{0.888} & \multicolumn{1}{c|}{0.074}& 0.000 *** \\ 
\cline{1-11}
\multirow{4}{\linewidth}{\centering AffectNet-Manual (Web, 420,299) \cite{mollahosseini2017affectNet}} & \multicolumn{1}{c|}{(0,0)} & \multicolumn{1}{c|}{0.086} & \multicolumn{1}{c|}{0.125} & \multicolumn{1}{c|}{0.039} & 0.000 *** & \multicolumn{1}{l|}{AU6=0} & \multicolumn{1}{c|}{0.165} & \multicolumn{1}{c|}{0.292} & \multicolumn{1}{c|}{0.127} & 0.000 *** \\
 & \multicolumn{1}{c|}{(1,0)} & \multicolumn{1}{c|}{0.251} & \multicolumn{1}{c|}{0.254} & \multicolumn{1}{c|}{0.004} & 0.920 & \multicolumn{1}{l|}{AU6=1} & \multicolumn{1}{c|}{0.676} & \multicolumn{1}{c|}{0.821} & \multicolumn{1}{c|}{0.145} & 0.000 *** \\
 & \multicolumn{1}{c|}{(0,1)} & \multicolumn{1}{c|}{0.608} & \multicolumn{1}{c|}{0.725} & \multicolumn{1}{c|}{0.117} & 0.000 *** & \multicolumn{1}{l|}{AU12=0} & \multicolumn{1}{c|}{0.093} & \multicolumn{1}{c|}{0.127} & \multicolumn{1}{c|}{0.034} & 0.000 *** \\
 & \multicolumn{1}{c|}{(1,1)} & \multicolumn{1}{c|}{0.778} & \multicolumn{1}{c|}{0.860} & \multicolumn{1}{c|}{0.082} & 0.000 *** & \multicolumn{1}{l|}{AU12=1} & \multicolumn{1}{c|}{0.699} & \multicolumn{1}{c|}{0.781} & \multicolumn{1}{c|}{0.082} & 0.000 *** \\ 
\cline{1-11}
\multirow{4}{\linewidth}{\centering AffectNet-Automatic (Web, 539,607) \cite{mollahosseini2017affectNet}} & \multicolumn{1}{c|}{(0,0)} & \multicolumn{1}{c|}{0.151} & \multicolumn{1}{c|}{0.236} & \multicolumn{1}{c|}{0.085} & 0.000 *** & \multicolumn{1}{l|}{AU6=0} & \multicolumn{1}{c|}{0.246} & \multicolumn{1}{c|}{0.410} & \multicolumn{1}{c|}{0.164} & 0.000 *** \\
 & \multicolumn{1}{c|}{(1,0)} & \multicolumn{1}{c|}{0.431} & \multicolumn{1}{c|}{0.518} & \multicolumn{1}{c|}{0.087} & 0.024 * & \multicolumn{1}{l|}{AU6=1} & \multicolumn{1}{c|}{0.822} & \multicolumn{1}{c|}{0.908} & \multicolumn{1}{c|}{0.086} & 0.000 *** \\
 & \multicolumn{1}{c|}{(0,1)} & \multicolumn{1}{c|}{0.811} & \multicolumn{1}{c|}{0.873} & \multicolumn{1}{c|}{0.062} & 0.000 *** & \multicolumn{1}{l|}{AU12=0} & \multicolumn{1}{c|}{0.162} & \multicolumn{1}{c|}{0.241} & \multicolumn{1}{c|}{0.079} & 0.000 *** \\
 & \multicolumn{1}{c|}{(1,1)} & \multicolumn{1}{c|}{0.907} & \multicolumn{1}{c|}{0.934} & \multicolumn{1}{c|}{0.027} & 0.000 *** & \multicolumn{1}{l|}{AU12=1} & \multicolumn{1}{c|}{0.861} & \multicolumn{1}{c|}{0.899} & \multicolumn{1}{c|}{0.038} & 0.000 *** \\
\cline{1-11}
\end{tabular*}
}
\begin{tablenotes}
\item[1] Signif. codes:  0 ‘***’ 0.001 ‘**’ 0.01 ‘*’ 0.05 ‘.’ 0.1 ‘ ’ 1
\end{tablenotes}
\end{threeparttable}
\end{center}
\vspace{-5pt}
\caption{Proportion of "happy" labels among males and females conditioned on AU6 and AU12 for each of the popular expression datasets. Here $Y \in \{0,1\}$ is the "happy" label, $Z \in \{M,F\}$ is the gender attribute. Blanks and omitted p-values indicate that the (AU6, AU12) combinations do not contain enough data for the chi-square tests.}
\label{tab:Anno_bias_summary}
\vspace{-5pt}
\end{table*}

Table \ref{tab:Anno_bias_summary} shows the proportion of "happy" labels among males and females conditioned on different values of AU6 and AU12. For each conditional distribution of "happy," a chi-square test of independence is used to determine whether there is a significant relationship between the labels and gender after controlling for the AUs. Due to the limited sizes of KDEF and CFD, some (AU6, AU12) combinations do not contain enough data for the chi-square tests and thus a single AU (\ie, AU6 only or AU12 only) is used as a condition. 
It is important to note that even though OpenFace is not perfectly accurate, we have demonstrated that it does not contain systematic bias with respect to gender (\ie, its errors are random), and thus any systematic bias in emotion annotations conditioned on the AUs must be due to the bias in emotion annotations, not the AUs.

From Table \ref{tab:Anno_bias_summary}, we can see significant differences between lab-controlled datasets and in-the-wild datasets. For both KDEF and CFD, the distribution of "happy" labels is independent of gender when AU6 and AU12 are controlled. On the other hand, for ExpW, RAF-DB, and AffectNet, the proportions of "happy" labels are significantly higher for females than males even when the AUs have been controlled. We believe that the significantly less annotation bias in lab-controlled datasets can be explained by the fact that those images are often carefully vetted by experts before being released while "in-the-wild" datasets are often annotated by laymen who are not specifically trained to overcome their cognitive bias or unconscious stereotyping.
Comparing AffectNet-Manual and AffectNet-Automatic, we see that 
the levels of annotation bias are similar, indicating that the model used to automatically label the 540K images inherits the label bias in the manually-labeled dataset.

Figure \ref{fig:Anno_bias_scatterplot} shows the proportion of "happy" labels as a function of AU6 and AU12 intensity for each in-the-wild dataset (the size of lab-controlled data is too small to calculate average proportions). As expected, the proportion of "happy" labels is higher when AU6 and AU12 intensities are higher, but the effect is different between males and females. ExpW, AffectNet-Manual, and AffectNet-Automatic all show large discrepancies in the conditional distributions of "happy" labels between males and females while the difference for RAF-DB is smaller. In fact, a logistic regression would show that gender is a significant predictor even when AU6 and AU12 are controlled for all four datasets. This is consistent with the result in Table \ref{tab:Anno_bias_summary}. 

For the anger annotations, we also observe a consistent pattern of systematic annotation bias among all in-the-wild datasets, whereas lab-controlled datasets show no signs of annotation bias. For all in-the-wild datasets, males are more likely than females to be labeled as ``angry'' after the AUs are controlled; see the Supplementary Material for the results. We also check other expression annotations but do not find significant annotation biases between males and females as those observed with "happy" and "angry" annotations. This is partially because many expression classes' occurrence rates are too low in these datasets. For example, surprise, fear, and disgust account for about only 4\%, 1\%, and 1\% of all images in AffectNet-Manual respectively and thus the differences between males and females are minor.

\revision{We also conduct analysis on the ``happy'' annotations across different age and racial groups following a similar procedure. We find that younger people are more likely to be annotated as ``happy'' compared to older people in general, although the saliency of such annotation bias varies across datasets. We do not find evidence of systematic annotation bias across different racial groups. The full results can be found in the Supplementary Material. For both age and race analyses, further analysis is needed on more balanced datasets (\ie, datasets that have more older people and minority races).}

To explain the seemingly contradictory observations that "happy" and "angry" expression labels suffer from significant annotation bias while many AU labels do not, we believe this is because facial action units are local attributes and so the gender information has little impact on the annotators' annotation, whereas when the annotators conduct expression annotation, they tend to look at the faces holistically, and so the gender of the face influences their annotation in a non-negligible way.

\section{Bias Correction}

\subsection{Learned Bias in Trained Models}
Having observed the existence of annotation bias in in-the-wild expression datasets, we hypothesize that a naive model trained on these data will learn such bias and that removing the annotation bias will reduce the bias of the model. To test our hypothesis, we use ExpW as our training data and CFD as our test data. We select CFD because the images are lab-controlled and thus contain fewer confounding factors (such as differences in the backgrounds) when we evaluate the model predictions between males and females. Following convention \cite{kamiran2009classifying,kamishima2011fairness,zliobaite2015relation}, we use the Calders-Verwer (CV) discrimination score \cite{calders2010three} as our metric for the bias of the trained model: 
\begin{equation}
Disc = P(\hat{Y} = Happy|F) - P(\hat{Y} = Happy|M)
\end{equation}

Since the probabilities are no longer conditioned on the AUs, we will need to balance the test data (CFD) so that the proportions of true happy faces are the same between males and females.

\begin{table}
\small
\begin{center}
\begin{tabular}{|p{0.12\linewidth}|p{0.19\linewidth}|p{0.19\linewidth}|p{0.1\linewidth}|} 
\hline
\multicolumn{1}{|c|}{Training Data} & \multicolumn{1}{>{\centering\arraybackslash}p{0.19\linewidth}|}{P($\hat{Y}$=1 \textbar{}F)} & \multicolumn{1}{>{\centering\arraybackslash}p{0.19\linewidth}|}{P($\hat{Y}$=1 \textbar{}M)} & \multicolumn{1}{>{\centering\arraybackslash}p{0.1\linewidth}|}{$Disc$} \\
\hline
\multicolumn{1}{|c|}{Raw ExpW}      & \multicolumn{1}{c|}{0.3916}                   & \multicolumn{1}{c|}{0.3342}                   & \multicolumn{1}{c|}{0.0574}                                                 \\
\multicolumn{1}{|c|}{Relabeled ExpW}     & \multicolumn{1}{c|}{0.3655}                   & \multicolumn{1}{c|}{0.3603}                   & \multicolumn{1}{c|}{0.0052}                                             \\
\hline
\end{tabular}
\end{center}
\vspace{-5pt}
\caption{Proportions of "happy" classification among males and females on the CFD test set by a ResNet-50 model trained on ExpW and relabeled ExpW data. For ground truth labels, the proportion of ``happy'' in the test set is 0.3629 for both males and females.}
\label{tab:relabeling_result}
\vspace{-9pt}
\end{table}

We first train a naive happiness classifier using the raw ExpW dataset. We use ResNet-50 \cite{he2016deep} pre-trained on ImageNet and fine-tuned by Adam optimization \cite{kingma2014adam} with a learning rate of 0.0001 in PyTorch. To evaluate the effect of annotation bias, we relabel the ExpW data as follows: For each (AU6, AU12) presence combination, we first calculate the average proportion of images that are labeled as "happy." As females are more likely to be labeled as "happy" than males, we randomly sample some "unhappy" male images and relabel them as "happy" and sample some "happy" female images and relabel them as "unhappy" so that they both have the same proportion of images labeled as "happy" conditioned on the AUs. Even though this potentially introduces label errors, these modified labels are statistically fair, or, in other words, systematically unbiased. We then train a happiness classifier on a balanced subset (where each AU and gender combination has 3,000 samples) of the modified data using the same procedure.

For the test set, we remove a few easy happy and unhappy faces from CFD (whose predicted scores from the naive classifier $>$0.99999 or $<$0.00001) and then balance the proportions of happiness between males and females by removing some happy female images. As ExpW and CFD use different labeling criteria, the thresholds for binarizing the output of the trained classifier are adjusted to maximize the accuracy on the test set. Table \ref{tab:relabeling_result} shows the model bias observed on the test set. We see significant bias in the prediction for the model trained on raw ExpW while there is little bias for the model trained on the relabeled ExpW data. This shows that annotation bias can have a significant impact on model fairness and thus should be actively managed.

\subsection{Bias Correction}

Since data massaging techniques such as changing the labels are intrusive and undesirable (it may have legal implications because it is a form of training on falsified data~\cite{jiang2020identifying}), in this section, we propose an AU-Calibrated Facial Expression Recognition (AUC-FER) framework that can effectively achieve similar results without the need to modify the labels.

Our goal is to ensure that the model classifies expressions based on the AUs and not the gender, so we want to encourage the model to treat two samples in a similar way if their AUs are similar, even if their genders are different and the labels are different. We note that this is related to the concept of individual fairness (as opposed to group fairness). Our method is motivated by techniques in metric learning, which aims to learn an embedding space where the embedded vectors of similar samples are encouraged to be closer, while dissimilar ones are kept far from each other \cite{sohn2016improved,wang2019multi}. In particular, we use the triplet loss function \cite{schroff2015facenet} as a regularizer to penalize unfairness.

From the training data, we construct triplets $\{X_i, X_j, X_k\}$ within each batch where $X_i$ and $X_j$ are images with the same AU presence (\eg, (AU6, AU12) for happiness), and $X_k$ is an image with a different AU presence from $X_i$. The triplet loss is then defined as:
\begin{equation}
    \mathcal{L}_{trp} = \sum_{i,j,k}^{N_{trp}} [||f(X_i)-f(X_j)||_2^2-||f(X_i)-f(X_k)||_2^2 + \alpha]_+,
\end{equation}
where $[z]_+ = \max(z, 0)$, and $f(.)$ is the feature representation of the images. The goal of the triplet loss function is to make the distance between $X_i$ and $X_j$ in the embedding space larger than the distance between $X_i$ and $X_k$ by at least a minimum margin $\alpha$.

As usual, we have cross-entropy loss for classification:
\begin{equation}
    \mathcal{L}_{softmax} = -\frac{1}{N}\sum_{i=1}^{N} \mathds{1}[\hat{y_i}=y_i] log(p(y_i)).
\end{equation}

The total loss function is then defined as the weighted sum of $L_{softmax}$ and $L_{trp}$:
\begin{equation}
    \mathcal{L} = \mathcal{L}_{softmax} + \lambda \mathcal{L}_{trp},
\end{equation}
where $\lambda$ measures how willing we are to deviate from the given biased labels and enforce fairness.

\subsection{Experiments}
We evaluate the proposed AUC-FER method by comparing it with other debiasing methods in the fairness literature. Popular methods include uniform confusion \cite{alvi2018turning}, gradient projection \cite{zhang2018mitigating}, domain discriminative training \cite{wang2020towards}, and domain independent training \cite{wang2020towards}. Many are motivated by techniques in domain adaptation and are designed to reduce data composition bias. We compare them with AUC-FER to evaluate their effectiveness in mitigating annotation bias.

For the first set of experiments, we use the ResNet-50 architecture \cite{he2016deep} pre-trained on ImageNet in PyTorch. For the four benchmark models, we follow 
Wang \etal \cite{wang2020towards} and replace the FC layer of the ResNet model with two consecutive FC layers both of size 2,048 with Dropout and ReLU in between. For AUC-FER, we use the PyTorch Metric Learning library \cite{musgrave2020pytorch} for the triplet loss implementation. All models are trained on random subsets of ExpW of size 20,000 and tested on the previously constructed CFD test set. The thresholds for binarizing the output scores are again chosen to maximize the accuracy on the test set, and the experiment is repeated 5 times for each model. To test the robustness of AUC-FER with respect to the model architecture and the size of training data, we repeat this experiment using MobileNetV2 \cite{sandler2018mobilenetv2} and a training set of size 8,000.

Tables \ref{tab:debiasing_result} and \ref{tab:othernetwork} show the discrimination scores for the models and compare them with baseline ResNet-50 and MobileNetV2 models. AUC-FER obtains the lowest discrimination score, which is a 64-89\% reduction in bias compared to the baseline models and is very close to the result we get by relabeling the training data. This shows that the proposed AUC-FER framework is effective in removing annotation bias. We also perform experiments for the angry expression using AffectNet-Automatic as training data, and AUC-FER again outperforms other debiasing techniques. 
The experiment details and analysis for the anger expression are included in the Supplementary Material. 

\begin{table}
\small
\begin{center}
\scalebox{0.95}{
\begin{tabular}{|p{0.26\linewidth}|p{0.24\linewidth}|p{0.22\linewidth}|} 
\hline
\multirow{1}{\linewidth}{\parbox{1.7\linewidth}{\vspace{0.09cm} \centering Methods\\ (ResNet-50~\cite{he2016deep})}}  & \multirow{1}{\linewidth}{\parbox{1\linewidth}{\vspace{0.25cm} \centering $Disc$}}        & \multicolumn{1}{>{\centering\arraybackslash}p{0.22\linewidth}|}{Compared to Baseline (\%)} \\ 
\hhline{|===|}
\multicolumn{1}{|c|}{Baseline}    & \multicolumn{1}{c|}{0.059 $\pm$ 0.035}     & \multicolumn{1}{c|}{-}         \\
\multicolumn{1}{|c|}{Uniform Confusion~\cite{alvi2018turning}}    & \multicolumn{1}{c|}{0.046 $\pm$ 0.008}     & \multicolumn{1}{c|}{77.6}         \\
\multicolumn{1}{|c|}{Gradient Projection~\cite{zhang2018mitigating}}  & \multicolumn{1}{c|}{0.036 $\pm$ 0.014}    & \multicolumn{1}{c|}{60.0}          \\
\multicolumn{1}{|c|}{Domain Discriminative~\cite{wang2020towards}}   & \multicolumn{1}{c|}{0.076 $\pm$ 0.024}  & \multicolumn{1}{c|}{128.8}        \\
\multicolumn{1}{|c|}{Domain Independent~\cite{wang2020towards}}   & \multicolumn{1}{c|}{0.029 $\pm$ 0.015}   & \multicolumn{1}{c|}{49.4}            \\
\multicolumn{1}{|c|}{AUC-FER (Ours)}   & \multicolumn{1}{c|}{\textbf{0.006 $\pm$ 0.020}}  & \multicolumn{1}{c|}{\textbf{10.6}} \\
\hline
\end{tabular}
}
\end{center}
\vspace{-5pt}
\caption{Discrimination scores for various debiasing methods using the ResNet-50 architecture trained on random subsets of ExpW of size 20,000 and tested on CFD for the ``happy'' expression. The average discrimination scores are compared against the baseline model and shown as a percentage.}
\label{tab:debiasing_result}
\vspace{-5pt}
\end{table}

\begin{table}
\small
\begin{center}
\scalebox{0.95}{
\begin{tabular}{|p{0.26\linewidth}|p{0.24\linewidth}|p{0.22\linewidth}|} 
\hline
\multirow{1}{\linewidth}{\parbox{1.7\linewidth}{\vspace{0.09cm} \centering Methods (MobileNetV2~\cite{sandler2018mobilenetv2})}}  & \multirow{1}{\linewidth}{\parbox{1\linewidth}{\vspace{0.25cm} \centering $Disc$}}        & \multicolumn{1}{>{\centering\arraybackslash}p{0.22\linewidth}|}{Compared to Baseline (\%)} \\ 
\hhline{|===|}
\multicolumn{1}{|c|}{Baseline}    & \multicolumn{1}{c|}{0.079 $\pm$ 0.009}     & \multicolumn{1}{c|}{-}         \\
\multicolumn{1}{|c|}{Uniform Confusion~\cite{alvi2018turning}}    & \multicolumn{1}{c|}{0.085 $\pm$ 0.021}     & \multicolumn{1}{c|}{107.8}         \\
\multicolumn{1}{|c|}{Gradient Projection~\cite{zhang2018mitigating}}  & \multicolumn{1}{c|}{0.070 $\pm$ 0.036}    & \multicolumn{1}{c|}{88.2}          \\
\multicolumn{1}{|c|}{Domain Discriminative~\cite{wang2020towards}}   & \multicolumn{1}{c|}{0.064 $\pm$ 0.029}  & \multicolumn{1}{c|}{80.4}        \\
\multicolumn{1}{|c|}{Domain Independent~\cite{wang2020towards}}   & \multicolumn{1}{c|}{0.062 $\pm$ 0.035}   & \multicolumn{1}{c|}{78.4}            \\
\multicolumn{1}{|c|}{AUC-FER (Ours)}   & \multicolumn{1}{c|}{\textbf{0.028 $\pm$ 0.029}}  & \multicolumn{1}{c|}{\textbf{35.9}} \\
\hline
\end{tabular}
}
\end{center}
\vspace{-5pt}
\caption{Discrimination scores for debiasing methods using the MobileNetV2 architecture trained on random subsets of ExpW of size 8,000 and tested on CFD for the ``happy'' expression.}
\label{tab:othernetwork}
\vspace{-10pt}
\end{table}

\section{Discussion}
In this paper, we study systematic biases in human annotations in public datasets on facial expressions. To our knowledge, this is the first work in computer vision to demonstrate the systematic effect of annotators’ perceptual bias as a potential source of bias that can be injected into computer vision models. 
We show that, contrary to the common assumption that annotation errors are just random noises, systematic biases exist in many facial expression datasets. The problem is more severe for in-the-wild datasets than lab-controlled datasets. We illustrate that if these biases are not addressed, trained models will also be biased.
We further develop an 
AUC-FER framework to address annotation bias for expression recognition tasks 
and demonstrate that it is more effective in reducing annotation bias than existing debiasing methods. 

The presented framework for facial expression recognition utilizes AUs as an auxiliary variable to enforce fairness since they are specifically designed to resolve subjectivity in facial analysis. 
This framework can be extended beyond expression recognition. In general, one can use any objective measures (\eg body keypoints) for tasks requiring subjective human labeling (\eg activity recognition or image captioning) within the proposed framework. Although such objective measures may not always be accurate in practice (\eg, 
applying OpenFace introduces additional noises), the belief is that because these measures (AUs, body keypoints) are often local attributes and less affected by other attributes of the subjects (\eg, gender, race, or age), they are fairer than the subjective labels in the training data and can thus be used as calibration for fairness.

\revision{For future work, we believe that combining our method with other debiasing techniques may potentially be effective when the training data suffers from multiple sources of biases (both composition bias and annotation bias).}

\revision{This paper focuses on the identification and mitigation of systematic annotation bias. It would be interesting for dataset curators to study if such annotation bias varies across annotator subgroups. Recent work has also pointed out that the prototypical framework of six expressions does not capture the full facial expressions of humans~\cite{du2014compound}, and compound emotions have been proposed to address the genuine ambivalence on some displayed facial expressions~\cite{du2014compound,emotionet,li2017reliable,blank2020emotional}. Future work can study the role of these definitions and their interaction with bias.}


\noindent
\textbf{Acknowledgement}
This work was supported by NSF SBE/SMA \#1831848 ``RIDIR: Integrated Communication Database and Computational Tools.''

\fontsize{9.7pt}{10.7pt} \selectfont
{\small
\bibliographystyle{ieee_fullname}
\bibliography{bibliography}
}

\end{document}


\title{Supplementary Material: Understanding and Mitigating Annotation Bias \\  in Facial Expression Recognition}

\author{Yunliang Chen \qquad Jungseock Joo\\
University of California, Los Angeles\\
{\tt\small chenyunliang@ucla.edu, jjoo@comm.ucla.edu}
}

\maketitle

\section{Angry Annotation Bias Between Genders}
\restylefloat{table}

\subsection{OpenFace Accuracy for Angry AUs}
We evaluate annotation bias for the angry expression. Recall that anger is defined as the combination of AU4 (brow lowerer), AU5 (upper lid raiser), AU7 (lid tightener), and AU23 (lip tightener). In this section, we check the quality of OpenFace AU recognition by comparing its output with the expert-coded EmotioNet dataset that consists of 24,600 images \cite{emotionet}. Since EmotioNet does not have labels for AU7 and AU23, we are only able to check for AU4 and AU5. 

Table \ref{tab:AU_accuracy_angry} shows the OpenFace AU accuracy results for AU4 and AU5. Similar to the case of happiness expression, we binarize the AU intensity outputs of OpenFace using a threshold that is chosen to maximize the overall accuracy of prediction. The accuracy result is shown as "Raw" in Table \ref{tab:AU_accuracy_angry}. The difference between males and females is small for AU5, while it is not negligible for AU4. Thus, we re-calibrate the OpenFace output so that different binarization thresholds are chosen to balance the accuracy between males and females. We remark that learning subgroup-specific thresholds is a common technique used to achieve fairness \cite{robinson2020face}. As shown in the "Recalibrated" column of Table \ref{tab:AU_accuracy_angry}, there is no statistically significant difference between males and females after calibration (\ie, the AU's are fair). Since there is little difference in AU5 before and after calibration, we will only apply calibration to AU4 in the below evaluations.

\begin{table}[h]
\small
\centering
\begin{tabular}{|l||cc|cc|} 
\hline
      & \multicolumn{2}{c|}{AU4 Accuracy} & \multicolumn{2}{c|}{AU5 Accuracy}  \\ 
\hline
  & Raw & Recalibrated      & Raw & Recalibrated      \\ 
\hline\hline
Male    & 0.854    & 0.857         & 0.968    & 0.958         \\
Female  & 0.928    & 0.855         & 0.958    & 0.958          \\ 
\hline
p-value & 0.000    &    0.779         & 0.00002    &    0.880          \\
\hline
\end{tabular}
\caption{Accuracy of OpenFace AU Recognition, evaluated on 24,600 EmotioNet images with expert-coded AUs. For the raw accuracy, the AU intensity output of OpenFace are binarized using a single threshold that maximizes the overall accuracy, while for the re-calibrated accuracy, different binarization thresholds are chosen to balance the accuracy between males and females.}
\label{tab:AU_accuracy_angry}
\end{table}

\subsection{Annotation Bias of the Angry Expression}

\label{tab:Anno_bias_angry_summary}

\begin{table*}[h]
\small
\centering
\begin{threeparttable}
\scalebox{0.95}{
\begin{tabular*}{1\textwidth}{|m{0.1\linewidth}||m{0.03\linewidth}|m{0.04\linewidth}|m{0.04\linewidth}|m{0.04\linewidth}|m{0.08\linewidth}||m{0.045\linewidth}|m{0.04\linewidth}|m{0.04\linewidth}|m{0.045\linewidth}|m{0.08\linewidth}|}
\cline{1-11}
\multirow{2}{\linewidth}{\parbox{1\linewidth}{\vspace{0.1cm} \centering Data \\ (Collecting Condition, Size)}} & \multicolumn{5}{c||}{Conditioned on Marginal AU} & \multicolumn{5}{c|}{Conditioned on Marginal AU} \\ 
\cline{2-11}
 & \multicolumn{1}{>{\centering\arraybackslash}m{0.06\linewidth}|}{AU} & \multicolumn{1}{>{\centering\arraybackslash}m{0.065\linewidth}|}{P(Angry \textbar{}AU, M)} & \multicolumn{1}{>{\centering\arraybackslash}m{0.06\linewidth}|}{P(Angry \textbar{}AU, F)} & \centering $\Delta$ & \multicolumn{1}{>{\centering\arraybackslash}m{0.08\linewidth}||}{p-value of $\chi^2$ test for $Y \perp \!\!\! \perp  Z$} & \multicolumn{1}{>{\centering\arraybackslash}m{0.06\linewidth}|}{AU} & \multicolumn{1}{>{\centering\arraybackslash}m{0.065\linewidth}|}{P(Angry \textbar{}AU, M)} & \multicolumn{1}{>{\centering\arraybackslash}m{0.06\linewidth}|}{P(Angry \textbar{}AU, F)} & \centering $\Delta$ & \multicolumn{1}{>{\centering\arraybackslash}m{0.08\linewidth}|}{p-value of $\chi^2$ test for $Y \perp \!\!\! \perp  Z$} \\ 
\hhline{|======::=====|} 
\multirow{4}{\linewidth}{\centering KDEF (Lab, 980) \cite{lundqvist1998karolinska}} & \multicolumn{1}{l|}{AU4=0} & \multicolumn{1}{c|}{0.090} & \multicolumn{1}{c|}{0.060} & \multicolumn{1}{c|}{0.030} & 0.103 & \multicolumn{1}{l|}{AU7=0} & \multicolumn{1}{c|}{0.124} & \multicolumn{1}{c|}{0.089} & \multicolumn{1}{c|}{0.035} & 0.246 \\
 & \multicolumn{1}{l|}{AU4=1} & \multicolumn{1}{c|}{0.405} & \multicolumn{1}{c|}{0.465} & \multicolumn{1}{c|}{-0.061} & 0.408 & \multicolumn{1}{l|}{AU7=1} & \multicolumn{1}{c|}{0.162} & \multicolumn{1}{c|}{0.178} & \multicolumn{1}{c|}{-0.016} & 0.609 \\
 & \multicolumn{1}{l|}{AU5=0} & \multicolumn{1}{c|}{0.207} & \multicolumn{1}{c|}{0.173} & \multicolumn{1}{c|}{0.034} & 0.320 & \multicolumn{1}{l|}{AU23=0} & \multicolumn{1}{c|}{0.123} & \multicolumn{1}{c|}{0.133} & \multicolumn{1}{c|}{-0.009} & 0.684 \\
 & \multicolumn{1}{l|}{AU5=1} & \multicolumn{1}{c|}{0.071} & \multicolumn{1}{c|}{0.104} & \multicolumn{1}{c|}{-0.032} & 0.231 & \multicolumn{1}{l|}{AU23=1} & \multicolumn{1}{c|}{0.268} & \multicolumn{1}{c|}{0.220} & \multicolumn{1}{c|}{0.047} & 0.533 \\ 
\cline{1-11}
\multirow{4}{\linewidth}{\centering CFD \\ (Lab, 1,207) \cite{ma2015chicago}} & \multicolumn{1}{l|}{AU4=0} & \multicolumn{1}{c|}{0.075} & \multicolumn{1}{c|}{0.091} & \multicolumn{1}{c|}{-0.016} & 0.324 & \multicolumn{1}{l|}{AU7=0} & \multicolumn{1}{c|}{0.026} & \multicolumn{1}{c|}{0.043} & \multicolumn{1}{c|}{-0.017} & 0.228  \\
 & \multicolumn{1}{l|}{AU4=1} & \multicolumn{1}{c|}{1.000} & \multicolumn{1}{c|}{0.933} & \multicolumn{1}{c|}{0.067} & - & \multicolumn{1}{l|}{AU7=1} & \multicolumn{1}{c|}{0.245} & \multicolumn{1}{c|}{0.233} & \multicolumn{1}{c|}{0.012} & 0.743 \\
 & \multicolumn{1}{l|}{AU5=0} & \multicolumn{1}{c|}{0.132} & \multicolumn{1}{c|}{0.156} & \multicolumn{1}{c|}{-0.023} & 0.326 & \multicolumn{1}{l|}{AU23=0} & \multicolumn{1}{c|}{0.081} & \multicolumn{1}{c|}{0.074} & \multicolumn{1}{c|}{0.007} & 0.711 \\
 & \multicolumn{1}{l|}{AU5=1} & \multicolumn{1}{c|}{0.096} & \multicolumn{1}{c|}{0.072} & \multicolumn{1}{c|}{0.024} & 0.448 & \multicolumn{1}{l|}{AU23=1} & \multicolumn{1}{c|}{0.208} & \multicolumn{1}{c|}{0.237} & \multicolumn{1}{c|}{-0.028} & 0.490  \\ 
\cline{1-11}
\multirow{4}{\linewidth}{\centering ExpW (Web, 91,793) \cite{SOCIALRELATION_ICCV2015,SOCIALRELATION_2017}} & \multicolumn{1}{l|}{AU4=0} & \multicolumn{1}{c|}{0.041} & \multicolumn{1}{c|}{0.033} & \multicolumn{1}{c|}{0.008} & 0.000 *** & \multicolumn{1}{l|}{AU7=0} & \multicolumn{1}{c|}{0.034} & \multicolumn{1}{c|}{0.026} & \multicolumn{1}{c|}{0.008} & 0.000 ***  \\
 & \multicolumn{1}{l|}{AU4=1} & \multicolumn{1}{c|}{0.145} & \multicolumn{1}{c|}{0.126} & \multicolumn{1}{c|}{0.019} & 0.235 & \multicolumn{1}{l|}{AU7=1} & \multicolumn{1}{c|}{0.053} & \multicolumn{1}{c|}{0.044} & \multicolumn{1}{c|}{0.008} & 0.000 *** \\
 & \multicolumn{1}{l|}{AU5=0} & \multicolumn{1}{c|}{0.042} & \multicolumn{1}{c|}{0.035} & \multicolumn{1}{c|}{0.007} & 0.000 *** & \multicolumn{1}{l|}{AU23=0} & \multicolumn{1}{c|}{0.045} & \multicolumn{1}{c|}{0.036} & \multicolumn{1}{c|}{0.009} & 0.000 *** \\
 & \multicolumn{1}{l|}{AU5=1} & \multicolumn{1}{c|}{0.046} & \multicolumn{1}{c|}{0.035} & \multicolumn{1}{c|}{0.010} & 0.000 *** & \multicolumn{1}{l|}{AU23=1} & \multicolumn{1}{c|}{0.041} & \multicolumn{1}{c|}{0.033} & \multicolumn{1}{c|}{0.008} & 0.010 ** \\ 
\cline{1-11}
\multirow{4}{\linewidth}{\centering RAF-DB (Web, 15,339) \cite{li2017reliable}} & \multicolumn{1}{l|}{AU4=0} & \multicolumn{1}{c|}{0.089} & \multicolumn{1}{c|}{0.030} & \multicolumn{1}{c|}{0.058} & 0.000 *** & \multicolumn{1}{l|}{AU7=0} & \multicolumn{1}{c|}{0.052} & \multicolumn{1}{c|}{0.017} & \multicolumn{1}{c|}{0.036} & 0.000 *** \\
 & \multicolumn{1}{l|}{AU4=1} & \multicolumn{1}{c|}{0.295} & \multicolumn{1}{c|}{0.061} & \multicolumn{1}{c|}{0.234} & 0.000 *** & \multicolumn{1}{l|}{AU7=1} & \multicolumn{1}{c|}{0.127} & \multicolumn{1}{c|}{0.046} & \multicolumn{1}{c|}{0.081} & 0.000 *** \\
 & \multicolumn{1}{l|}{AU5=0} & \multicolumn{1}{c|}{0.108} & \multicolumn{1}{c|}{0.037} & \multicolumn{1}{c|}{0.071} & 0.000 *** & \multicolumn{1}{l|}{AU23=0} & \multicolumn{1}{c|}{0.089} & \multicolumn{1}{c|}{0.031} & \multicolumn{1}{c|}{.058} & 0.000 *** \\
 & \multicolumn{1}{l|}{AU5=1} & \multicolumn{1}{c|}{0.043} & \multicolumn{1}{c|}{0.013} & \multicolumn{1}{c|}{0.030} & 0.000 *** & \multicolumn{1}{l|}{AU23=1} & \multicolumn{1}{c|}{0.097} & \multicolumn{1}{c|}{0.032} & \multicolumn{1}{c|}{.066} & 0.000 *** \\ 
\cline{1-11}
\multirow{4}{\linewidth}{\centering AffectNet-Manual (Web, 420,299) \cite{mollahosseini2017affectNet}} & \multicolumn{1}{l|}{AU4=0} & \multicolumn{1}{c|}{0.083} & \multicolumn{1}{c|}{0.036} & \multicolumn{1}{c|}{0.047} & 0.000 *** & \multicolumn{1}{l|}{AU7=0} & \multicolumn{1}{c|}{0.088} & \multicolumn{1}{c|}{0.037} & \multicolumn{1}{c|}{0.051} & 0.000 *** \\
 & \multicolumn{1}{l|}{AU4=1} & \multicolumn{1}{c|}{0.287} & \multicolumn{1}{c|}{0.112} & \multicolumn{1}{c|}{0.175} & 0.000 *** & \multicolumn{1}{l|}{AU7=1} & \multicolumn{1}{c|}{0.093} & \multicolumn{1}{c|}{0.039} & \multicolumn{1}{c|}{0.054} & 0.000 *** \\
 & \multicolumn{1}{l|}{AU5=0} & \multicolumn{1}{c|}{0.093} & \multicolumn{1}{c|}{0.037} & \multicolumn{1}{c|}{0.056} & 0.000 *** & \multicolumn{1}{l|}{AU23=0} & \multicolumn{1}{c|}{0.092} & \multicolumn{1}{c|}{0.037} & \multicolumn{1}{c|}{0.054} & 0.000 *** \\
 & \multicolumn{1}{l|}{AU5=1} & \multicolumn{1}{c|}{0.084} & \multicolumn{1}{c|}{0.038} & \multicolumn{1}{c|}{0.046} & 0.000 *** & \multicolumn{1}{l|}{AU23=1} & \multicolumn{1}{c|}{0.086} & \multicolumn{1}{c|}{0.038} & \multicolumn{1}{c|}{0.048} & 0.000 *** \\ 
\cline{1-11}
\multirow{4}{\linewidth}{\centering AffectNet-Automatic (Web, 539,607) \cite{mollahosseini2017affectNet}} & \multicolumn{1}{l|}{AU4=0} & \multicolumn{1}{c|}{0.095} & \multicolumn{1}{c|}{0.019} & \multicolumn{1}{c|}{0.066} & 0.000 *** & \multicolumn{1}{l|}{AU7=0} & \multicolumn{1}{c|}{0.081} & \multicolumn{1}{c|}{0.017} & \multicolumn{1}{c|}{0.064} & 0.000 ***  \\
 & \multicolumn{1}{l|}{AU4=1} & \multicolumn{1}{c|}{0.374} & \multicolumn{1}{c|}{0.118} & \multicolumn{1}{c|}{0.256} & 0.000 *** & \multicolumn{1}{l|}{AU7=1} & \multicolumn{1}{c|}{0.108} & \multicolumn{1}{c|}{0.026} & \multicolumn{1}{c|}{0.083} & 0.000 *** \\
 & \multicolumn{1}{l|}{AU5=0} & \multicolumn{1}{c|}{0.103} & \multicolumn{1}{c|}{0.023} & \multicolumn{1}{c|}{0.080} & 0.000 *** & \multicolumn{1}{l|}{AU23=0} & \multicolumn{1}{c|}{0.096} & \multicolumn{1}{c|}{0.019} & \multicolumn{1}{c|}{0.077} & 0.000 *** \\
 & \multicolumn{1}{l|}{AU5=1} & \multicolumn{1}{c|}{0.071} & \multicolumn{1}{c|}{0.018} & \multicolumn{1}{c|}{0.053} & 0.000 *** & \multicolumn{1}{l|}{AU23=1} & \multicolumn{1}{c|}{0.085} & \multicolumn{1}{c|}{0.029} & \multicolumn{1}{c|}{0.056} & 0.000 ***  \\
\cline{1-11}
\end{tabular*}
}
\begin{tablenotes}
\item[] Signif. codes:  0 ‘***’ 0.001 ‘**’ 0.01 ‘*’ 0.05 ‘.’ 0.1 ‘ ’ 1
\end{tablenotes}
\end{threeparttable}
\caption{Proportion of "angry" labels among males and females conditioned on AU4, AU5, AU7, and AU23 for each of the popular expression datasets. Here $Y \in \{0,1\}$ is the "angry" label, $Z \in \{M,F\}$ is the gender attribute. Blanks and omitted p-values indicate that the AUs do not contain enough data for the chi-square tests.}
\label{tab:Anno_bias_angry_summary}
\end{table*}

In this section, we present evaluation results for the angry annotation bias in major public datasets. Similar to the evaluation of happy annotations, we apply the OpenFace AU detector and obtain the AU presence and AU intensity information for each image. AU4 intensities are then binarized into AU4 presence variable using the calibrated thresholds found earlier. All other AUs (AU5, AU7, and AU23) use the raw AU presence outputs from OpenFace since no adjustment is needed or available as shown in the previous section. We also apply our gender classifier when the gender information is not available (\ie, for ExpW and AffectNet). 

Table \ref{tab:Anno_bias_angry_summary} shows the proportion of "angry" labels among males and females conditioned on different values of AU4, AU5, AU7, and AU23. For each of the conditional distributions of "angry," a chi-square test of independence is used to determine whether there is a significant relationship between the labels and gender after controlling for the AUs. Unlike Table 
2 in the paper, only marginal distributions are shown since the joint distribution of (AU4, AU5, AU7, AU23) can take $2^4 = 16$ possible values and in many cases do not contain enough data or significantly reduce the power of statistical testing. This is expected since many AU's are correlated and so some combinations of (AU4, AU5, AU7, AU23) are much more likely than others. We believe that the distributions of "angry" labels conditioned on marginal AU still provides useful information for comparing the labels of lab-controlled datasets and in-the-wild datasets.

From Table \ref{tab:Anno_bias_angry_summary}, we can see significant differences between lab-controlled datasets and in-the-wild datasets. The pattern is similar to that of the happiness expression. For both KDEF and CFD, the distribution of "angry" labels is independent of gender when the AUs are controlled. On the other hand, for ExpW, RAF-DB, and AffectNet, the proportion of "angry" labels is significantly higher for males than females even when the AUs have been controlled. 

\begin{figure*}[t]
\centering
\includegraphics[width=0.24\textwidth]{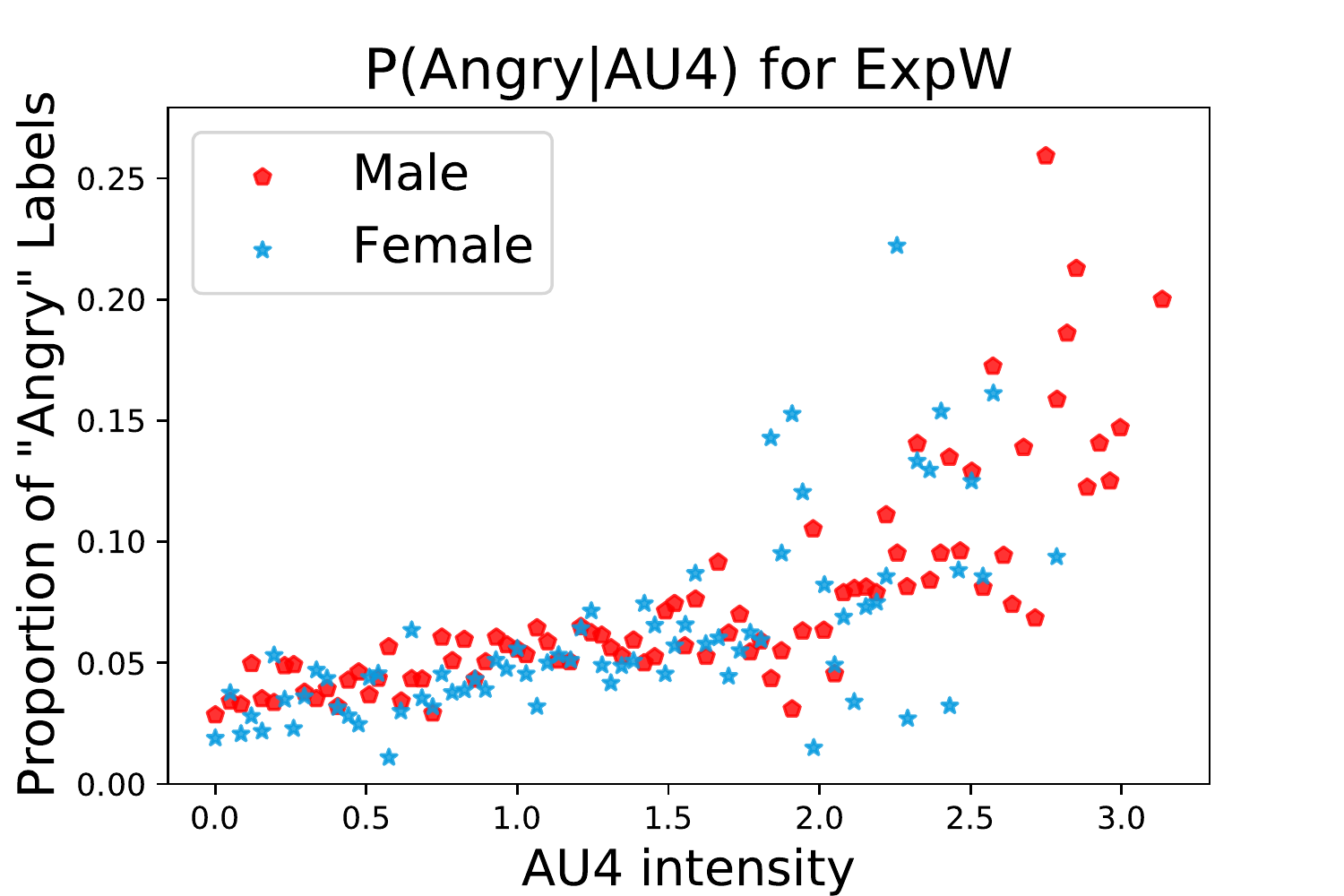}
\includegraphics[width=0.24\textwidth]{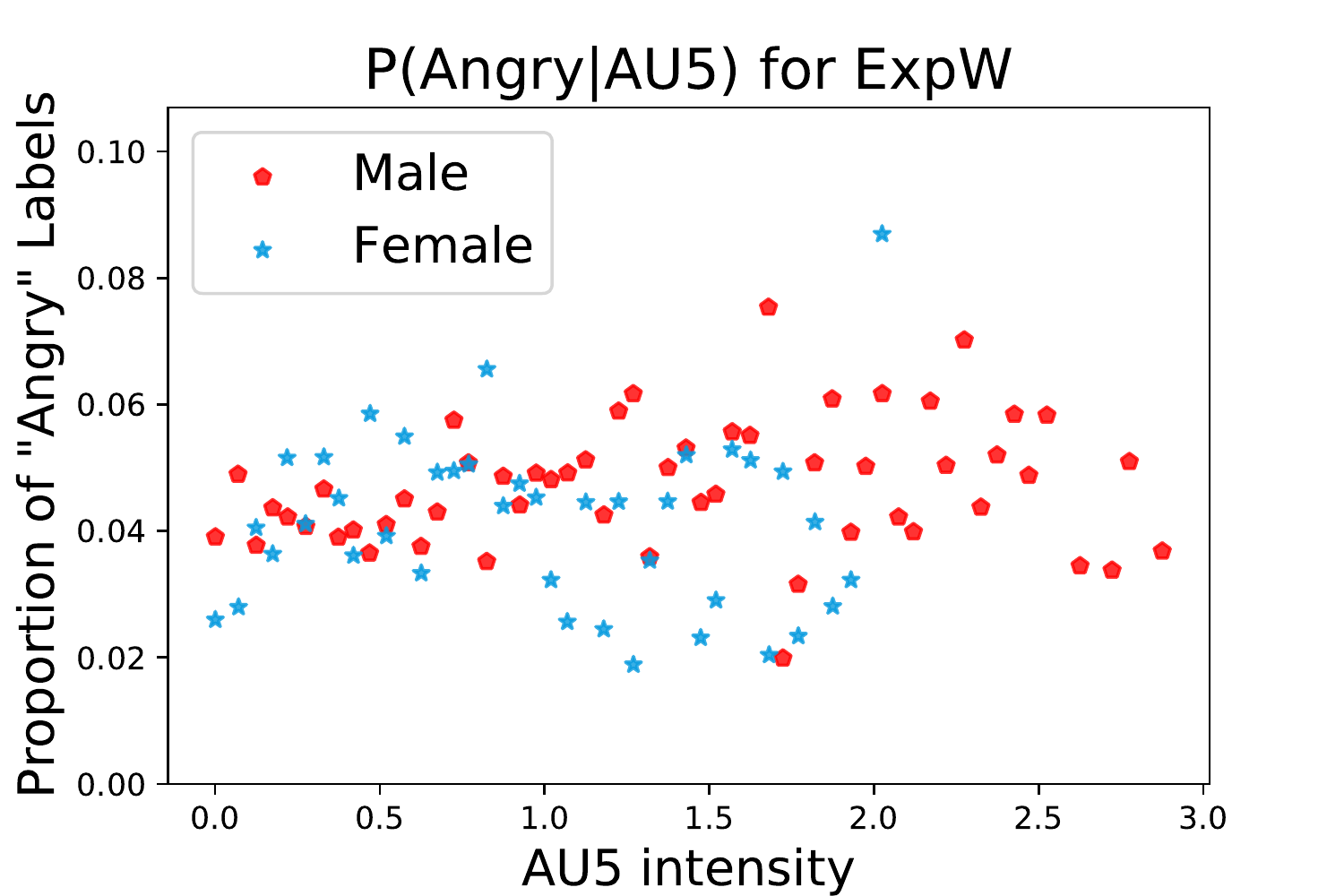}
\includegraphics[width=0.24\textwidth]{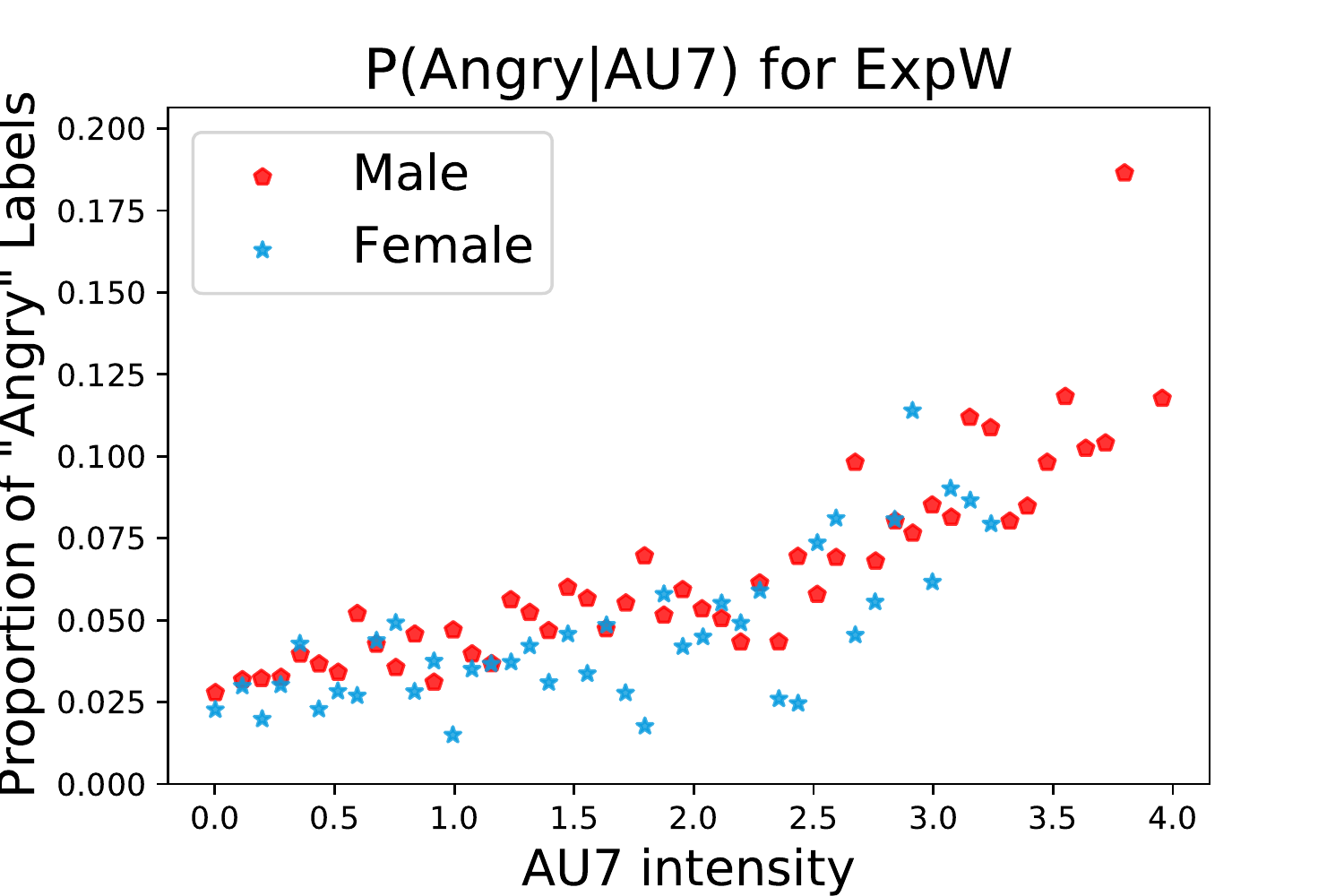}
\includegraphics[width=0.24\textwidth]{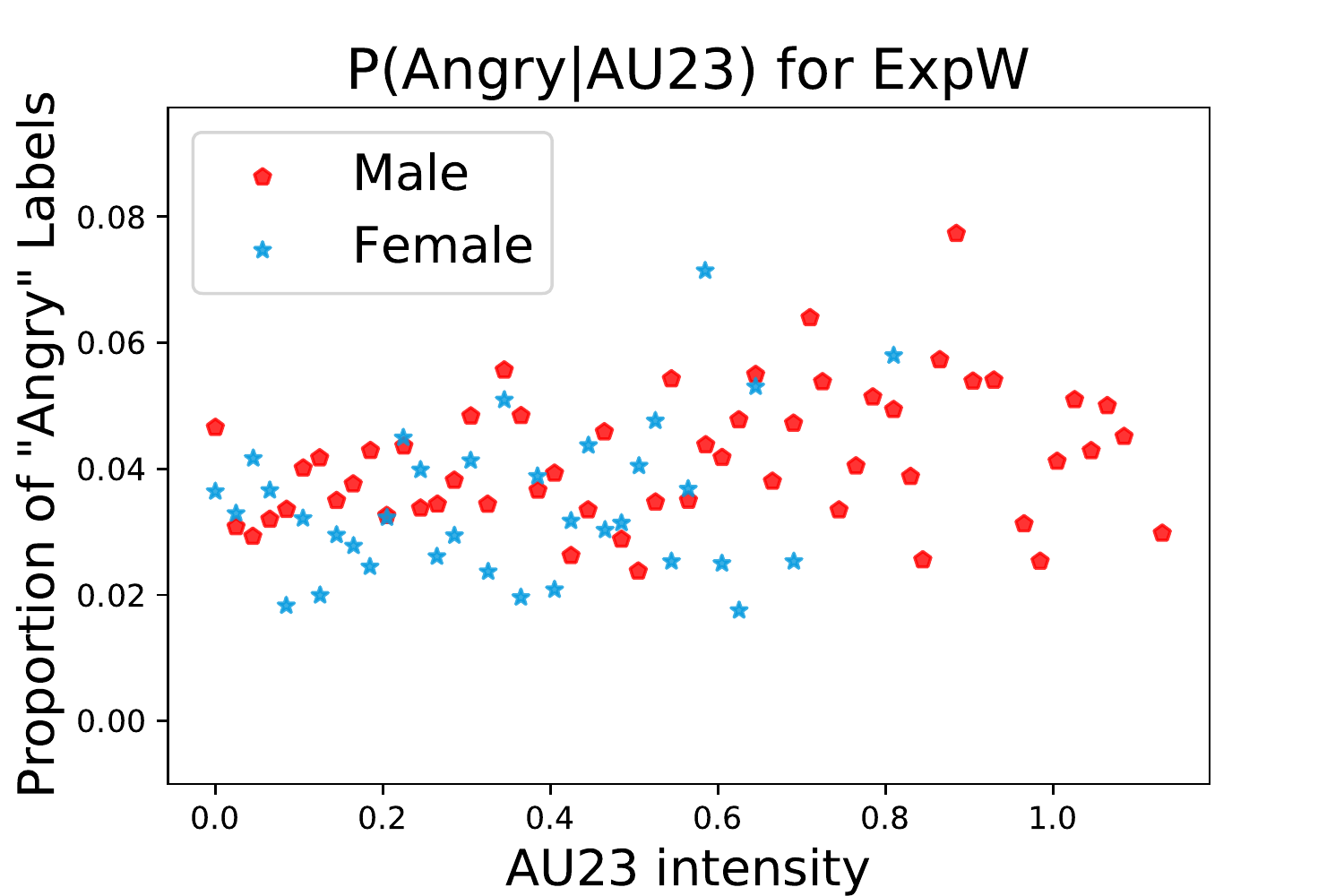} 
\includegraphics[width=0.24\textwidth]{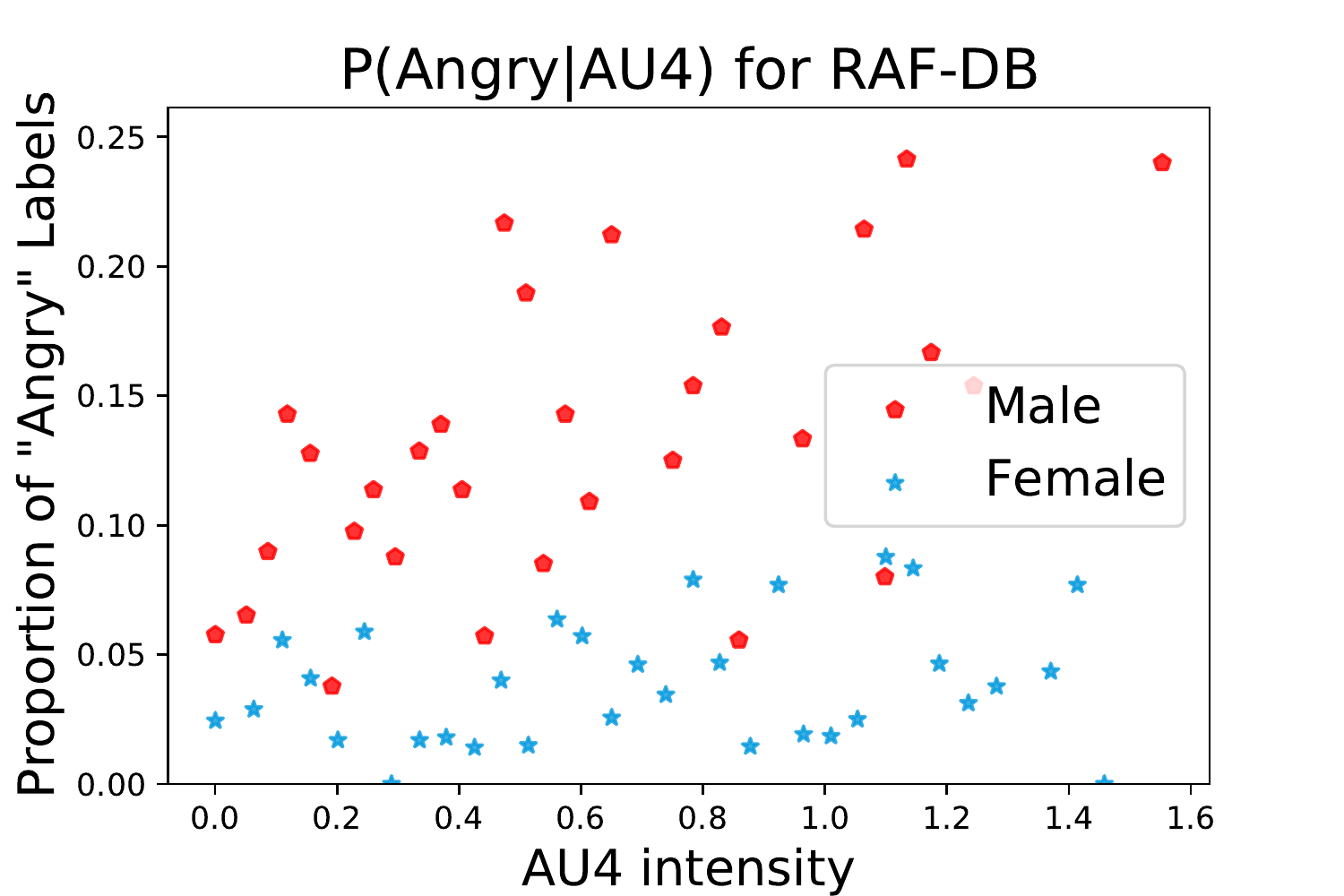}
\includegraphics[width=0.24\textwidth]{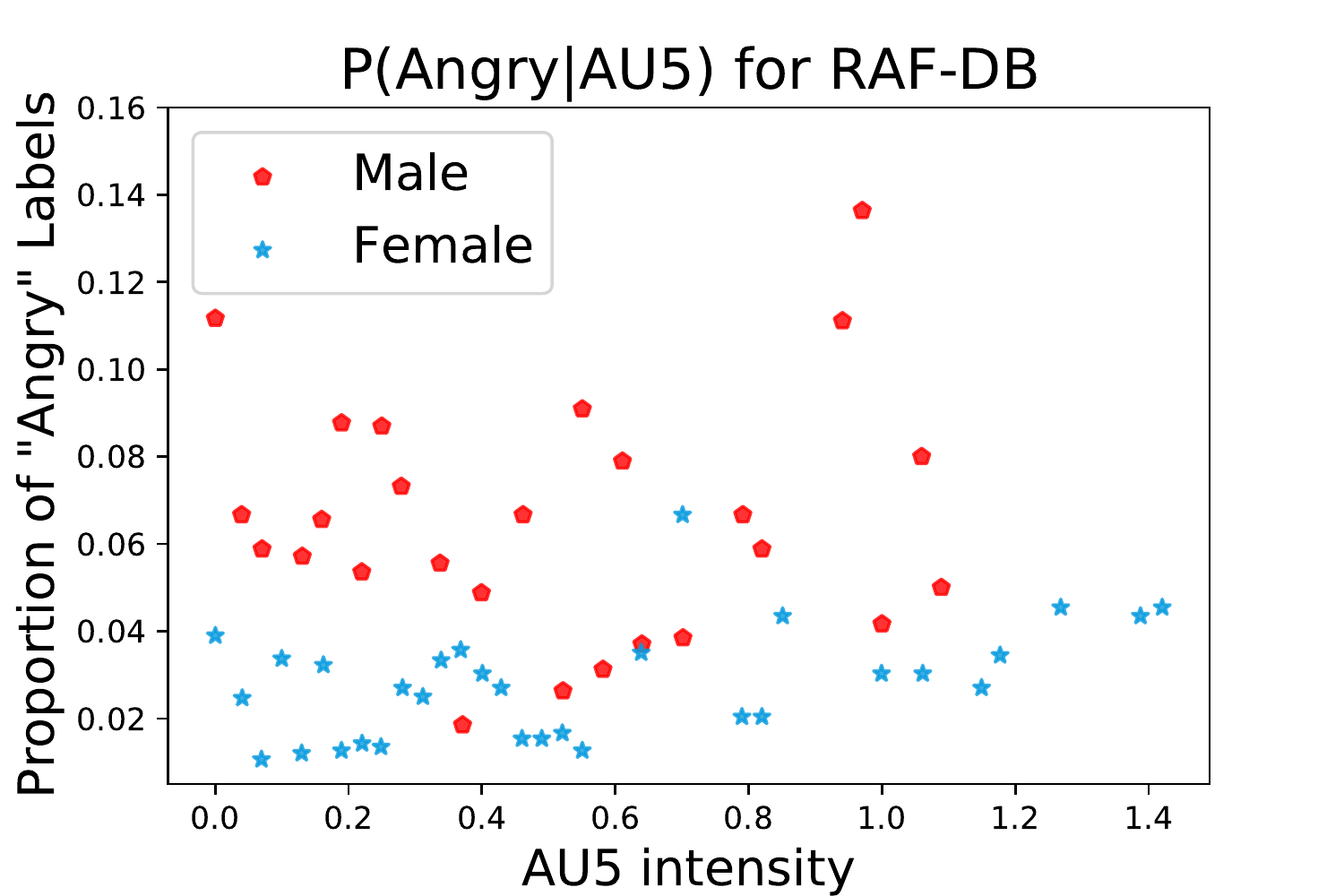}
\includegraphics[width=0.24\textwidth]{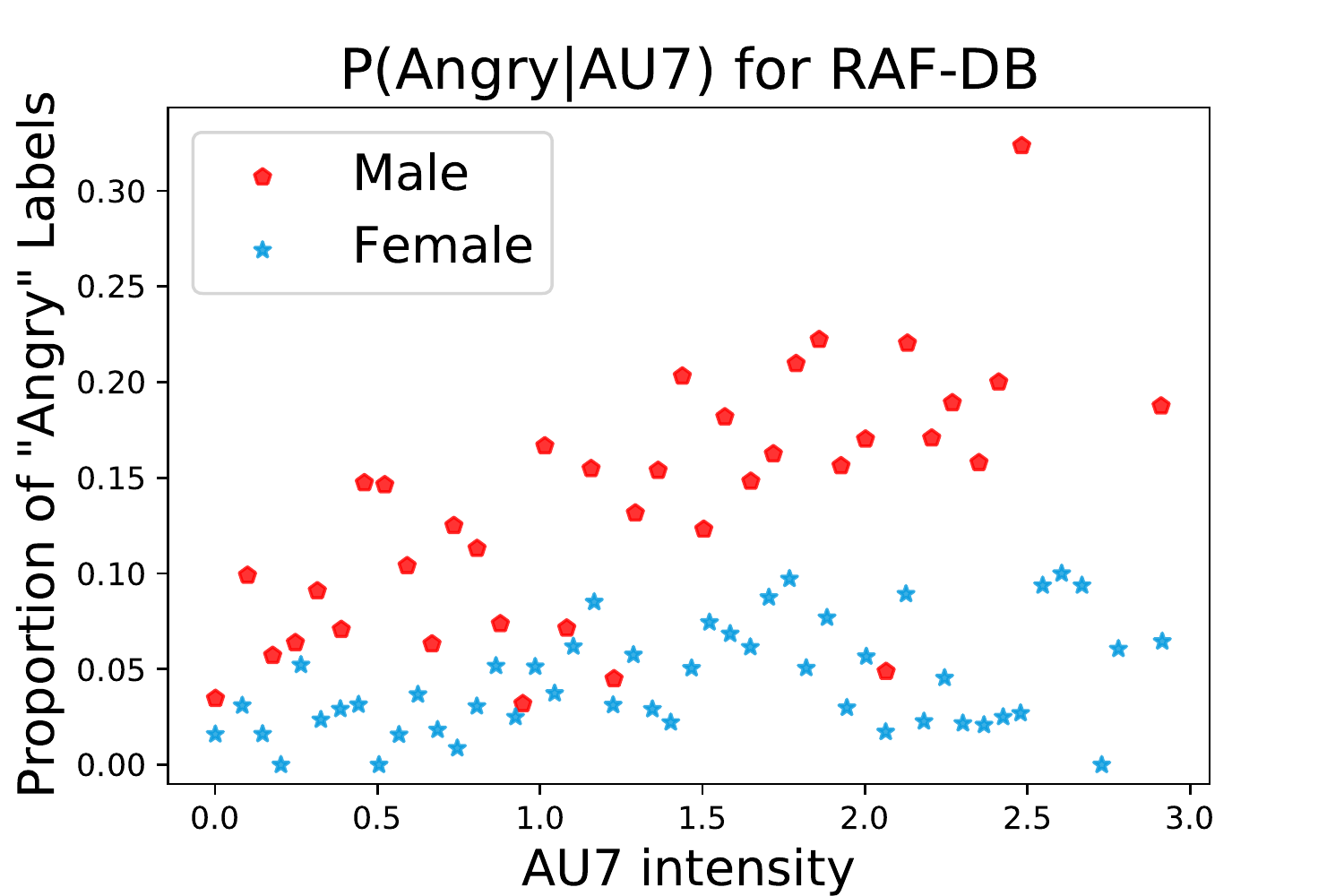}
\includegraphics[width=0.24\textwidth]{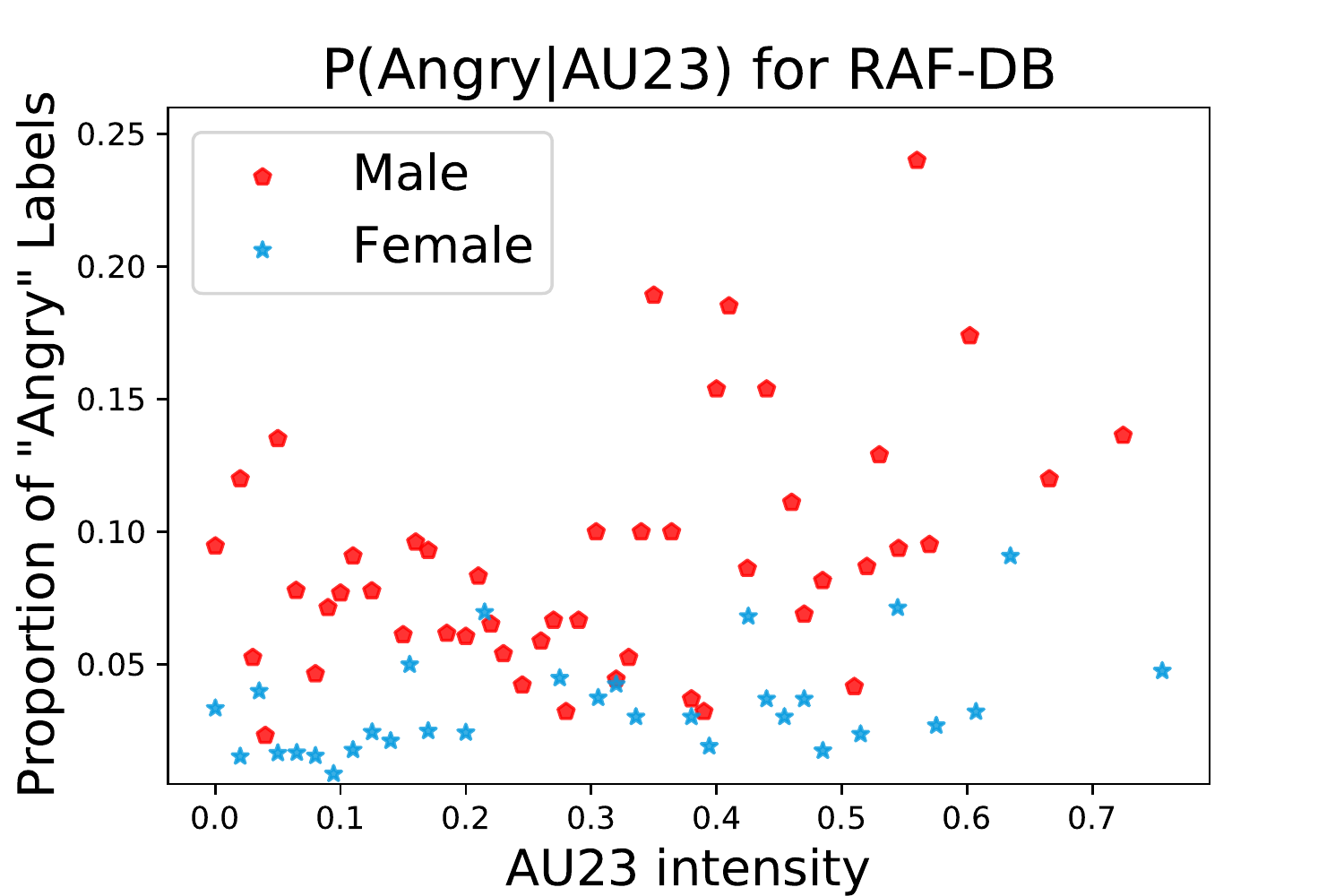}
\includegraphics[width=0.24\textwidth]{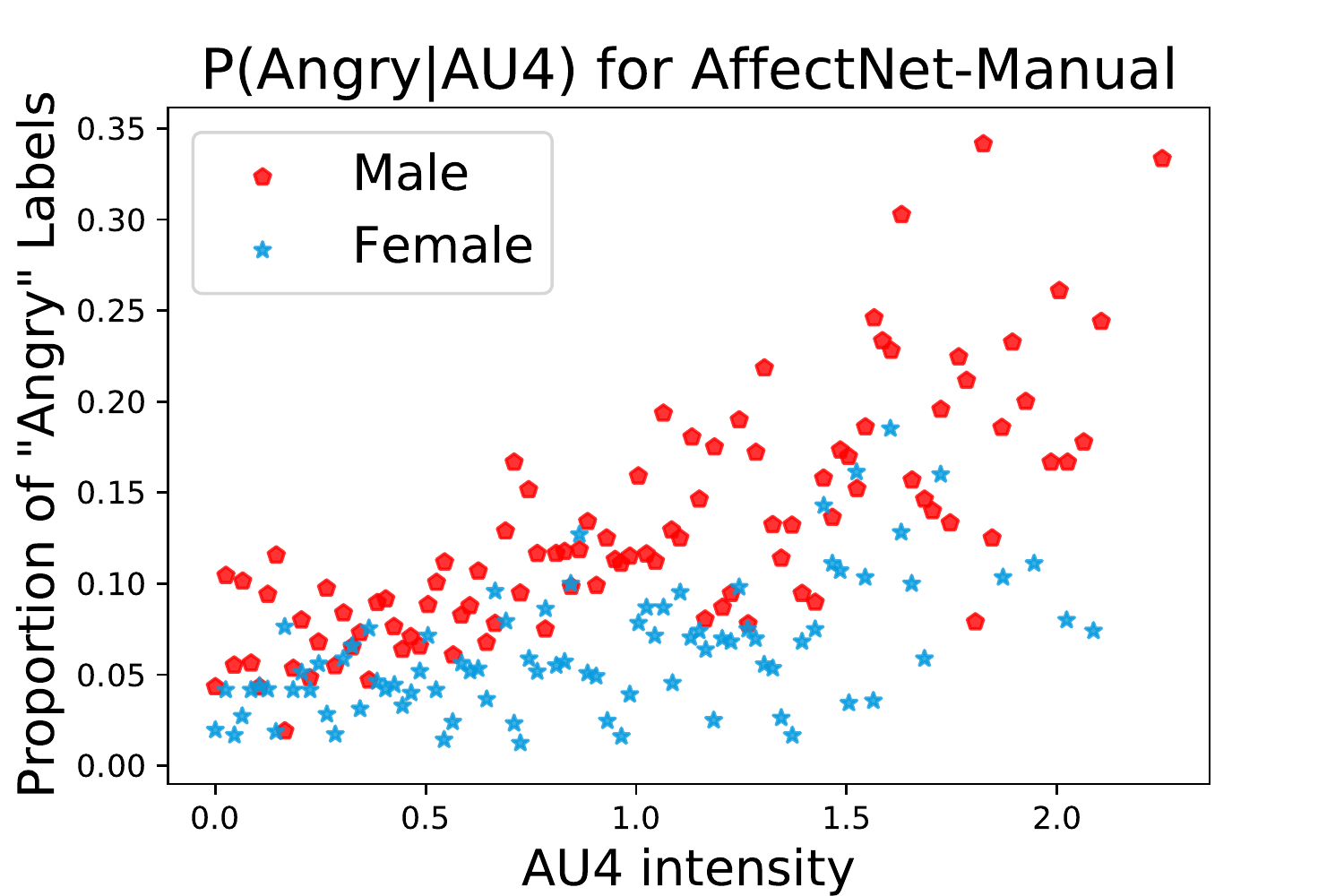}
\includegraphics[width=0.24\textwidth]{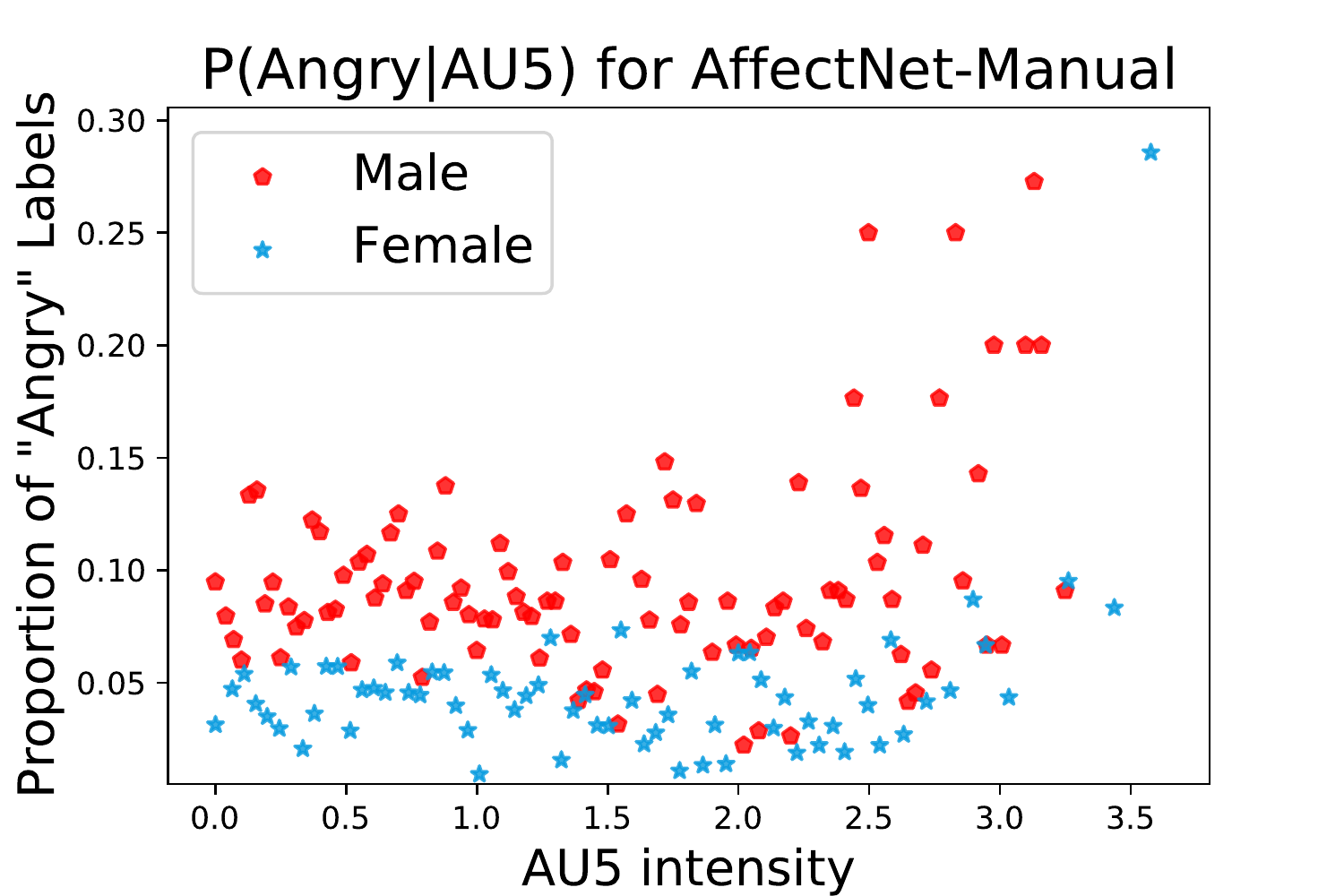}
\includegraphics[width=0.24\textwidth]{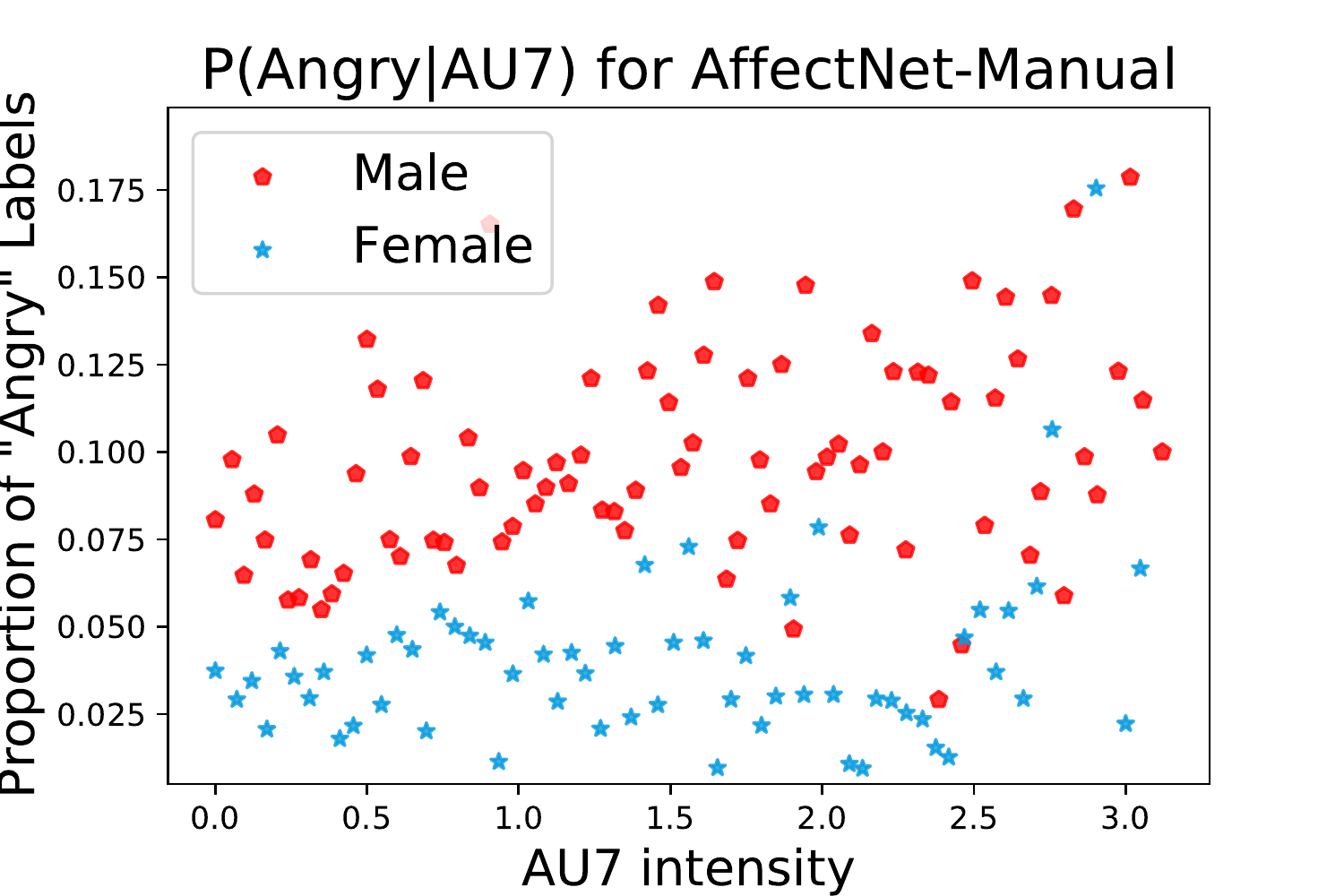}
\includegraphics[width=0.24\textwidth]{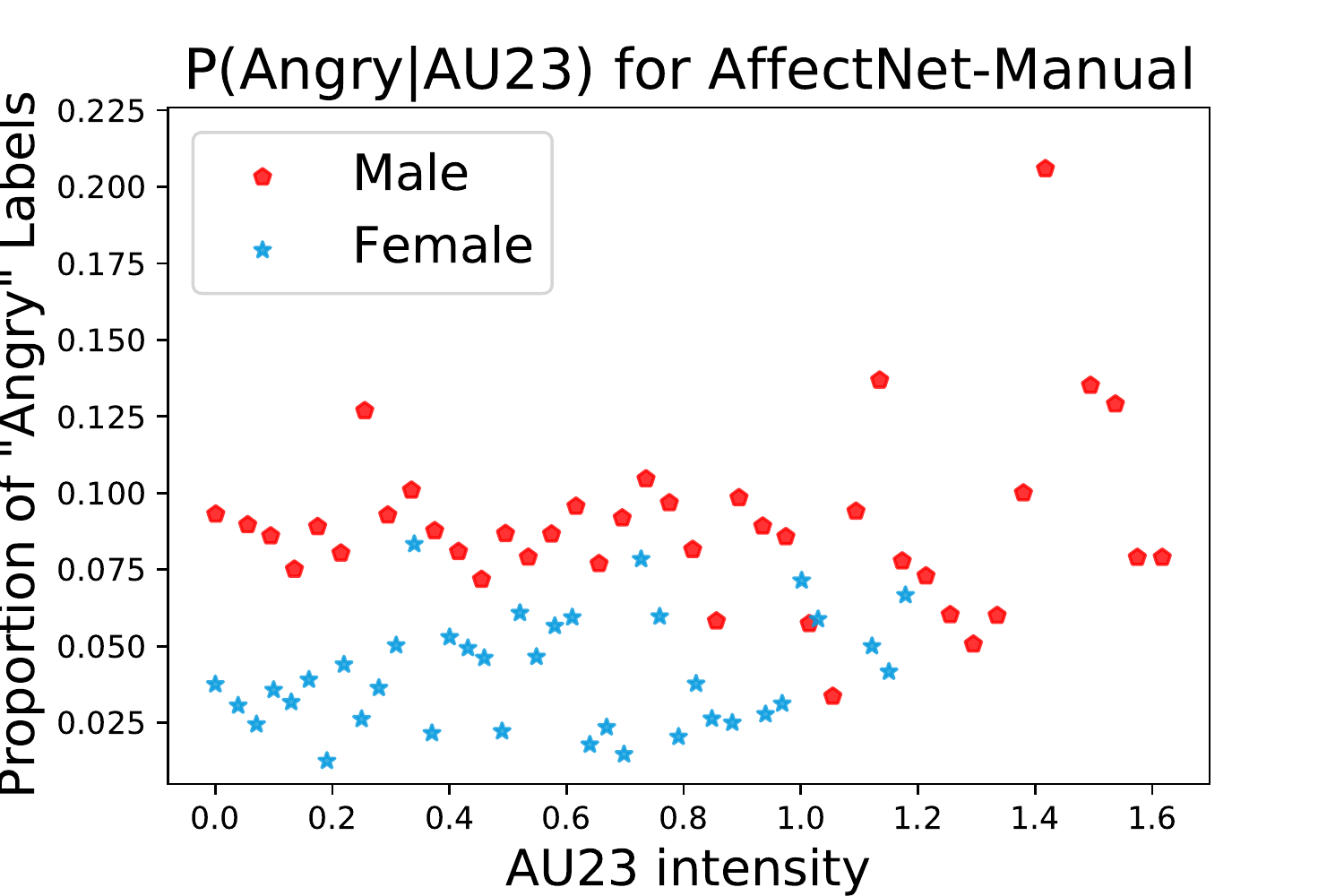}
\includegraphics[width=0.24\textwidth]{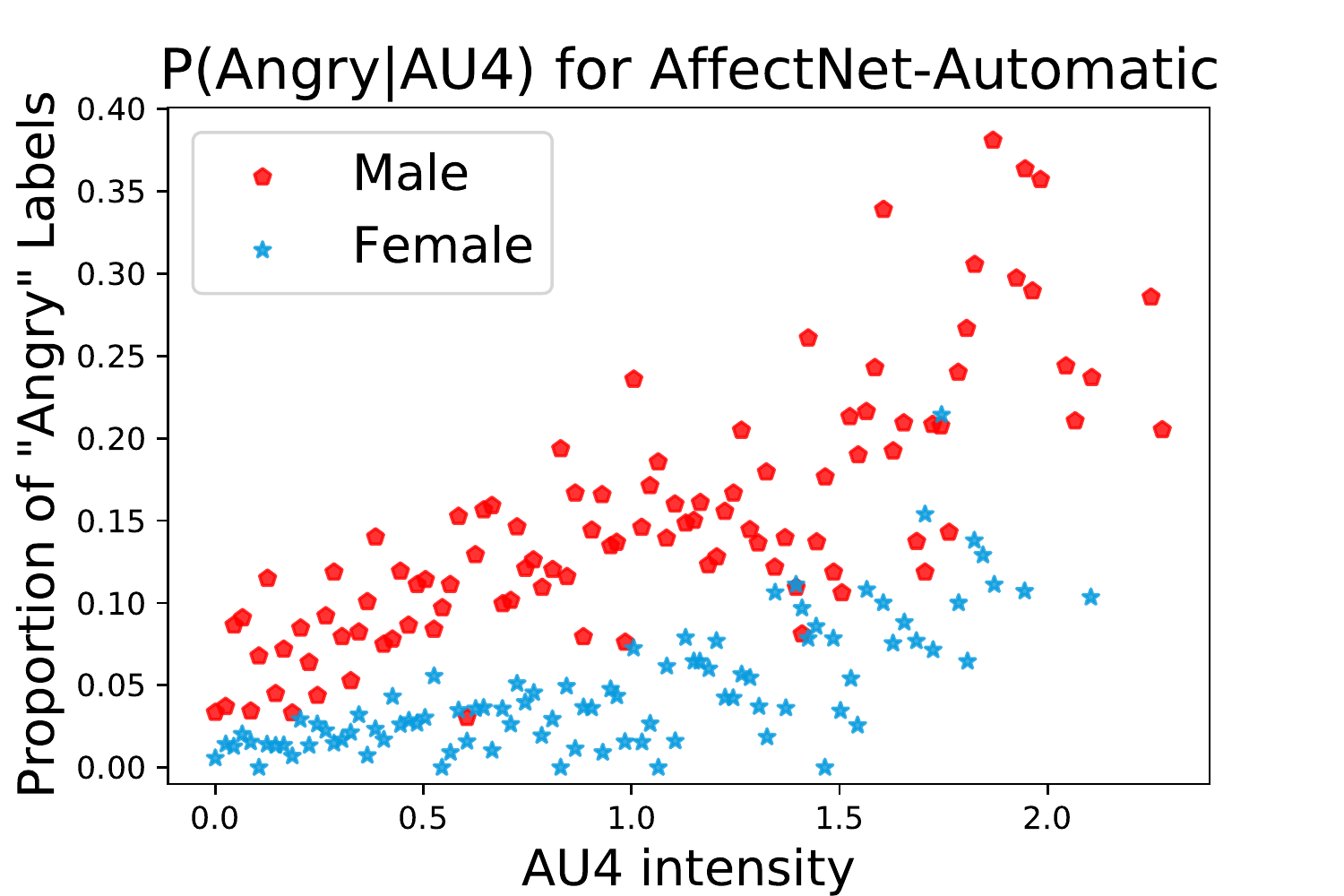}
\includegraphics[width=0.24\textwidth]{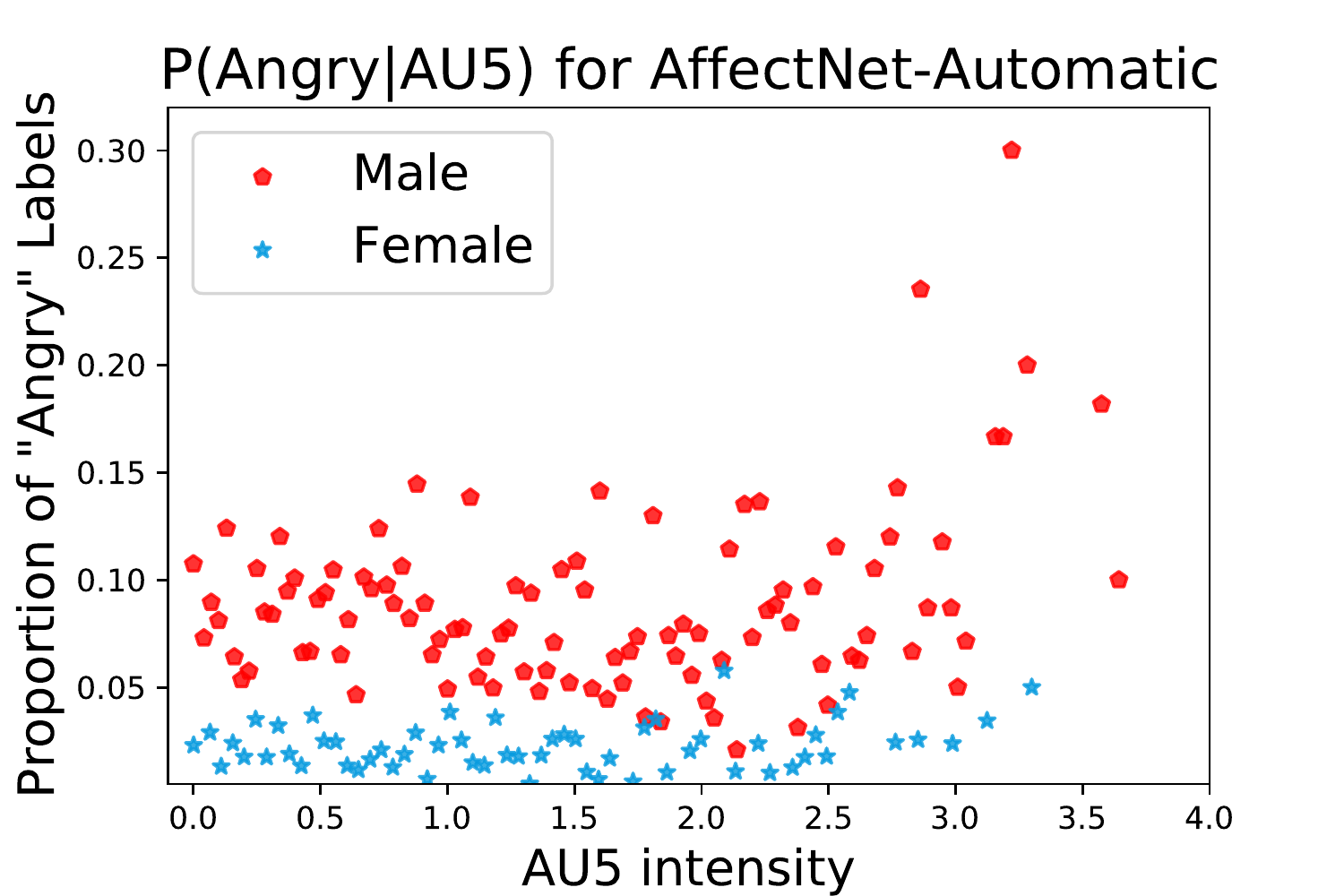}
\includegraphics[width=0.24\textwidth]{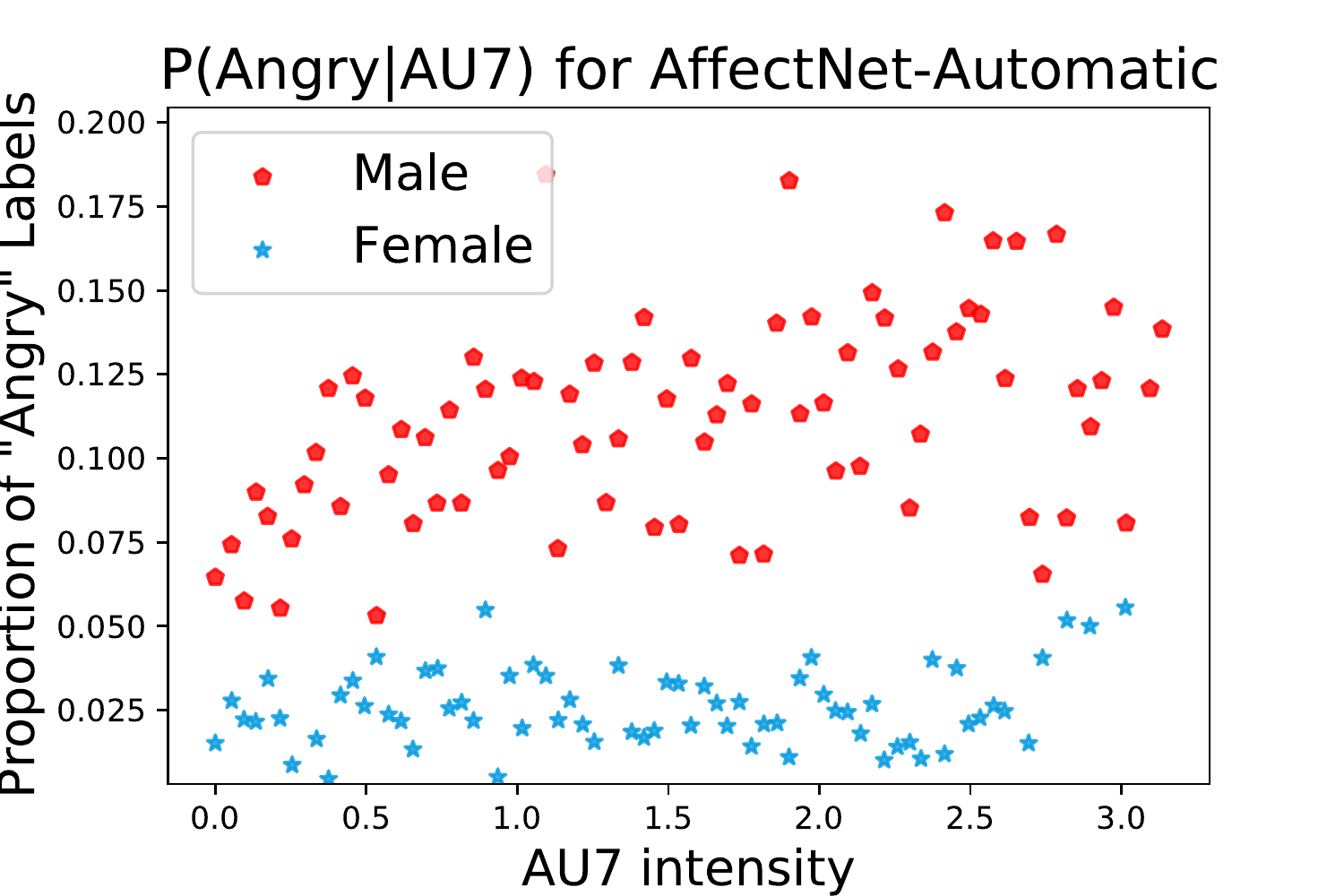}
\includegraphics[width=0.24\textwidth]{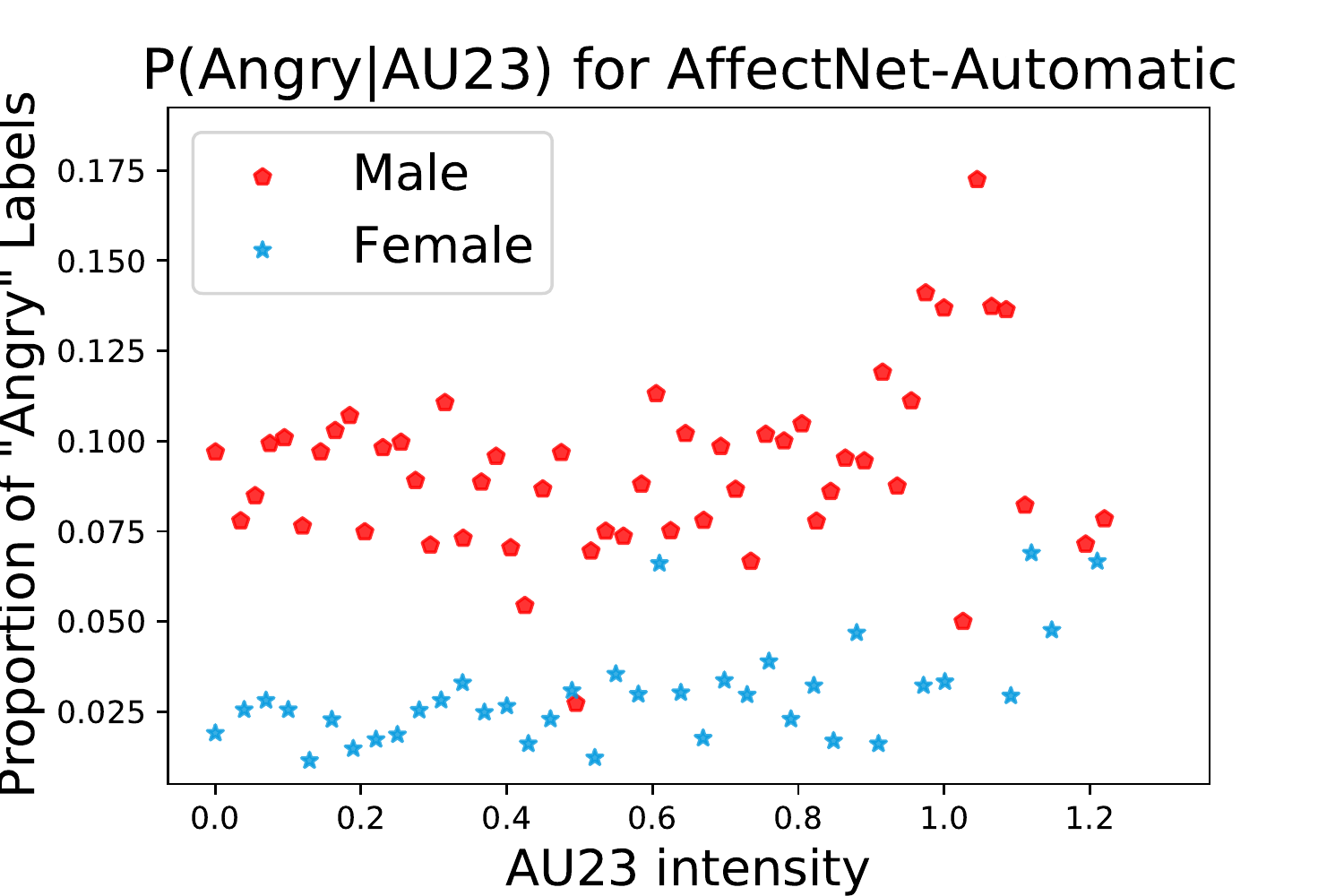}
\caption{Proportion of "angry" labels among males and females conditioned on AU4, AU5, AU7, and AU23 intensity for each in-the-wild expression dataset. Significant differences can be seen between males and females, indicating the presence of annotation bias.}
\label{fig:Anno_bias_scatterplot_angry}
\end{figure*}

Figure \ref{fig:Anno_bias_scatterplot_angry} shows the proportion of "angry" labels as a function of AU intensities for each in-the-wild dataset. The proportion of "angry" labels is higher when the AU intensities are higher, but the effect is different between males and females. All in-the-wild datasets show large discrepancies in the conditional distributions of "angry" labels between males and females. This is consistent with the result in Table \ref{tab:Anno_bias_angry_summary}, and we conclude that angry annotation bias is a prominent issue for in-the-wild datasets.

\subsection{Bias Correction for the Angry Expression}
In this section, we examine the effectiveness of the proposed AUC-FER algorithm in removing angry annotation bias. 
Similar to the experiments we did for the happiness expression, we compare our algorithm with other debiasing methods in the fairness literature, including uniform confusion \cite{alvi2018turning}, gradient reversal \cite{zhang2018mitigating}, domain discriminative training \cite{wang2020towards}, and domain independent training \cite{wang2020towards}. 

To test for robustness of our algorithm, we use AffectNet-Automatic as our training data instead of ExpW as we did for the happiness expression evaluation. The training data is of size 20,000 and is randomly sampled from AffectNet-Automatic. The test set is still constructed from CFD \cite{ma2015chicago}. In particular, we remove from CFD a few easy non-angry images (whose predicted scores from a pretrained naive classifier are less than 0.05) and then balance the number of male and female images in each AU combination. Doing so also balances the number of angry images between males and females. Similar to the experiments for the happiness expression, the thresholds for binarizing the output of the trained classifier are adjusted to maximize the accuracy on the test set. 

For all experiments, we use the ResNet-50 architecture \cite{he2016deep} pre-trained on ImageNet in PyTorch. The baseline model is a naive classifier  fine-tuned by Adam optimization \cite{kingma2014adam} with a learning rate of 0.0001 in PyTorch. 
For the four benchmark models, we follow 
Wang \etal \cite{wang2020towards} and replace the FC layer of the ResNet-50 model with two consecutive FC layers both of size 2048 with Dropout and ReLU in between. For AUC-FER, we use the PyTorch Metric Learning library \cite{musgrave2020pytorch} for the triplet loss implementation, and the hyperparameter $\lambda$ which trades off $\mathcal{L}_{softmax}$ and $\lambda \mathcal{L}_{trp}$ is set to 10. The training data is resampled so that the number of images is balanced for each gender and AU combination. The experiment is repeated 5 times for each model.

Since the attribute we are interested in is ``angry'', the Calders-Verwer (CV) discrimination score \cite{calders2010three} here is defined as: 
\begin{equation}
Disc = |P(\hat{Y} = Angry|M) - P(\hat{Y} = Angry|F)|.
\end{equation}
Table \ref{tab:debiasing_result_angry} 
shows the discrimination scores for all debiasing methods and compares them against the baseline model. All methods achieve a reduction in bias, but AUC-FER obtains the lowest discrimination score, which is a 72\% reduction in bias compared to the baseline model. 
This again shows that the proposed AUC-FER algorithm is effective in removing annotation bias. 

\begin{table}[t]
\small
\begin{center}
\scalebox{0.95}{
\begin{tabular}{|p{0.26\linewidth}|p{0.24\linewidth}|p{0.22\linewidth}|} 
\hline
\multirow{1}{\linewidth}{\parbox{1.7\linewidth}{\vspace{0.09cm} \centering Methods\\ (ResNet-50~\cite{he2016deep})}}  & \multirow{1}{\linewidth}{\parbox{1\linewidth}{\vspace{0.25cm} \centering $Disc$}}        & \multicolumn{1}{>{\centering\arraybackslash}p{0.22\linewidth}|}{Compared to Baseline (\%)} \\ 
\hhline{|===|}
\multicolumn{1}{|c|}{Baseline}    & \multicolumn{1}{c|}{0.064 $\pm$ 0.020}     & \multicolumn{1}{c|}{-}         \\
\multicolumn{1}{|c|}{Uniform Confusion~\cite{alvi2018turning}}    & \multicolumn{1}{c|}{0.044 $\pm$ 0.025}     & \multicolumn{1}{c|}{68.3}         \\
\multicolumn{1}{|c|}{Gradient Projection~\cite{zhang2018mitigating}}  & \multicolumn{1}{c|}{0.031 $\pm$ 0.031}    & \multicolumn{1}{c|}{49.2}          \\
\multicolumn{1}{|c|}{Domain Discriminative~\cite{wang2020towards}}   & \multicolumn{1}{c|}{0.041 $\pm$ 0.051}  & \multicolumn{1}{c|}{63.5}        \\
\multicolumn{1}{|c|}{Domain Independent~\cite{wang2020towards}}   & \multicolumn{1}{c|}{0.021 $\pm$ 0.018}   & \multicolumn{1}{c|}{33.3}            \\
\multicolumn{1}{|c|}{AUC-FER (Ours)}   & \multicolumn{1}{c|}{\textbf{0.018 $\pm$ 0.039}}  & \multicolumn{1}{c|}{\textbf{28.2}} \\
\hline
\end{tabular}}
\end{center}
\vspace{-5pt}
\caption{Discrimination scores for various debiasing methods using the ResNet-50 architecture trained on random subsets of AffectNet-Automatic of size 20,000 and tested on CFD for the anger expression. 
}
\label{tab:debiasing_result_angry}
\vspace{-10pt}
\end{table}

\begin{revision}
\section{Happy Annotation Bias Across Age Groups}

\begin{table*}[h]
\small
\centering
\begin{threeparttable}\scalebox{0.95}{
\begin{tabular*}{1\textwidth}{|m{0.1\linewidth}||m{0.17\linewidth}|m{0.045\linewidth}|m{0.045\linewidth}|m{0.05\linewidth}|m{0.045\linewidth}|m{0.075\linewidth}||m{0.045\linewidth}|m{0.045\linewidth}|m{0.075\linewidth}|}
\cline{1-10}
\multirow{2}{\linewidth}{\parbox{1\linewidth}{\vspace{0.1cm} \centering Data}} &\multirow{2}{\linewidth}{\parbox{1\linewidth}{\centering Metrics \\ P(Happy\textbar{}AU6, AU12)}} & \multicolumn{5}{c||}{4 Age Groups} & \multicolumn{3}{c|}{2 Age Groups} \\ 

\cline{3-10}
             &                   & $\leq$ 19    & 20-39  & 40-59\tnote{1}  &  $\geq$ 60\tnote{1}    & \multicolumn{1}{c|}{p-value}   &  $\leq$ 39    &  $\geq$ 40    & \multicolumn{1}{c|}{p-value}   \\ 
\hhline{|=::======::===|}
\multirow{6}{\linewidth}{\centering ExpW \cite{SOCIALRELATION_ICCV2015,SOCIALRELATION_2017}}                & P(Happy\textbar{}(0, 0)) & 0.169 & \textbf{0.190} & 0.177 & 0.207 & 0.000 *** & 0.187 & 0.191 & 0.382     \\
                    & P(Happy\textbar{}(1, 0)) & 0.190 & 0.262 & 0.251  & \textbf{0.303}  & 0.017 *   & 0.251 & 0.274 & 0.306     \\
                    & P(Happy\textbar{}(0, 1)) & \textbf{0.763} & 0.708 & 0.586 & 0.626 & 0.000 *** & \textbf{0.716} & 0.606 & 0.000 ***  \\
                    & P(Happy\textbar{}(1, 1)) & \textbf{0.838} & 0.832 & 0.765 & 0.806  & 0.000 *** & \textbf{0.833} & 0.785 & 0.000 ***  \\ 
\cline{2-10}
                    & P(Happy)                 & 0.321 & 0.335 & 0.299 & 0.346  &           & 0.333 & 0.320 &            \\ 
\cline{2-10}
                    & Number of samples        & 11,561  & 62,622  & 6,033   & 5,139   &           & 74,183  & 11,172  &            \\ 
\hhline{|=::======::===|}
\multirow{6}{\linewidth}{\centering RAF-DB \cite{li2017reliable}}             & P(Happy\textbar{}(0, 0)) & 0.231 & 0.168 & \textbf{0.249} & 0.130 & 0.000 *** & \textbf{0.234} & 0.168 & 0.000 ***  \\
                    & P(Happy\textbar{}(1, 0)) & \textbf{0.320} & 0.180 & 0.285 & 0.250   & 0.003 **    & \textbf{0.410} & 0.180 & 0.000 ***  \\
                    & P(Happy\textbar{}(0, 1)) & 0.859 & 0.837 & 0.891 & 0.741 & 0.209     & 0.877  & 0.837 & 0.076 .    \\
                    & P(Happy\textbar{}(1, 1)) & 0.837  & 0.857 & \textbf{0.903} & 0.879 & 0.010 **   & \textbf{0.903} & 0.857 & 0.048 *    \\ 
\cline{2-10}
                    & P(Happy)                 & 0.379 & 0.340 & 0.536 & 0.433 &           & 0.438 & 0.340 &            \\ 
\cline{2-10}
                    & Number of samples        & 3,205   & 6,805   & 1,911   & 293    &           & 10,111  & 2,103   &            \\ 
\hhline{|=::======::===|}
\multirow{6}{\linewidth}{\centering AffectNet-Manual \cite{mollahosseini2017affectNet}}     & P(Happy\textbar{}(0, 0)) & 0.141  & \textbf{0.157} & 0.078 & 0.078 & 0.000 *** & \textbf{0.112} & 0.078 & 0.000 ***  \\
                    & P(Happy\textbar{}(1, 0)) & \textbf{0.407} & 0.241 & 0.225 & 0.272 & 0.023 *   & 0.272 & 0.240 & 0.287      \\
                    & P(Happy\textbar{}(0, 1)) & \textbf{0.746} & 0.697 & 0.599 & 0.525 & 0.000 *** & \textbf{0.705} & 0.584 & 0.000 ***  \\
                    & P(Happy\textbar{}(1, 1)) & 0.824 & \textbf{0.839} & 0.797 & 0.749 & 0.000 *** & \textbf{0.837}  & 0.784  & 0.000 ***  \\ 
\cline{2-10}
                    & P(Happy)                 & 0.342 & 0.324 & 0.282 & 0.282 &           & 0.327 & 0.282 &            \\ 
\cline{2-10}
                    & Number of samples        & 4,041   & 20,486  & 8,477   & 2,690   &           & 24,527  & 11,167  &            \\ 
\hhline{|=::======::===|}
\multirow{6}{\linewidth}{\centering AffectNet-Automatic \cite{mollahosseini2017affectNet}}  & P(Happy\textbar{}(0, 0)) & \textbf{0.244} & 0.184 & 0.160 & 0.153 & 0.000 *** & \textbf{0.198}  & 0.158 & 0.000 ***  \\
                    & P(Happy\textbar{}(1, 0)) & 0.591 & 0.428 & 0.435 & \textbf{0.436} & 0.017 *   & 0.478 & 0.436 & 0.210       \\
                    & P(Happy\textbar{}(0, 1)) & \textbf{0.867} & 0.858  & 0.822 & 0.711 & 0.000 *** & \textbf{0.860} & 0.800 & 0.000 ***  \\
                    & P(Happy\textbar{}(1, 1)) & 0.887 & 0.932 & \textbf{0.932} & 0.882 & 0.000 *** & 0.924 & 0.918  & 0.436      \\ 
\cline{2-10}
                    & P(Happy)                 & 0.437  & 0.423 & 0.394 & 0.379 &           & 0.426 & 0.390   &            \\ 
\cline{2-10}
                    & Number of samples        & 6,876   & 24,111  & 8,432   & 2,865   &           & 30,987  & 11,297  &            \\
\cline{1-10}
\end{tabular*}
}
\begin{tablenotes}
\item[] Signif. codes:  0 ‘***’ 0.001 ‘**’ 0.01 ‘*’ 0.05 ‘.’ 0.1 ‘ ’ 1
\item[1] For RAF-DB, the 4 age groups are $\leq$ 19, 20 - 39, 40 - 69, and $\geq$ 70.
\end{tablenotes}
\end{threeparttable}
\caption{Conditional and marginal distributions of "happy" labels along with the numbers of samples for each age group for each in-the-wild dataset. The p-values are the $\chi^2$ tests for independence of the "happy" labels and the age groups. When the p-values are significant at the 0.05 level, the age groups with the highest proportion of "happy" labels are highlighted.}
\label{tab:Anno_bias_happy_age_summary}
\end{table*}

In this section, we describe the evaluation of annotation bias for the happy expression across age groups. We note that most of the datasets (both lab-controlled and in-the-wild) are severely dominated by younger people, making the statistical tests challenging.

Since only RAF-DB includes age labels, we first train a simple age classifier with ResNet-34 architecture \cite{he2016deep} using the FairFace dataset \cite{karkkainen2021fairface} similar to that for training a gender classifier. We then apply the trained classifier on the datasets that do not have age labels (\ie, KDEF, CFD, ExpW, and AffectNet). The age predictions are then grouped into the following 4 groups: ``less than 19'', ``20-39'', ``40-59'', and ``more than 60.'' The original labels provided by RAF-DB are ``0-3,'' ``4-19,'' ``20-39,'' ``40-69,'' and ``70+.'' We group the ``0-3'' and ``4-19'' age groups together to increase the number of samples in each age group. In addition to this 4-group categorization, we also group the images into ``less than 40'' and ``more than 40'' to further alleviate the problem of having too few samples in some age groups when conditioning on the AUs and to investigate general discrepancies between younger and older people. However, the lab-controlled datasets contain too few images in the older age groups (\eg, only 26 out of the 1,207 total images are predicted to be older than 40 years old). As a result, we are unable to evaluate the annotation bias for the two lab-controlled datasets.

\begin{figure*}[t]
\centering
\includegraphics[width=0.23\textwidth]{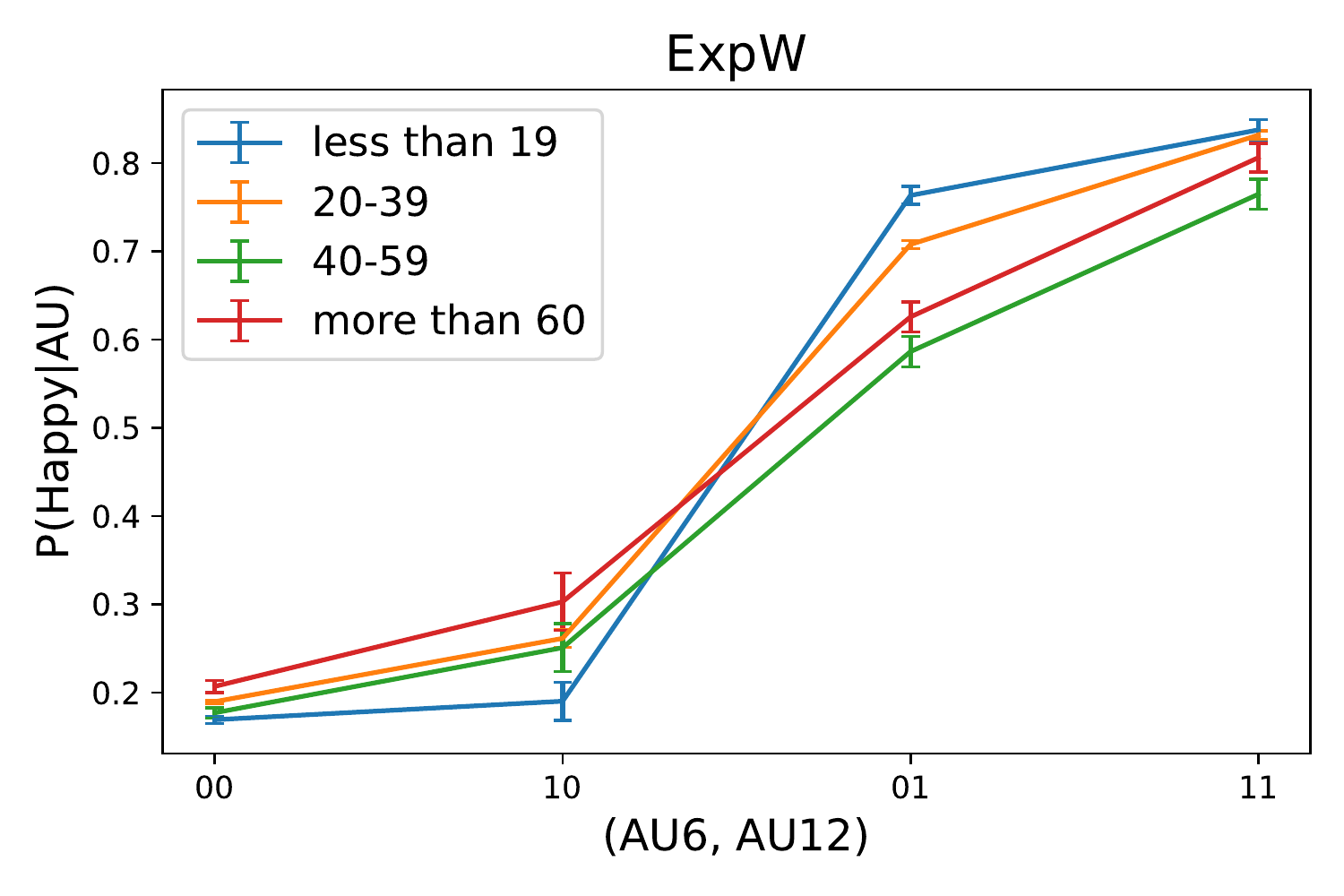}
\includegraphics[width=0.23\textwidth]{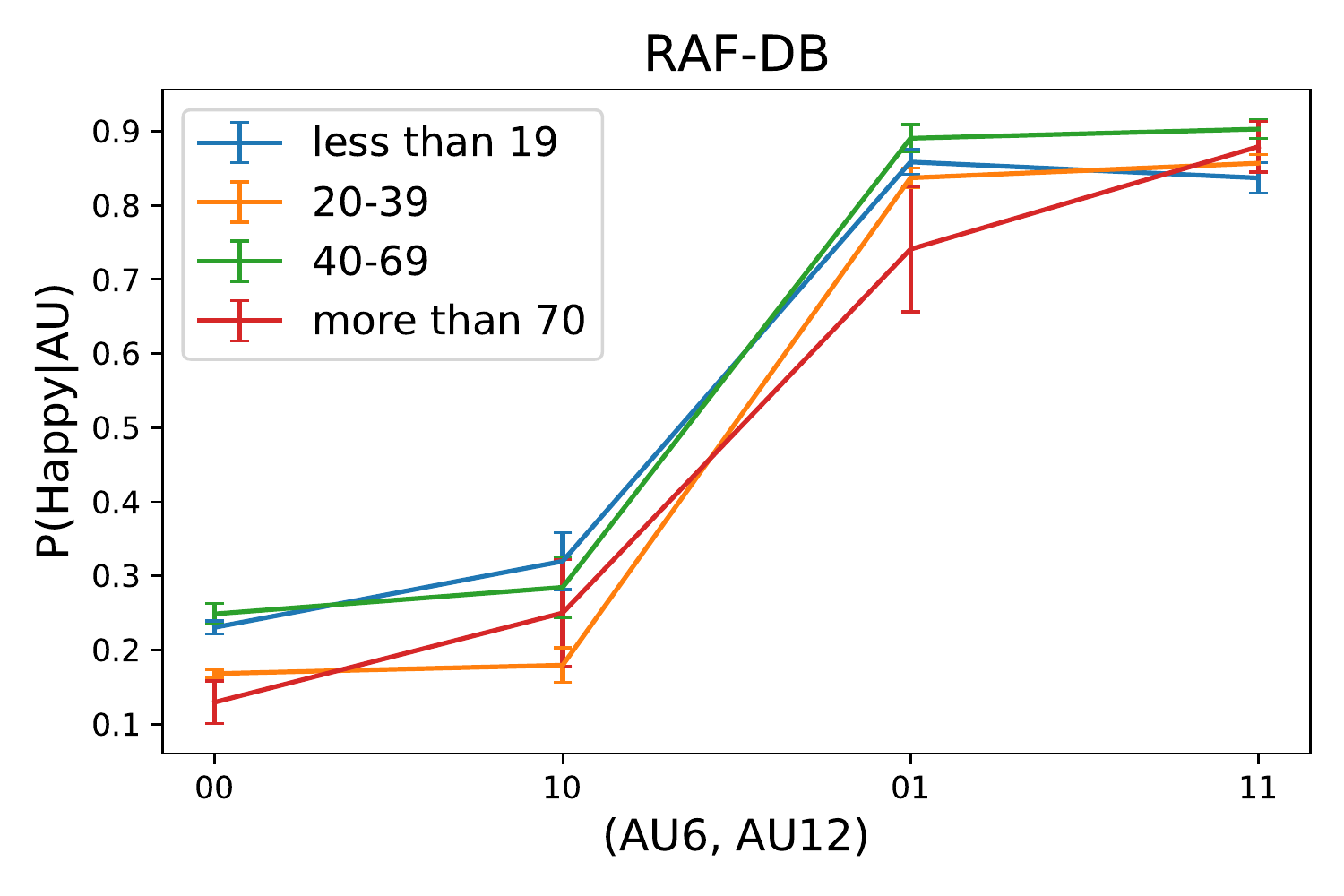}
\includegraphics[width=0.23\textwidth]{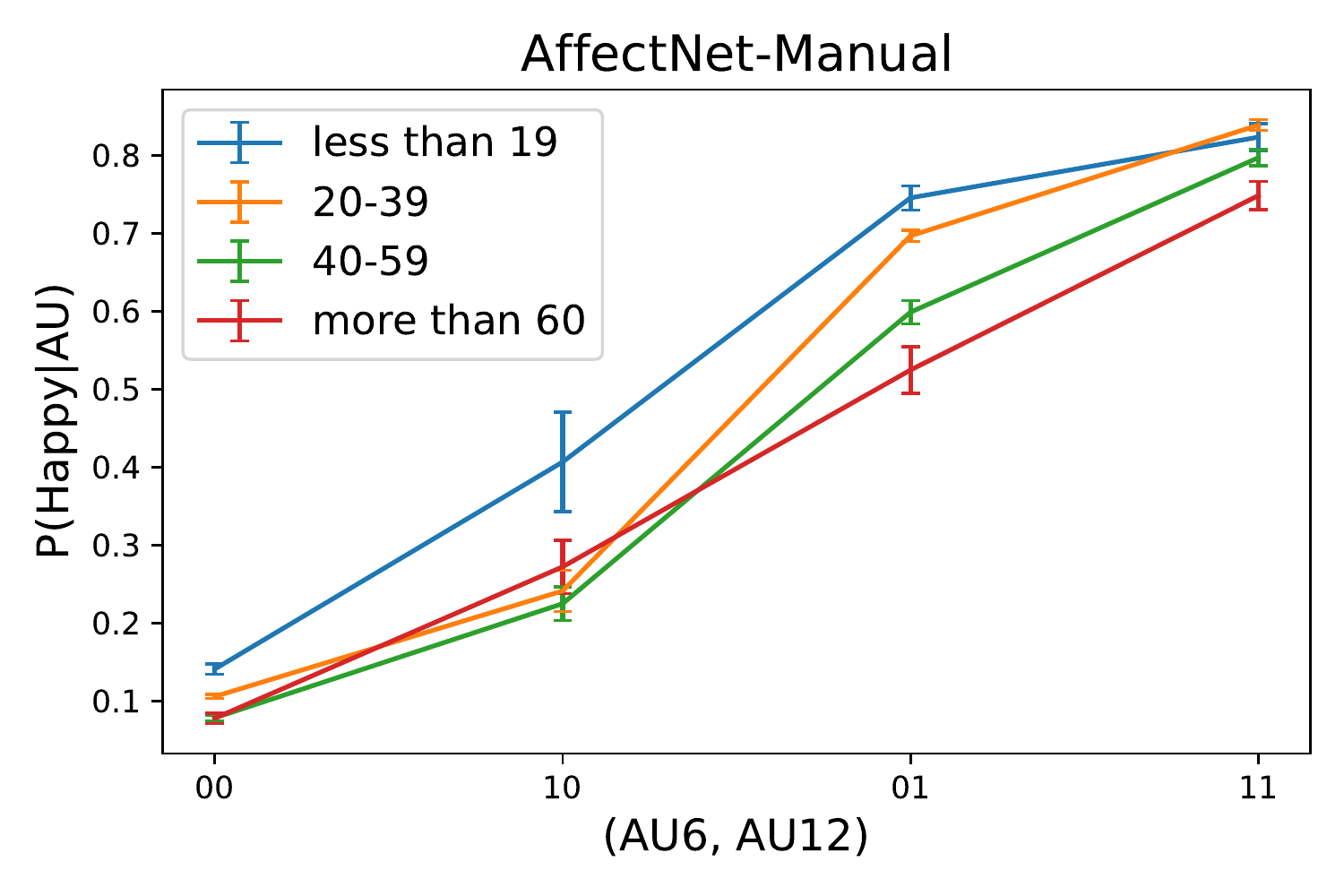}
\includegraphics[width=0.23\textwidth]{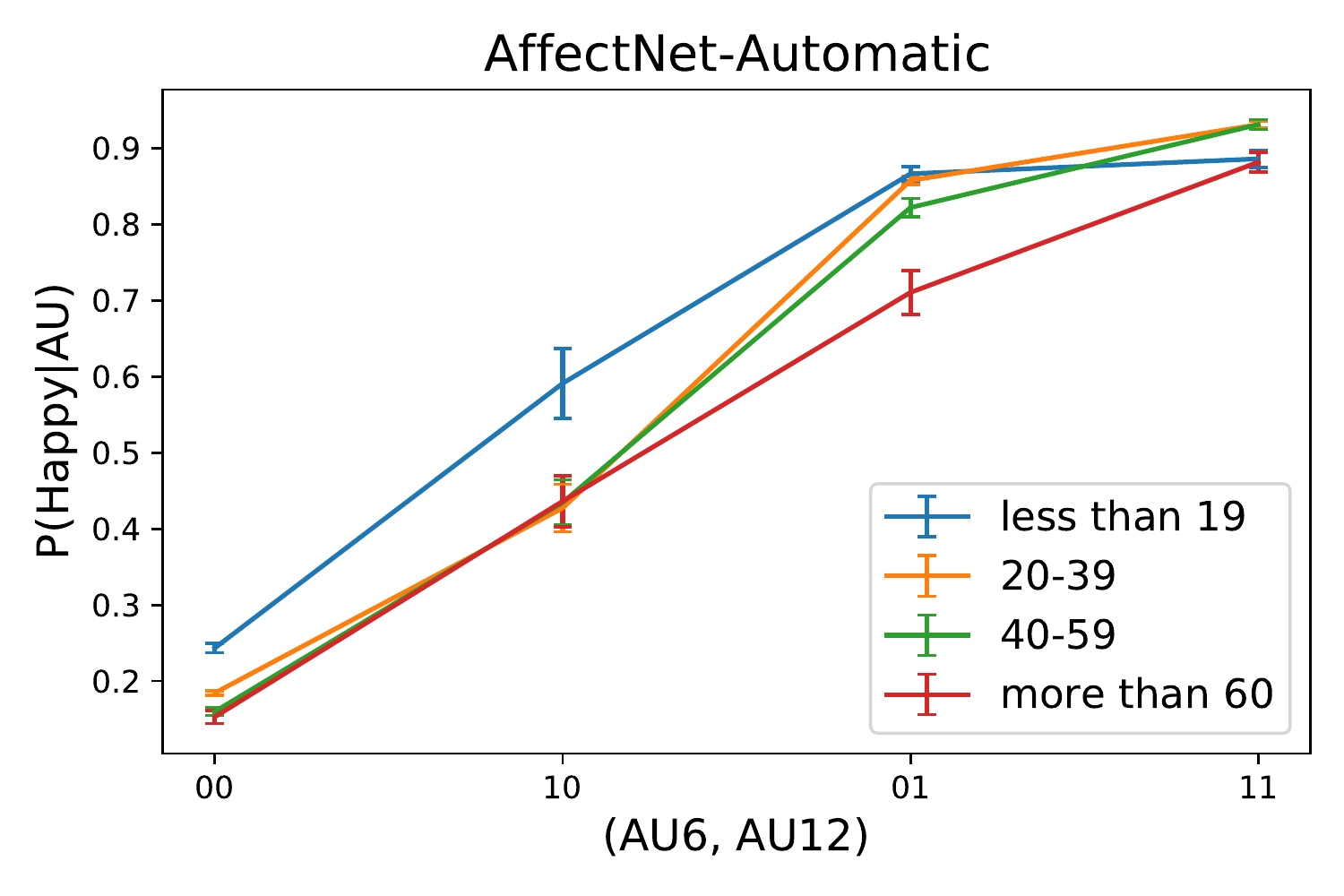}
\includegraphics[width=0.23\textwidth]{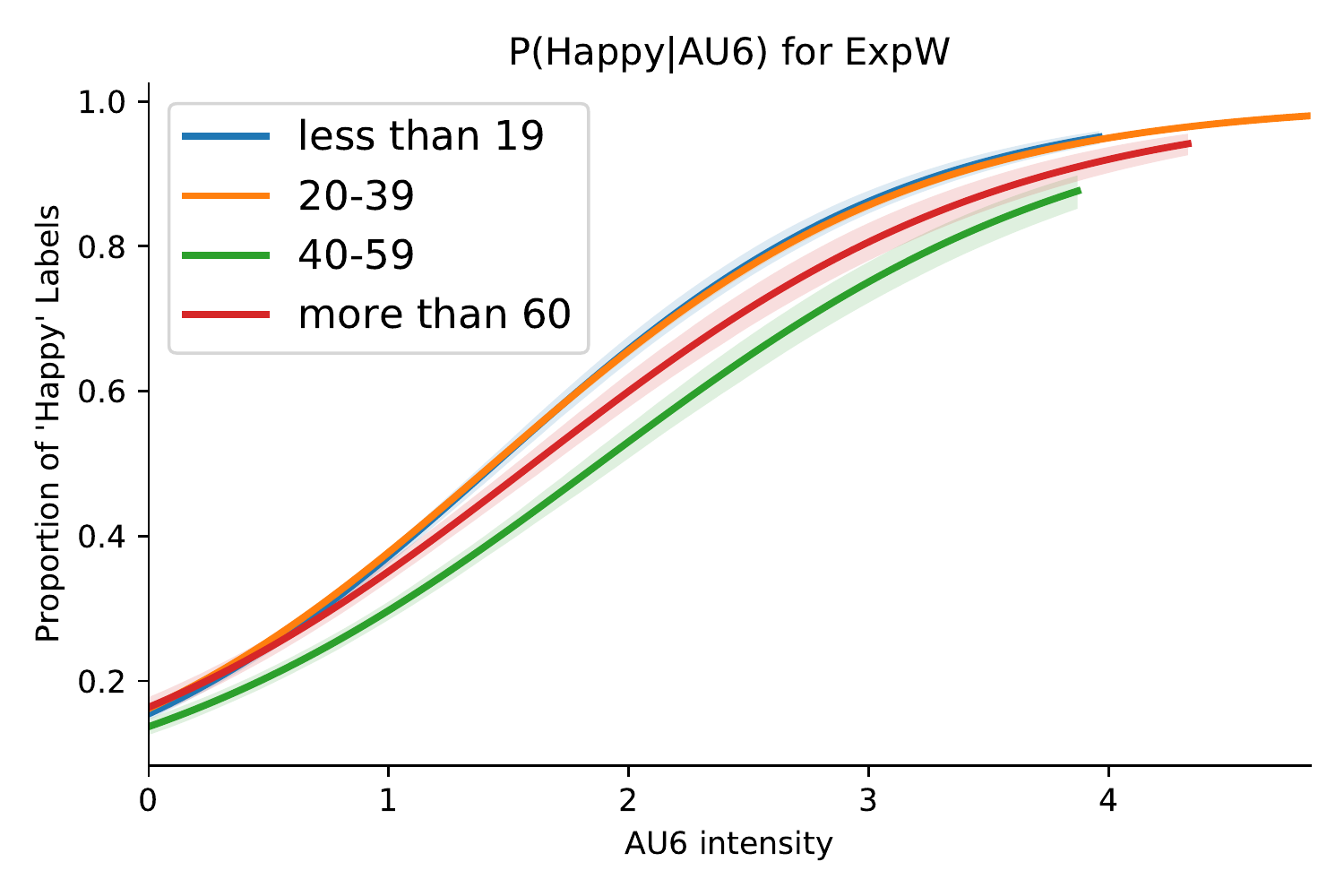}
\includegraphics[width=0.23\textwidth]{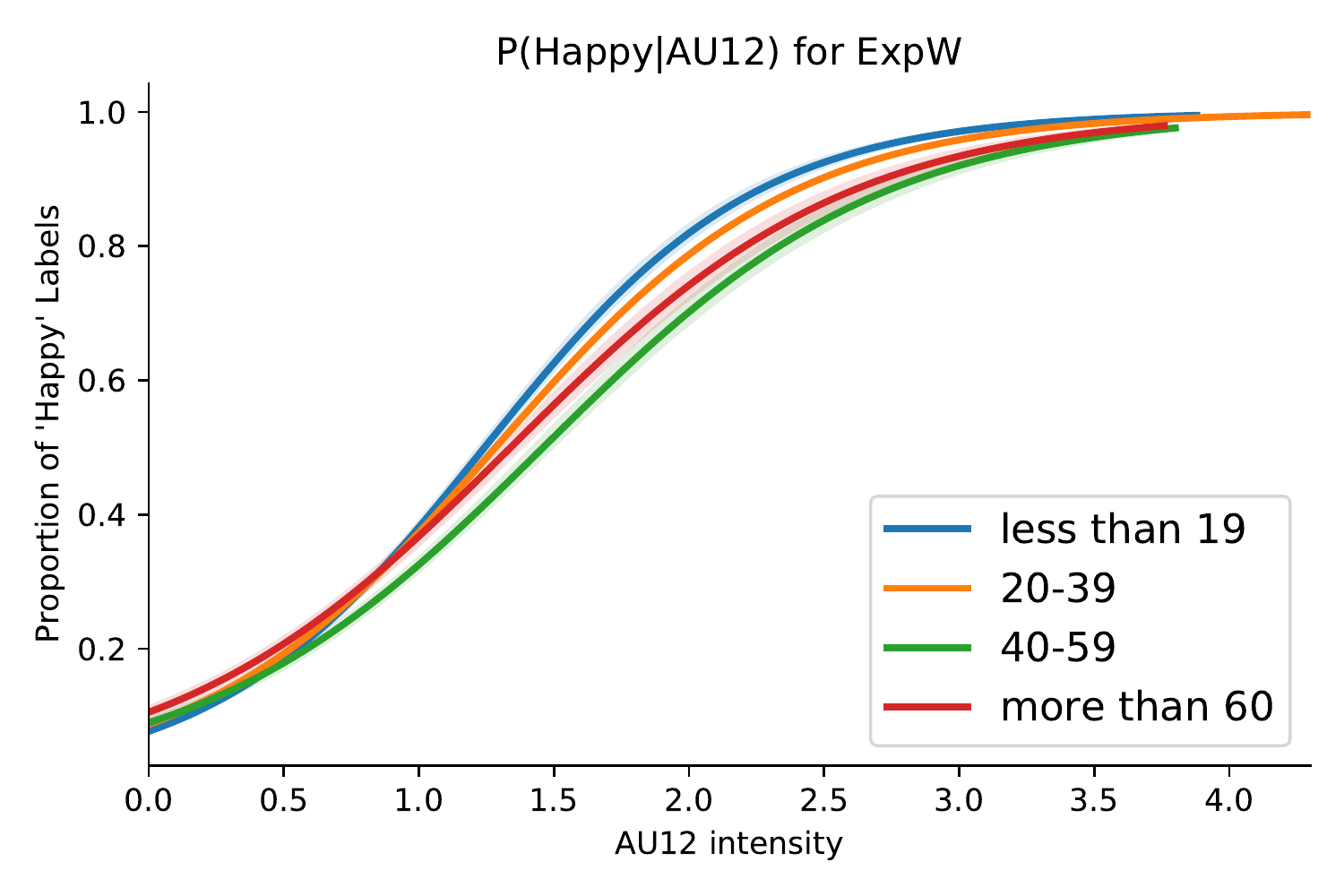}
\includegraphics[width=0.23\textwidth]{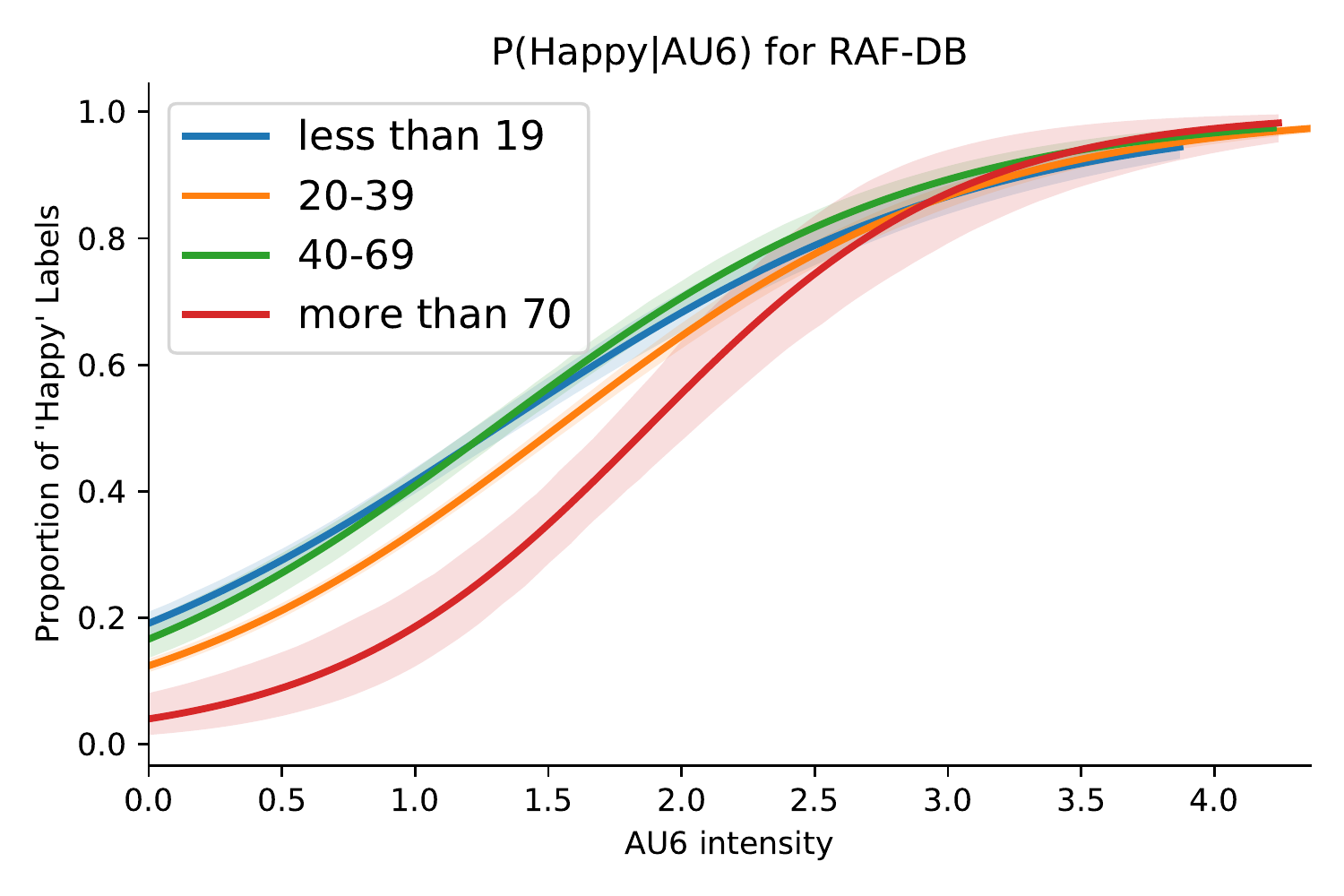}
\includegraphics[width=0.23\textwidth]{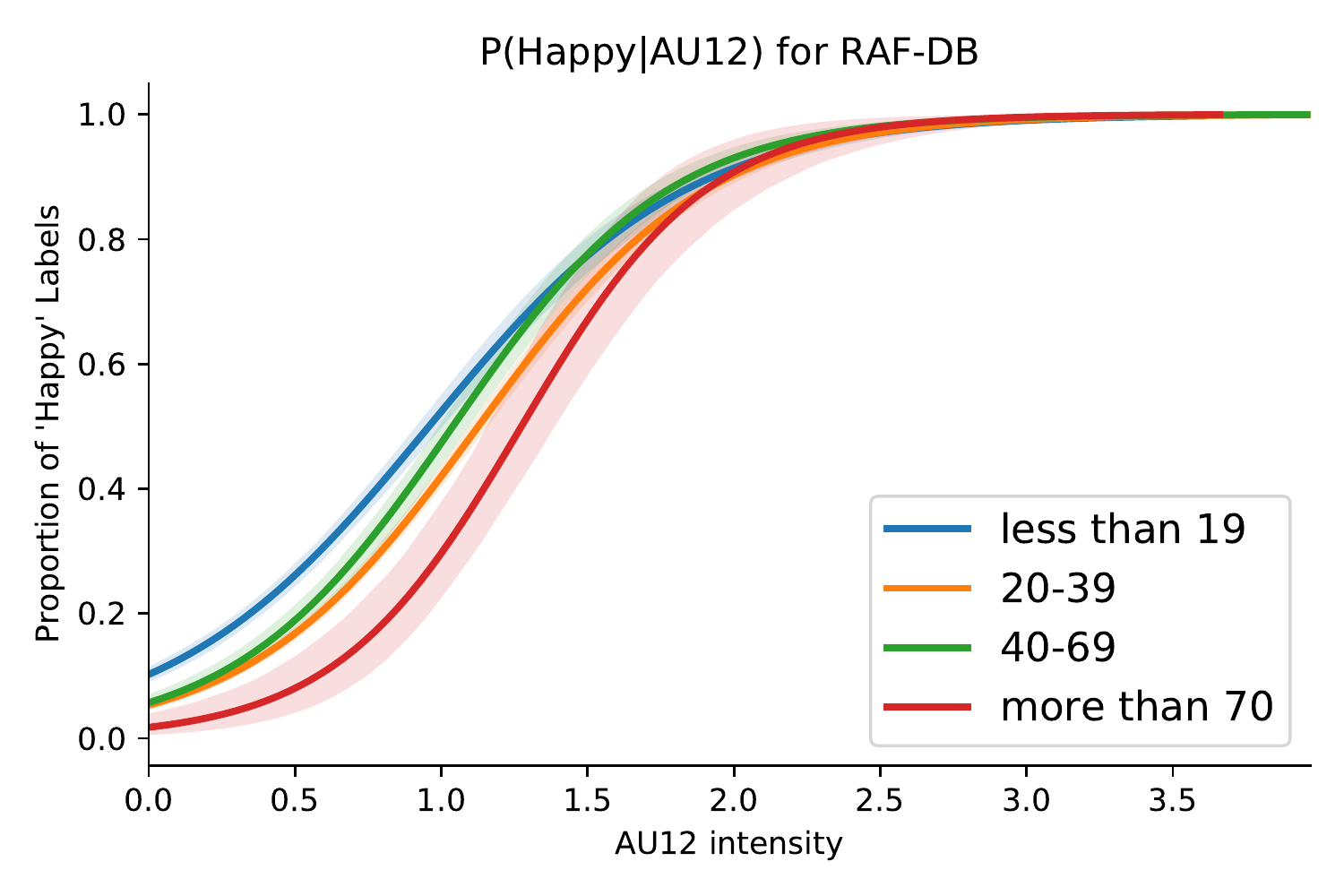}
\includegraphics[width=0.23\textwidth]{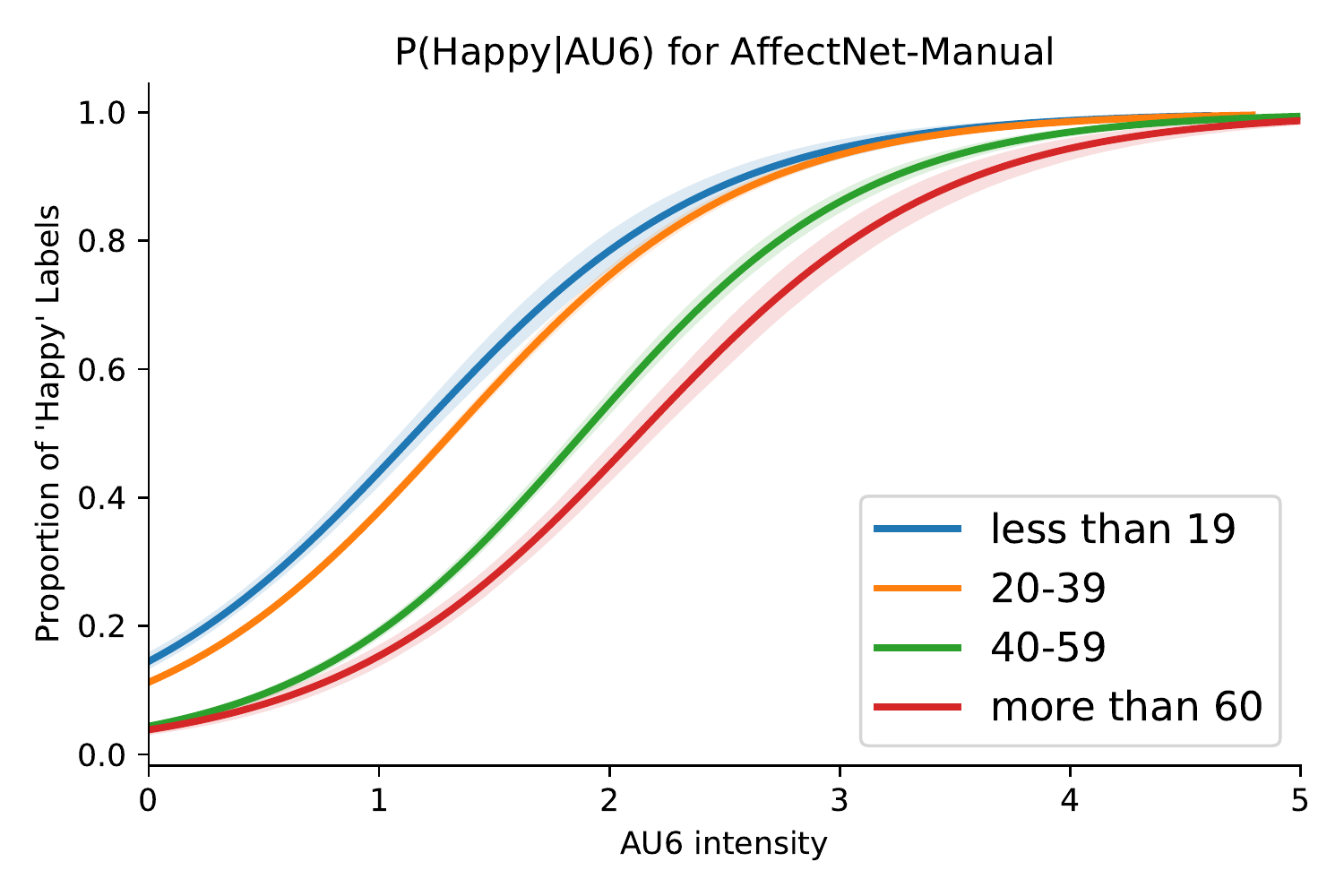}
\includegraphics[width=0.23\textwidth]{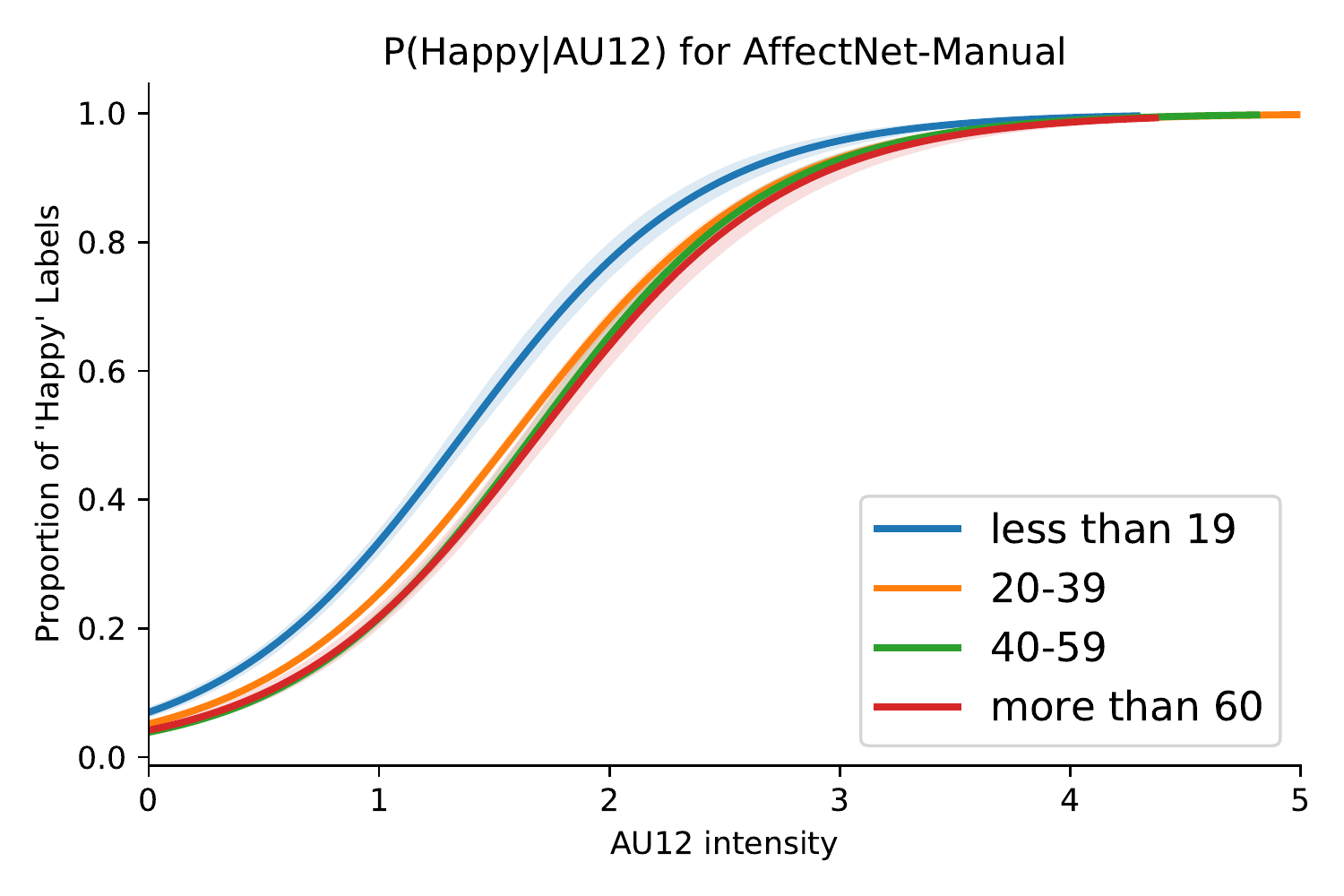}
\includegraphics[width=0.23\textwidth]{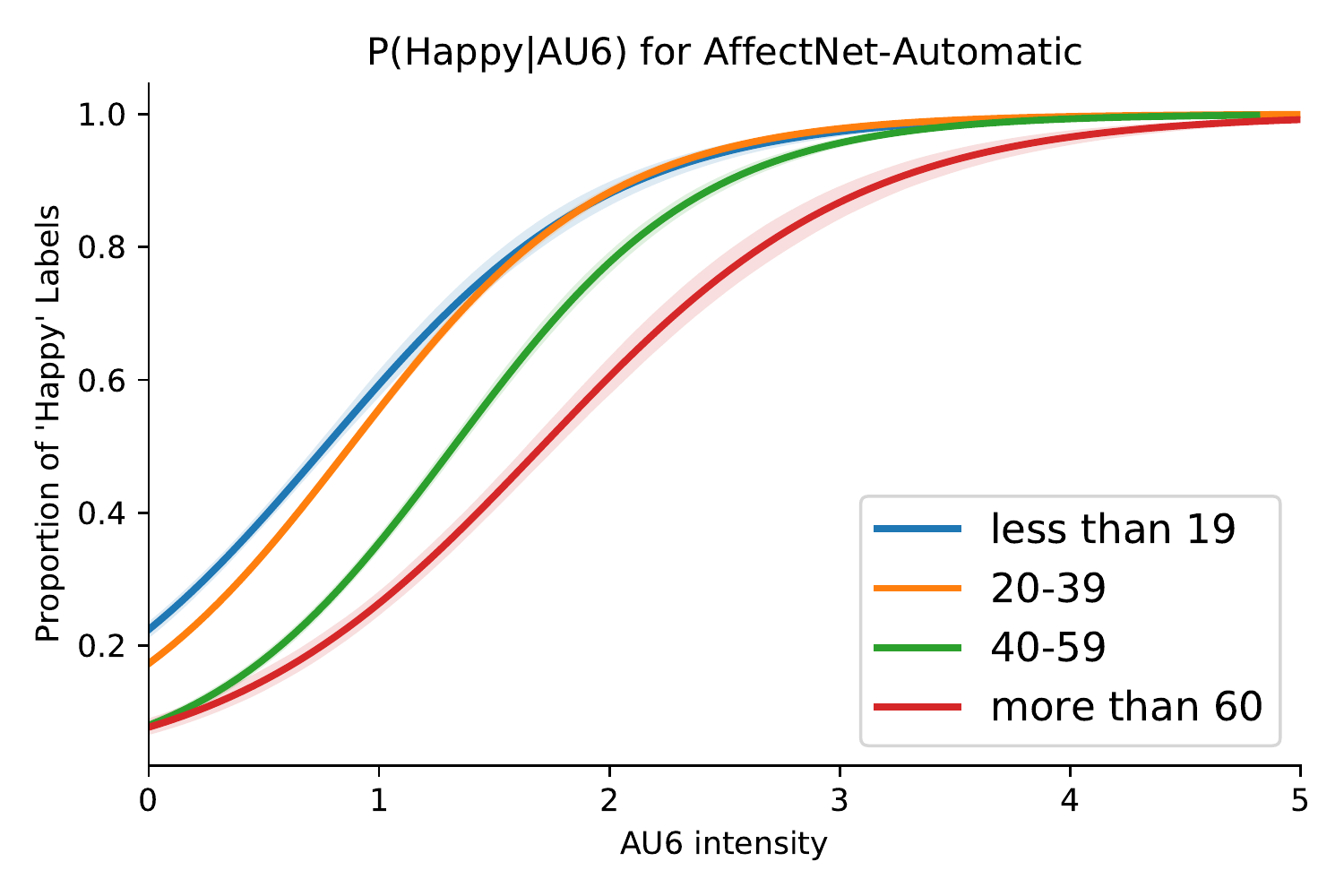}
\includegraphics[width=0.23\textwidth]{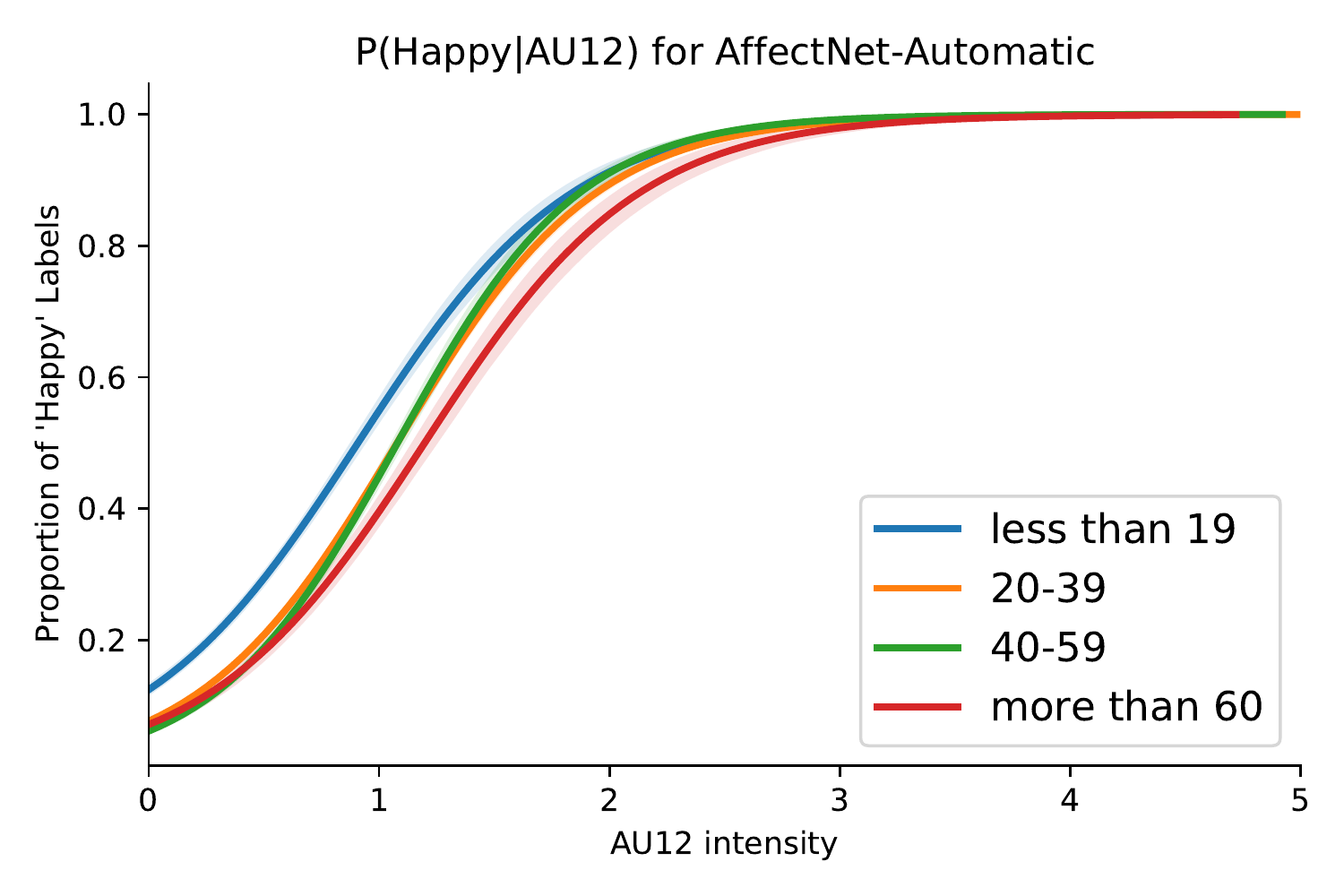}
\caption{Annotation bias of the ``happy'' expression across 4 age groups for each in-the-wild expression dataset. The first row shows the proportions of "happy" labels for each of the 4 age groups conditioned on AU6 and AU12 presence variables. The error bars indicate one standard error of the proportion. The second and third rows show the fitted logistic regression curves as a function of AU intensities. The shaded regions indicate 95\% confidence intervals. AU6 exhibits a larger bias than AU12.}
\label{fig:happy_annotation_bias_4age}
\end{figure*}

\begin{figure*}[t]
\centering

\includegraphics[width=0.23\textwidth]{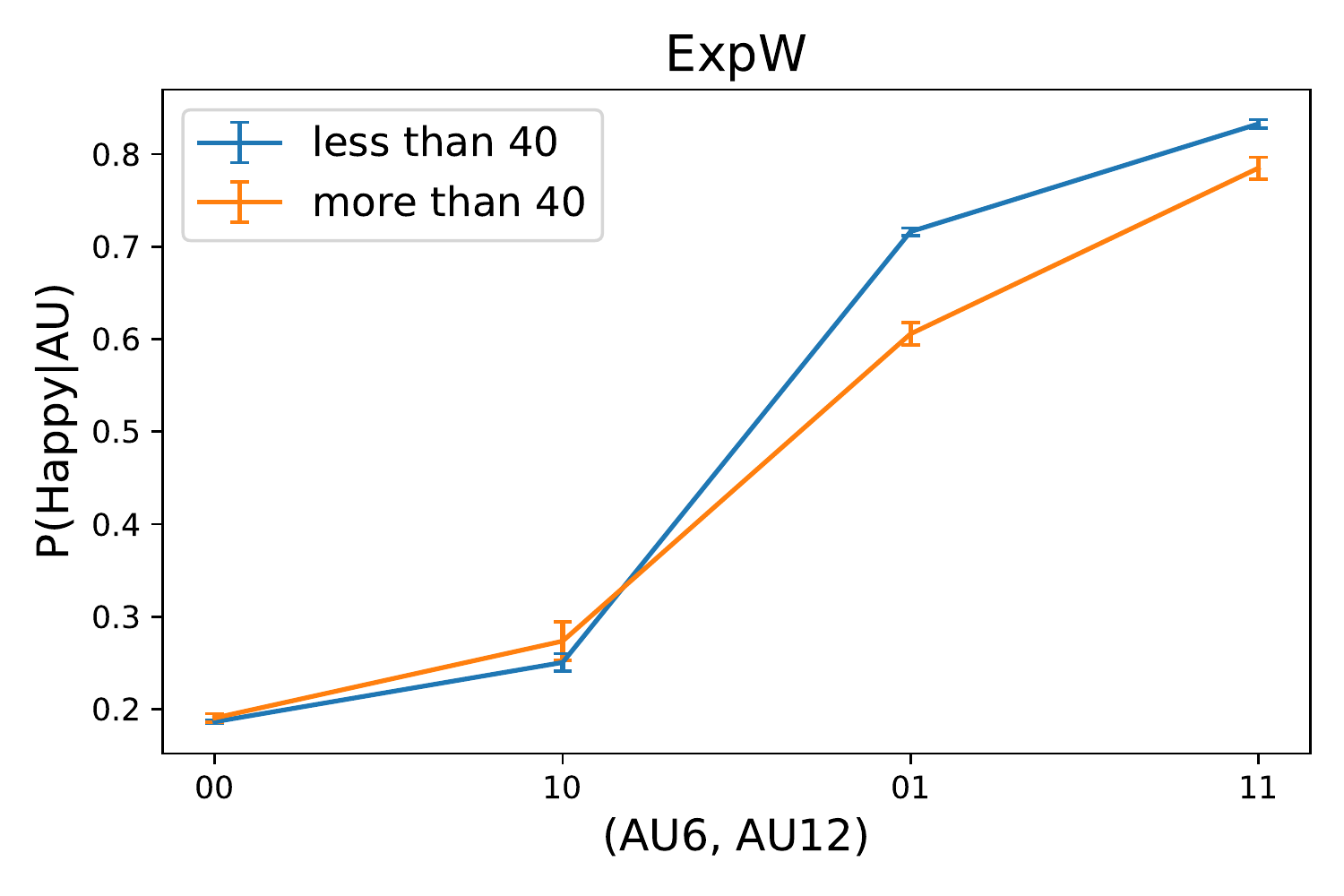}
\includegraphics[width=0.23\textwidth]{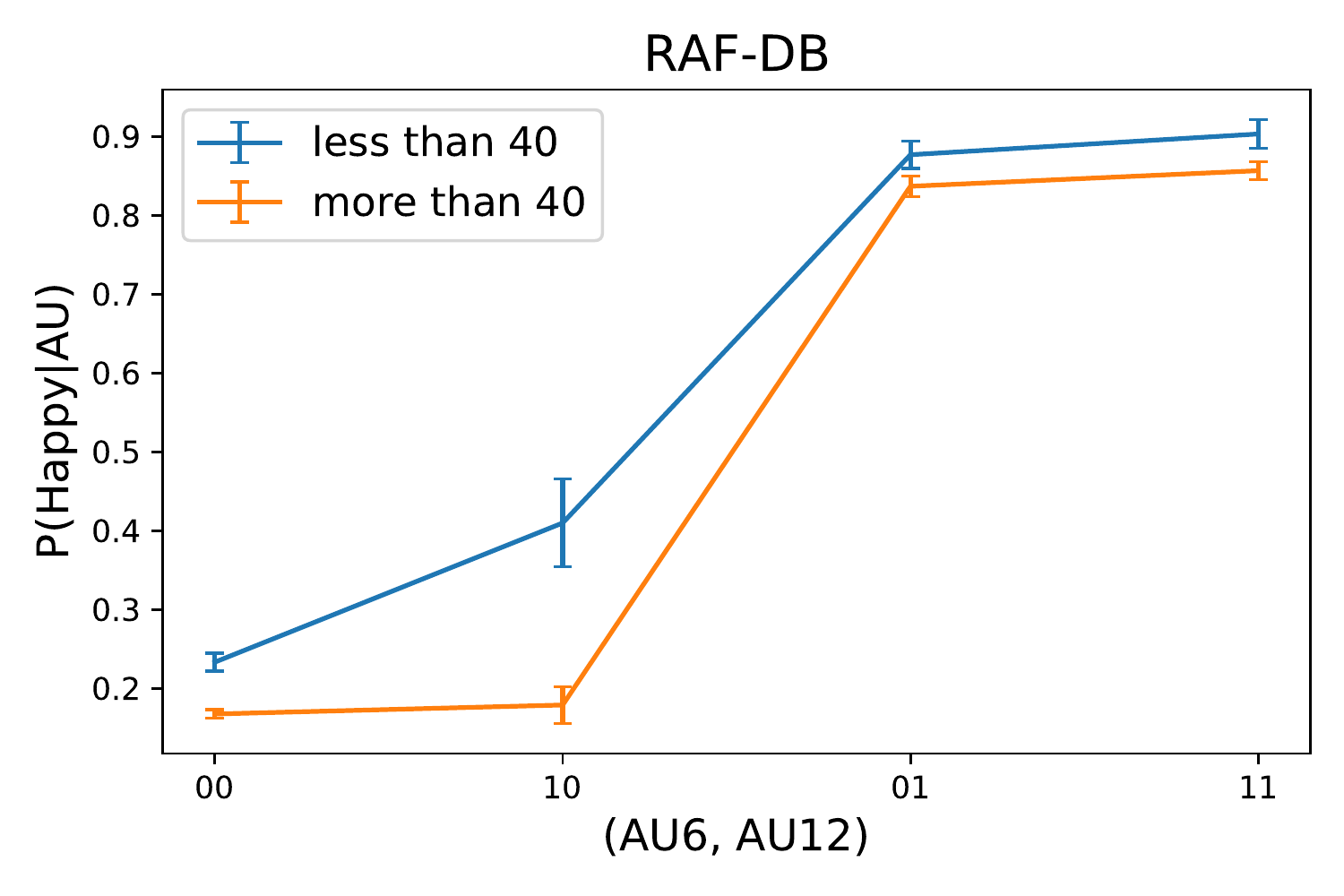}
\includegraphics[width=0.23\textwidth]{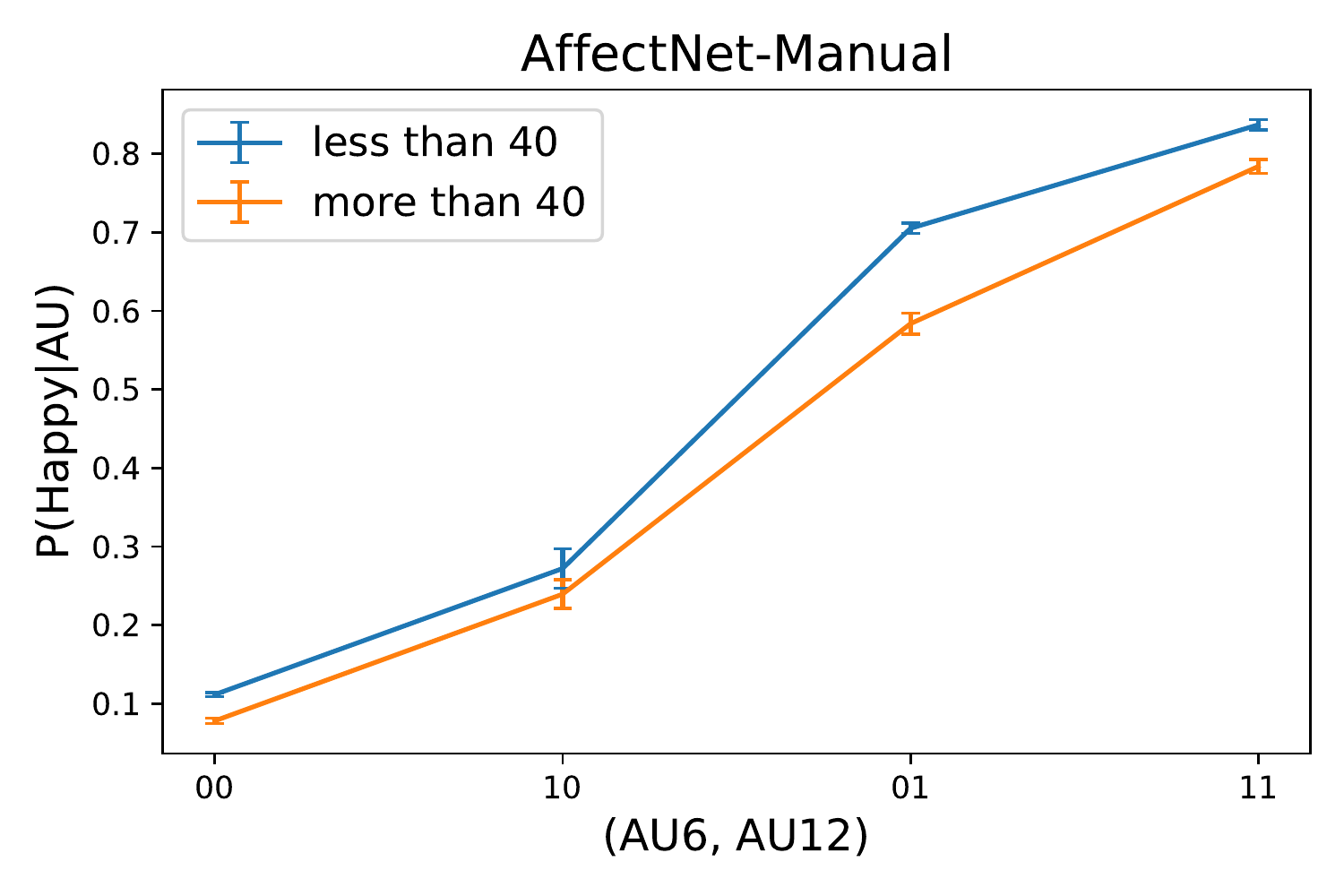}
\includegraphics[width=0.23\textwidth]{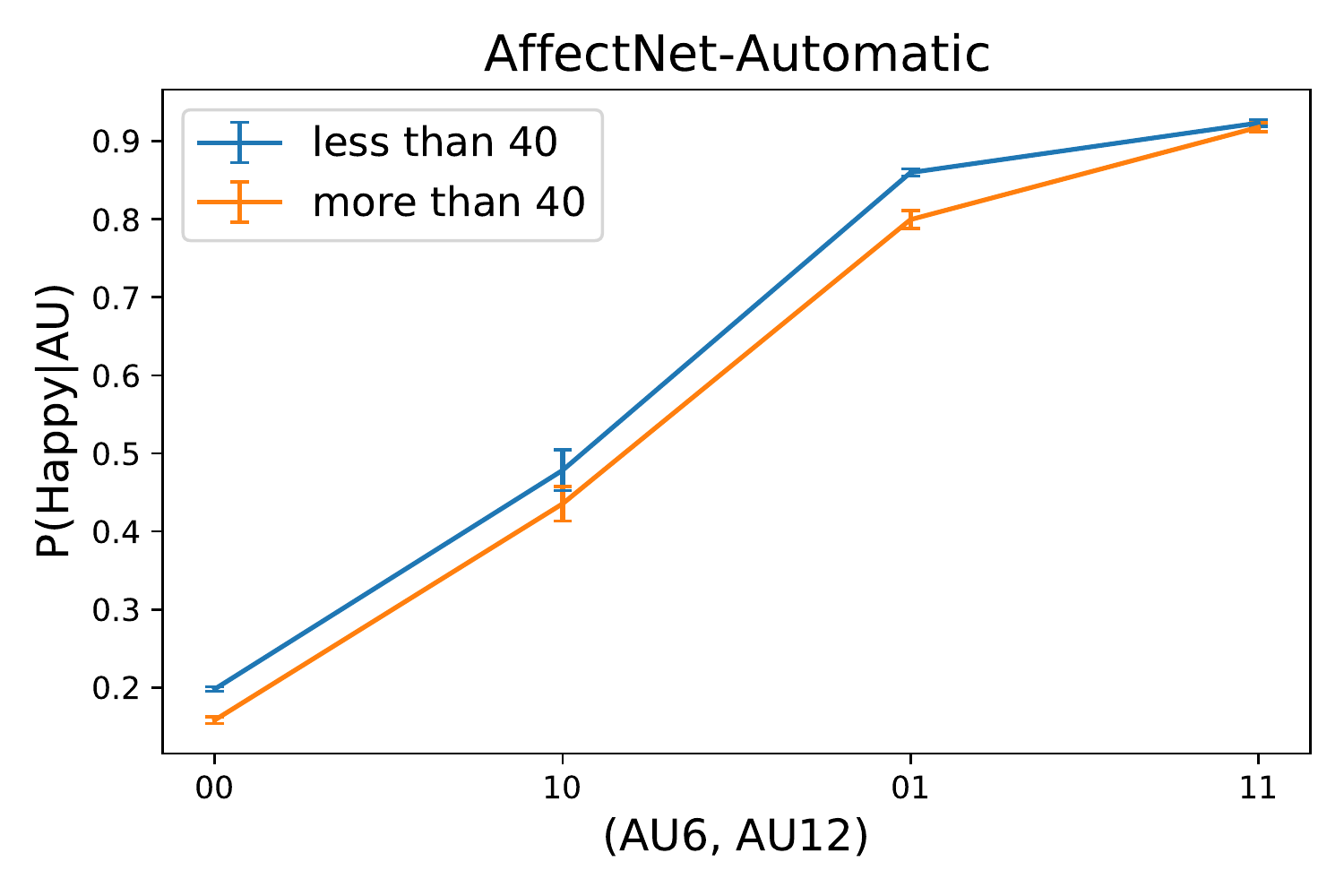}
\includegraphics[width=0.23\textwidth]{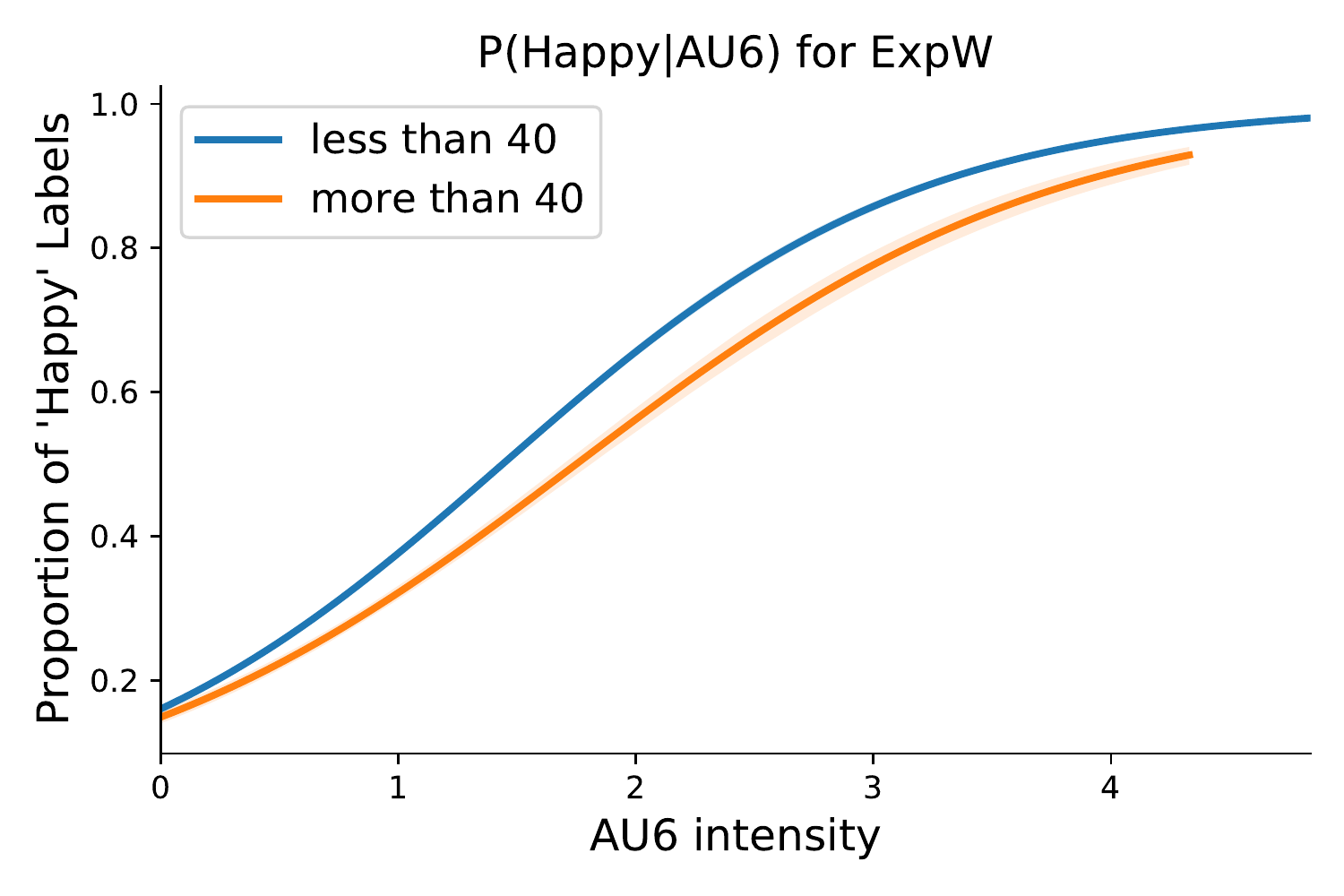}
\includegraphics[width=0.23\textwidth]{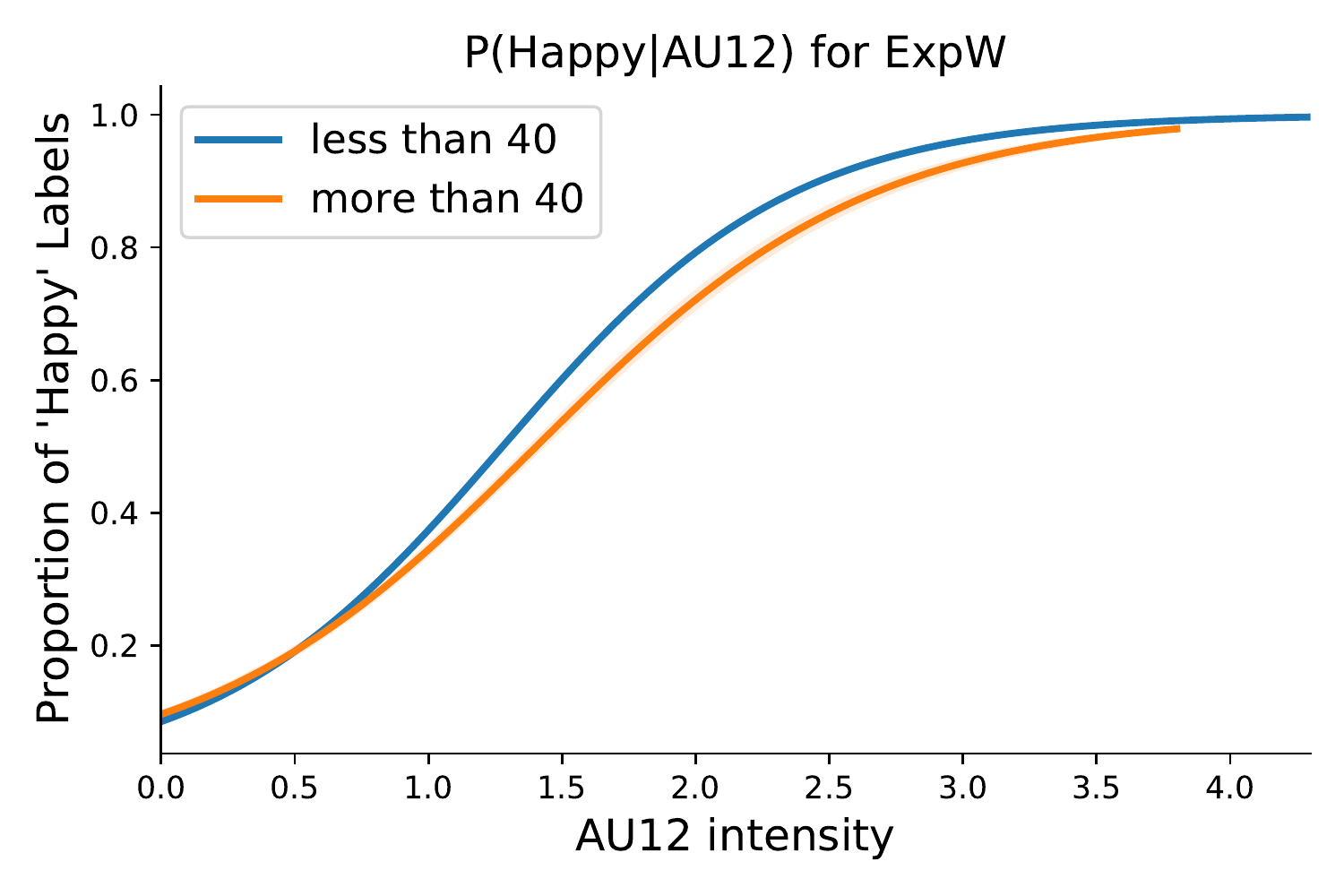}
\includegraphics[width=0.23\textwidth]{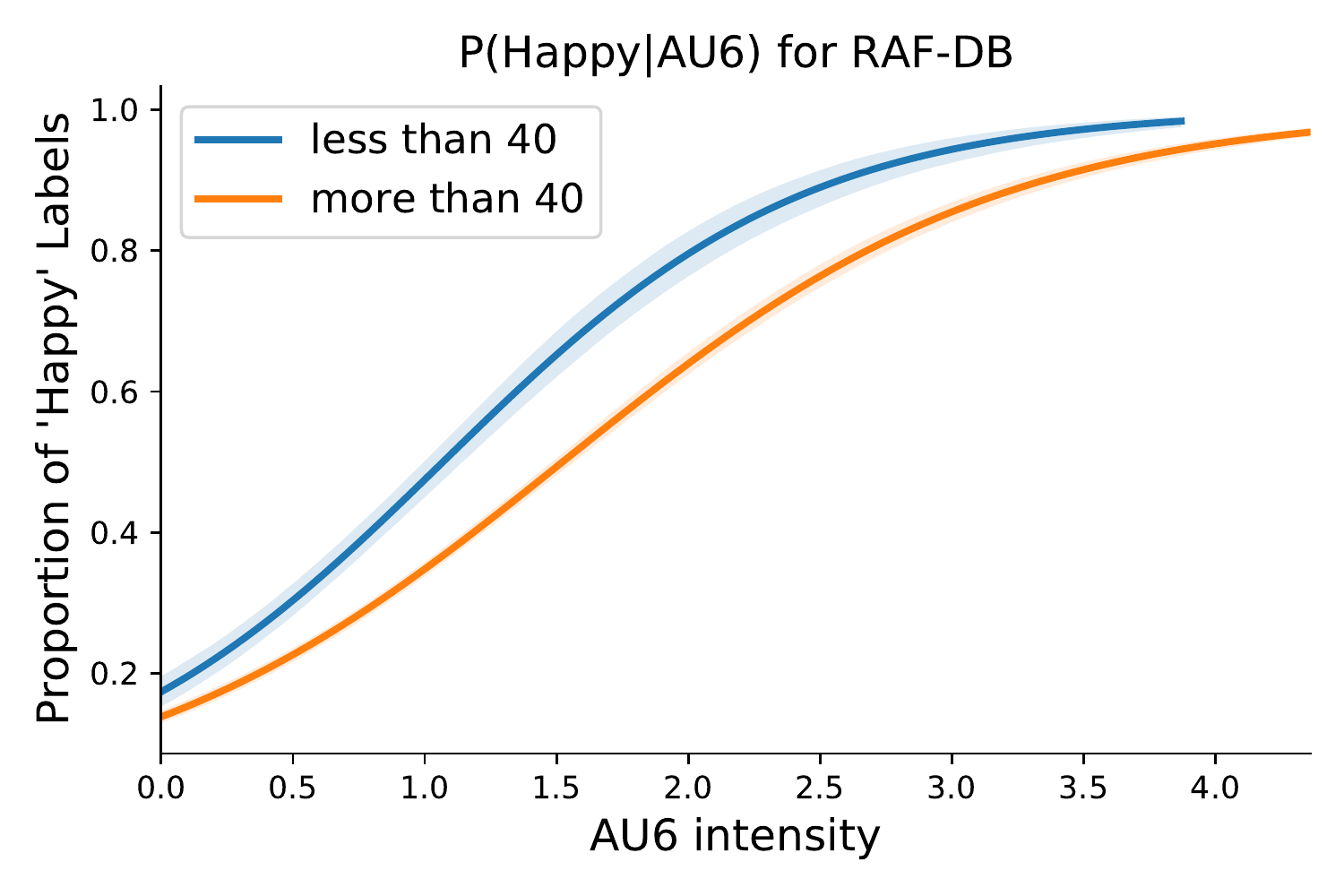}
\includegraphics[width=0.23\textwidth]{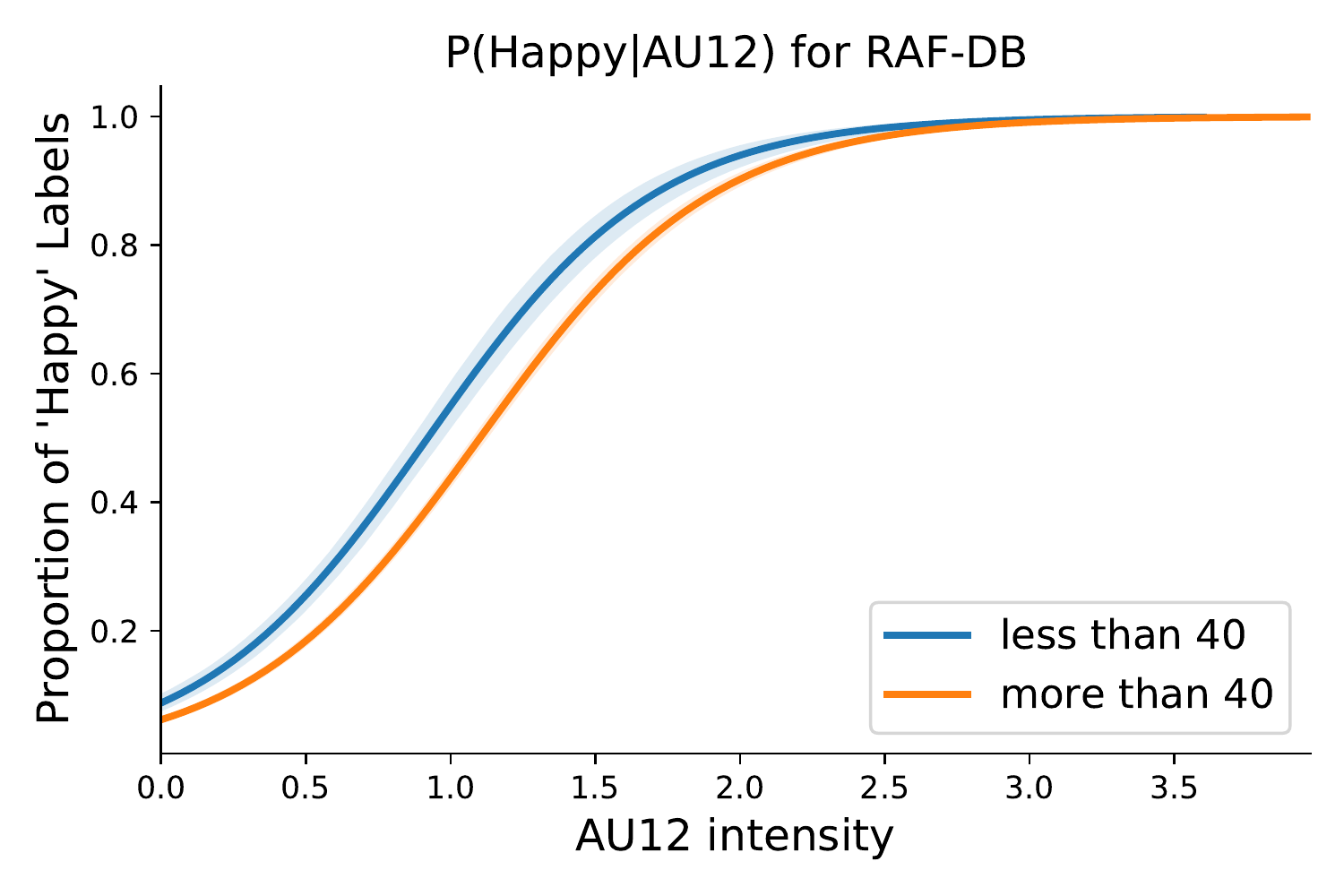}
\includegraphics[width=0.23\textwidth]{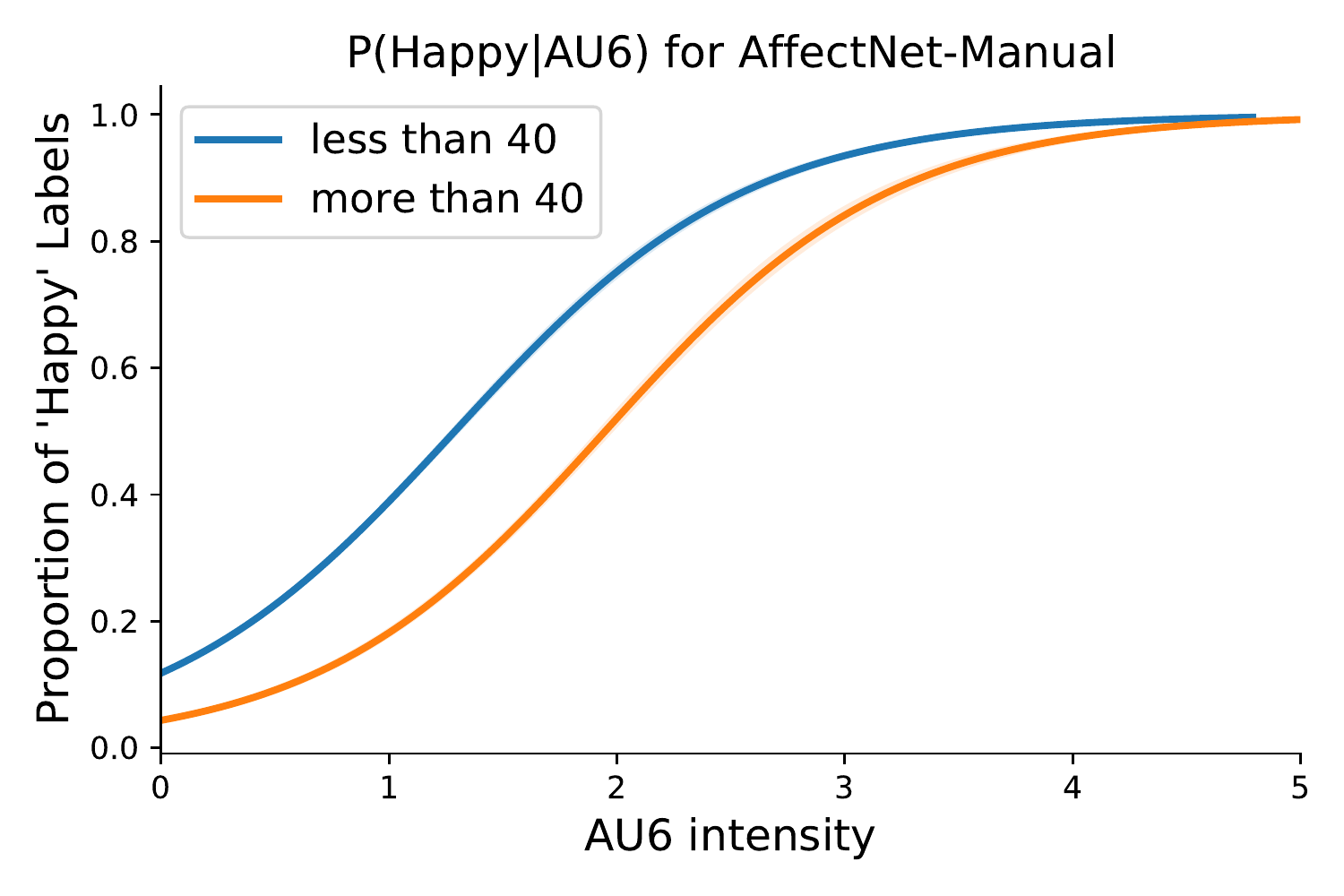}
\includegraphics[width=0.23\textwidth]{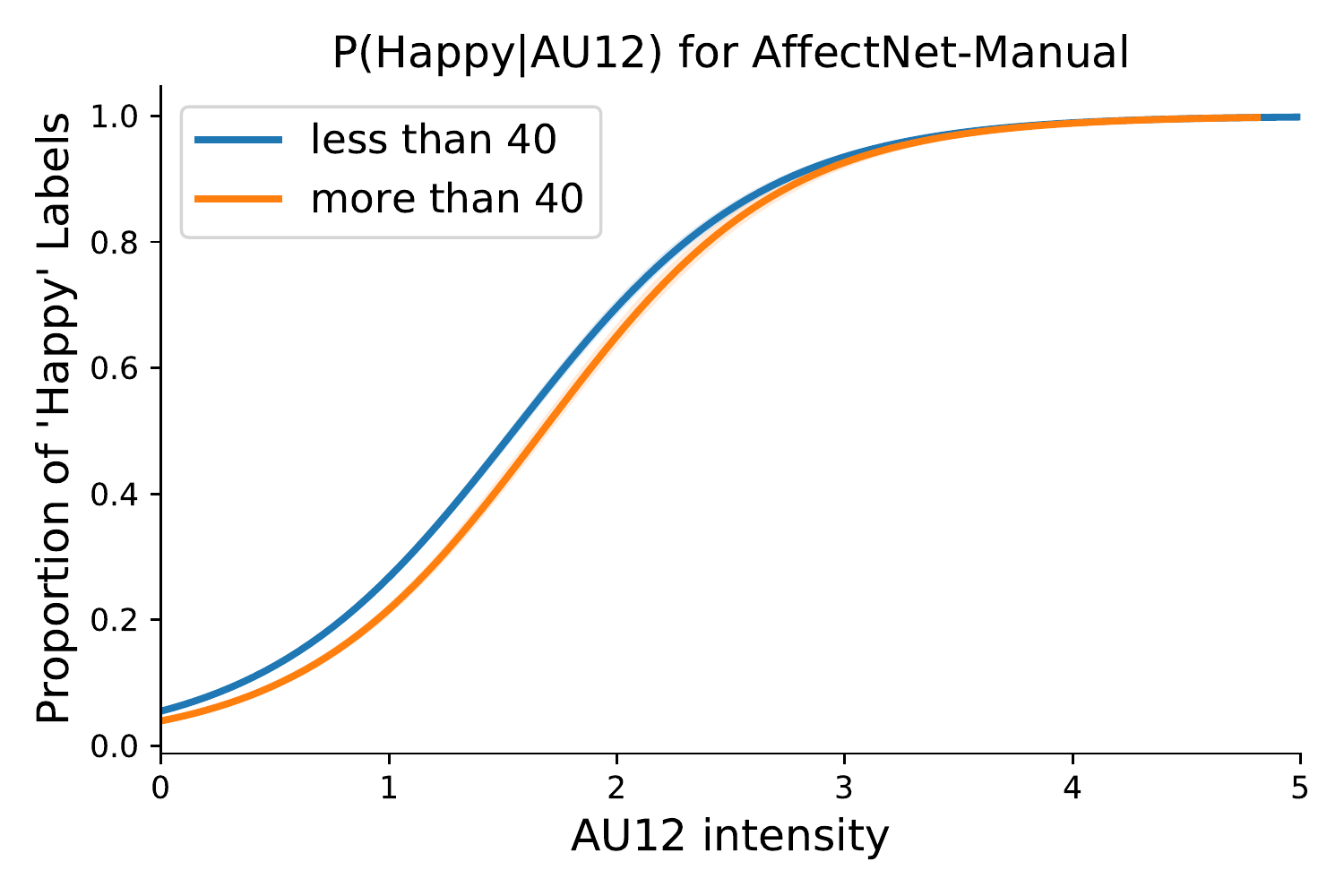}
\includegraphics[width=0.23\textwidth]{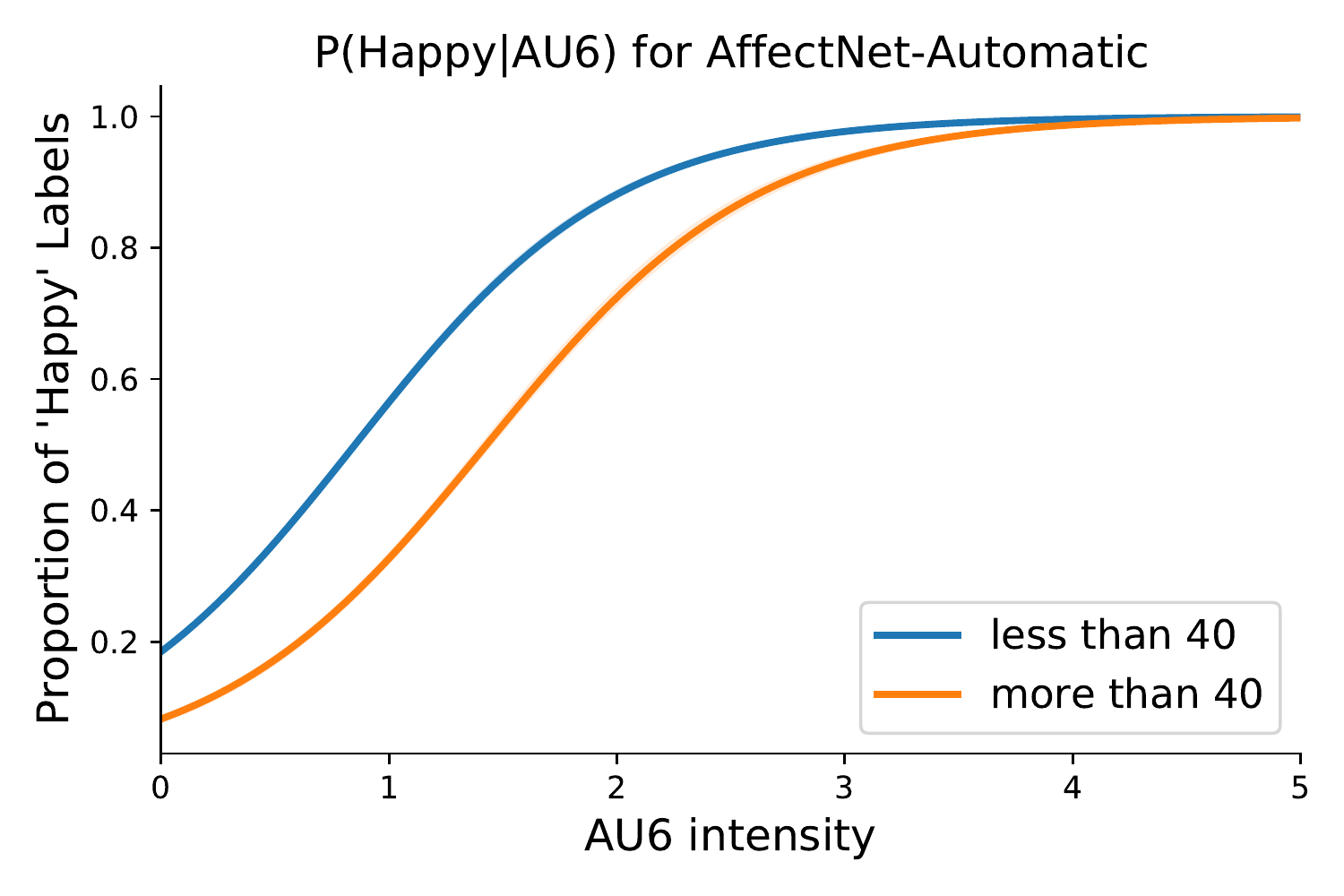}
\includegraphics[width=0.23\textwidth]{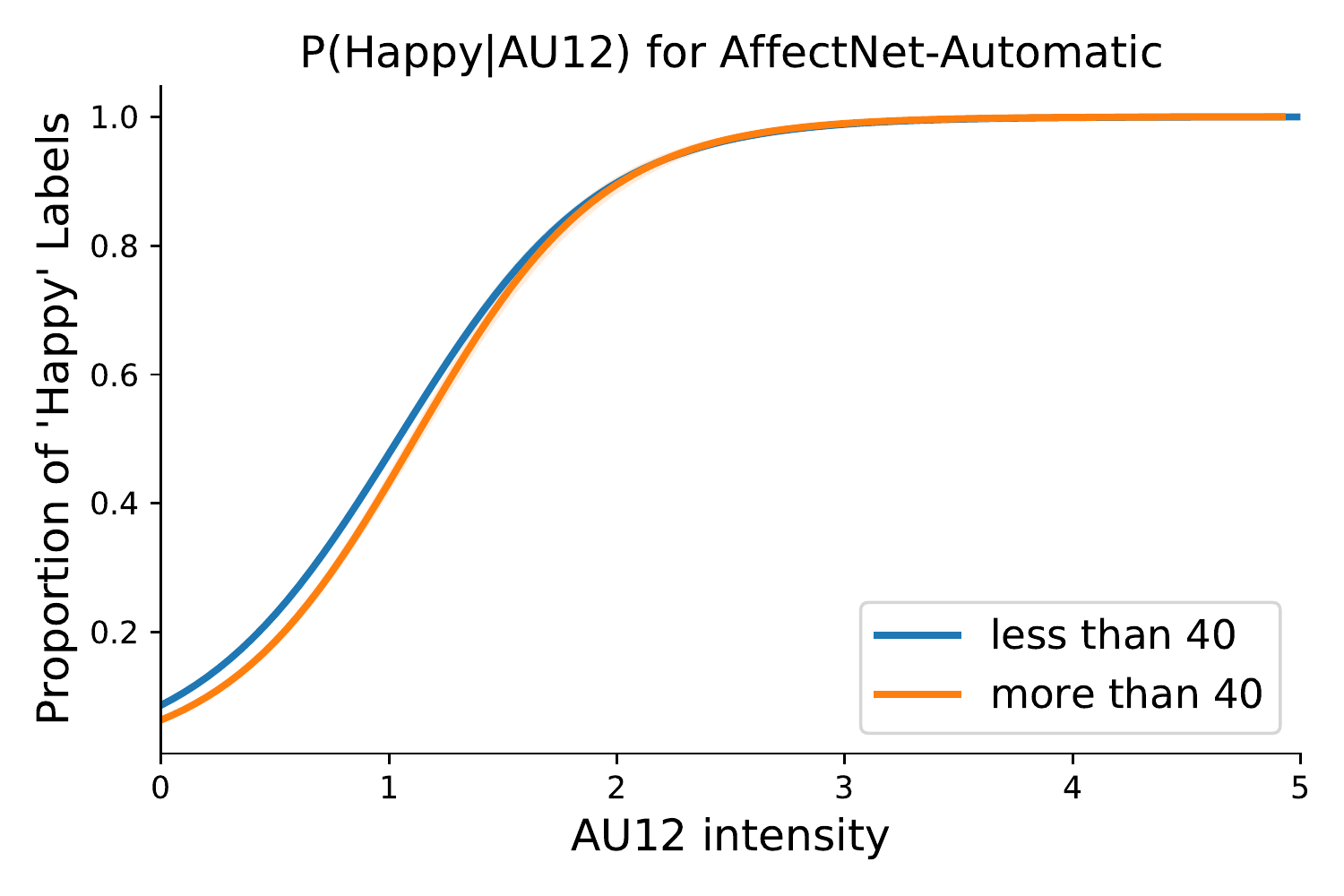}
\caption{Annotation bias of the ``happy'' expression between younger and older populations. The first row shows the proportions of "happy" labels conditioned on AU6 and AU12 presence variables. The error bars indicate one standard error of the proportion. The second and third rows show the fitted logistic regression curves as a function of AU intensities. The shaded regions indicate 95\% confidence intervals. Younger people seems more likely to be annotated as ``happy'' compared to older people.}
\label{fig:happy_annotation_bias_2age}
\end{figure*}

Table \ref{tab:Anno_bias_happy_age_summary} shows the conditional and marginal distributions of ``happy'' labels along with the numbers of samples for each age group for each in-the-wild expression dataset. We see that all datasets are heavily dominated by younger people. The sum of the numbers for each age group does not add up to those of the full datasets due to the fact that OpenFace fails to produce AU labels (possibly due to occlusion or blur) for a small fraction of the images. The p-values are the $\chi^2$ tests for independence of the ``happy'' labels and the age groups. When the p-values are significant at the 0.05 level, the age groups with the highest proportion of "happy" labels are highlighted. We can see that the differences in the proportion of ``happy'' labels are statistically significant in most cases. Moreover, younger age groups have a higher proportion of ``happy'' labels in general, suggesting the existence of systematic annotation bias. 

Figure \ref{fig:happy_annotation_bias_4age} plots the annotation bias of the ``happy'' expression across the 4 age groups. The first row shows the proportions of ``happy'' labels for each of the 4 age groups conditioned on AU6 and AU12 presence variables. The error bars indicate one standard error of the proportion. The second and third rows show the fitted logistic regression curves as a function of AU intensities. The shaded regions indicate 95\% confidence intervals. From the plots, we see that AU6 exhibits a larger bias than AU12.
Figure \ref{fig:happy_annotation_bias_2age} compares younger and older populations. Consistent with Table \ref{tab:Anno_bias_happy_age_summary}, we see that younger people are more likely to be annotated as ``happy'' compared to older people, although the saliency varies across datasets. Further analysis on more balanced datasets (\ie, datasets that have more older-than-40 and especially older-than-60 people) is needed.

\section{Happy Annotation Bias Across Racial Groups}

\begin{table*}[h]
\small
\centering
\begin{threeparttable}\scalebox{0.97}{
\begin{tabular*}{1\textwidth}{|m{0.1\linewidth}||m{0.17\linewidth}|m{0.045\linewidth}|m{0.045\linewidth}|m{0.045\linewidth}|m{0.075\linewidth}||m{0.045\linewidth}|m{0.065\linewidth}|m{0.06\linewidth}|m{0.075\linewidth}|}
\cline{1-10}
\multirow{2}{\linewidth}{\parbox{1\linewidth}{\vspace{0.5cm} \centering Data}} &\multirow{2}{\linewidth}{\parbox{1\linewidth}{\vspace{0.3cm} \centering Metrics \\ P(Happy\textbar{}AU6, AU12)}} & \multicolumn{4}{c||}{3 Groups} & \multicolumn{4}{c|}{6 Groups} \\ 
\cline{3-10}
             &                   & Asian          & Black          & White          &  \multicolumn{1}{c||}{p-value} & Indian & Latino-Hispanic & Middle Eastern &  \multicolumn{1}{c|}{p-value}  \\
\cline{1-10}
\multirow{6}{\linewidth}{\centering ExpW \cite{SOCIALRELATION_ICCV2015,SOCIALRELATION_2017}}                 & P(Happy\textbar{}(0, 0)) & 0.177          & 0.181          & \textbf{0.194} & 0.000 ***              & 0.242  & 0.21            & 0.177          & 0.000 ***               \\
                    & P(Happy\textbar{}(0, 1)) & 0.697          & \textbf{0.723} & 0.678          & 0.004 **           & 0.75   & 0.763           & 0.685          & 0.000 ***               \\
                    & P(Happy\textbar{}(1, 0)) & 0.253          & 0.267          & 0.243          & 0.67               & 0.327  & 0.275           & 0.235          & 0.654               \\
                    & P(Happy\textbar{}(1, 1)) & 0.815          & 0.833          & 0.81           & 0.252              & 0.882  & 0.88            & 0.798          & 0.000 ***               \\ 
\cline{2-10}
                    & P(Happy)                 & 0.319          & 0.323          & 0.328          &                    & 0.406  & 0.398           & 0.302          &                     \\ 
\cline{2-10}
                    & Number of samples        & 31,791          & 11,381          & 25,826          &                    & 1,701   & 8,751            & 5,905           &                     \\ 
\hhline{|=::=====::====|}
\multirow{6}{\linewidth}{\centering RAF-DB \cite{li2017reliable}}                & P(Happy\textbar{}(0, 0)) & 0.2            & \textbf{0.249} & 0.187          & 0.000 ***              &        &                 &                &                     \\
                    & P(Happy\textbar{}(0, 1)) & 0.883          & 0.833          & 0.847          & 0.286              &        &                 &                &                     \\
                    & P(Happy\textbar{}(1, 0)) & \textbf{0.343} & 0.244          & 0.216          & 0.023 *            &        &                 &                &                     \\
                    & P(Happy\textbar{}(1, 1)) & 0.873          & 0.919          & 0.862          & 0.177              &        &                 &                &                     \\ 
\cline{2-6}
                    & P(Happy)                 & 0.403          & 0.421          & 0.375          &                    &        &                 &                &                     \\ 
\cline{2-6}
                    & Number of samples        & 1,904           & 1,011           & 9,299           &                    &        &                 &                &                     \\ 
\hhline{|=::=====::====|}
\multirow{6}{\linewidth}{\centering AffectNet-Manual \cite{mollahosseini2017affectNet}}     & P(Happy\textbar{}(0, 0)) & 0.094          & 0.083          & \textbf{0.106} & 0.005 **           & 0.099  & 0.104           & 0.083          & 0.003 **            \\
                    & P(Happy\textbar{}(0, 1)) & 0.657          & 0.693          & 0.683          & 0.499              & 0.68   & 0.68            & 0.627          & 0.228               \\
                    & P(Happy\textbar{}(1, 0)) & 0.208          & 0.275          & 0.266          & 0.663              & 0.229  & 0.212           & 0.19           & 0.654               \\
                    & P(Happy\textbar{}(1, 1)) & 0.781          & 0.838          & 0.821          & 0.083 .            & 0.817  & 0.826           & 0.718          & 0.004 **            \\ 
\cline{2-10}
                    & P(Happy)                 & 0.311          & 0.320          & 0.320          &                    & 0.312  & 0.321           & 0.222          &                     \\ 
\cline{2-10}
                    & Number of samples        & 2,211           & 2,833           & 23,780          &                    & 1,159   & 3,179            & 2,532           &                     \\ 
\hhline{|=::=====::====|}
\multirow{6}{\linewidth}{\centering AffectNet-Automatic \cite{mollahosseini2017affectNet}} & P(Happy\textbar{}(0, 0)) & \textbf{0.207} & 0.154          & 0.193          & 0.000 ***              & 0.176  & 0.204           & 0.153          & 0.000 ***               \\
                    & P(Happy\textbar{}(0, 1)) & 0.828          & 0.847          & 0.853          & 0.254              & 0.837  & 0.885           & 0.784          & 0.002 **            \\
                    & P(Happy\textbar{}(1, 0)) & 0.436          & 0.435          & 0.473          & 0.709              & 0.262  & 0.483           & 0.413          & 0.164               \\
                    & P(Happy\textbar{}(1, 1)) & 0.874          & 0.918          & \textbf{0.930} & 0.000 ***              & 0.917  & 0.932           & 0.874          & 0.000 ***               \\ 
\cline{2-10}
                    & P(Happy)                 & 0.427          & 0.391          & 0.430          &                    & 0.403  & 0.453           & 0.281          &                     \\ 
\cline{2-10}
                    & Number of samples        & 3,619           & 4,454           & 26,798          &                    & 1,548   & 3,198            & 2,667           &                     \\
\cline{1-10}
\end{tabular*}
}
\begin{tablenotes}
\item[] Signif. codes:  0 ‘***’ 0.001 ‘**’ 0.01 ‘*’ 0.05 ‘.’ 0.1 ‘ ’ 1
\item[1] The provided labels from RAF-DB are ``Asian,'' ``White,'' and ``Black'' only.
\end{tablenotes}
\end{threeparttable}
\caption{Conditional and marginal distributions of ``happy'' labels along with the numbers of samples for each racial/ethnic group for each in-the-wild dataset. The p-values are the $\chi^2$ tests for independence of the "happy" labels and the racial/ethnic groups. When the p-values for the 3 racial groups are significant at the 0.05 level, the racial groups with the highest proportion of ``happy'' labels are highlighted.}
\label{tab:Anno_bias_happy_race_summary}
\end{table*}

In this section, we describe the evaluation of annotation bias for the happy expression across racial groups. Similar to the issue of imbalanced classes we encounter with age groups, many of the in-the-wild datasets (with the exception of ExpW) are severely dominated by white people, making the statistical tests challenging. Again, lab-controlled datasets are too small and lack diversity in their populations (\eg, CFD has only ~100 images for Asian and Latino groups each), and so we only focus on in-the-wild datasets.

Since only RAF-DB includes race labels, we first train a simple race classifier with ResNet-34 architecture \cite{he2016deep} using the FairFace dataset \cite{karkkainen2021fairface} similar to the procedure for training the gender and age classifiers. We follow the convention of the race/ethnicity categorization of FairFace, which classifies each image into one of seven groups: White, Black, Latino/Hispanic, East Asian, Southeast Asian, Indian, and Middle Eastern. We then apply the trained classifier on the datasets that do not have age labels (\ie, KDEF, CFD, ExpW, and AffectNet). 
We combine the East Asian and Southeast Asian groups together. This results in 6 racial/ethnic groups. However, since the race labels provided by RAF-DB only include White, Black, and Asian, and many datasets contain relatively few images for Indian, Middle Eastern, and Latino/Hispanic groups, we also conduct analysis that only focuses on the three major races.

\begin{figure*}[t]
\centering
\includegraphics[width=0.24\textwidth]{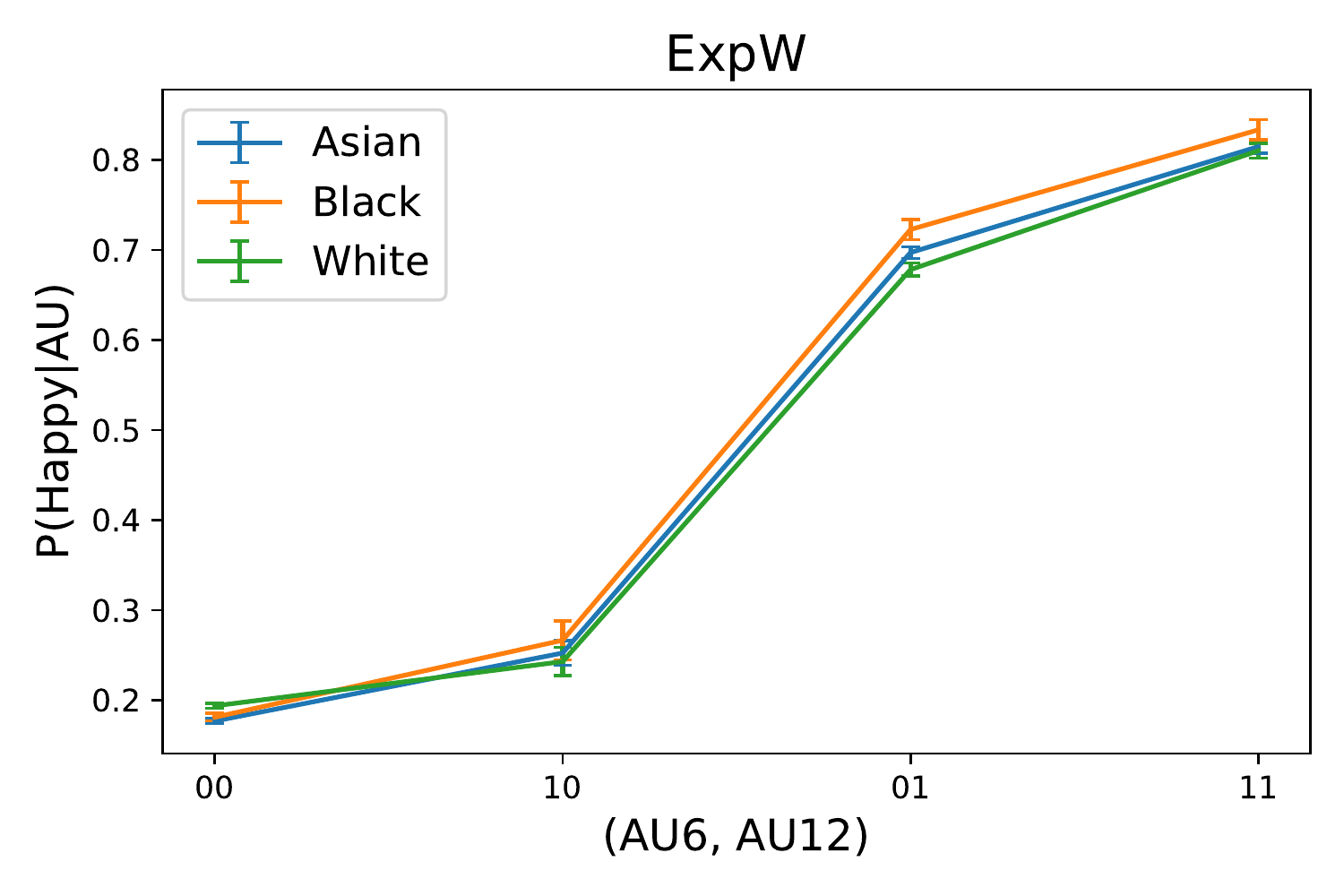}
\includegraphics[width=0.24\textwidth]{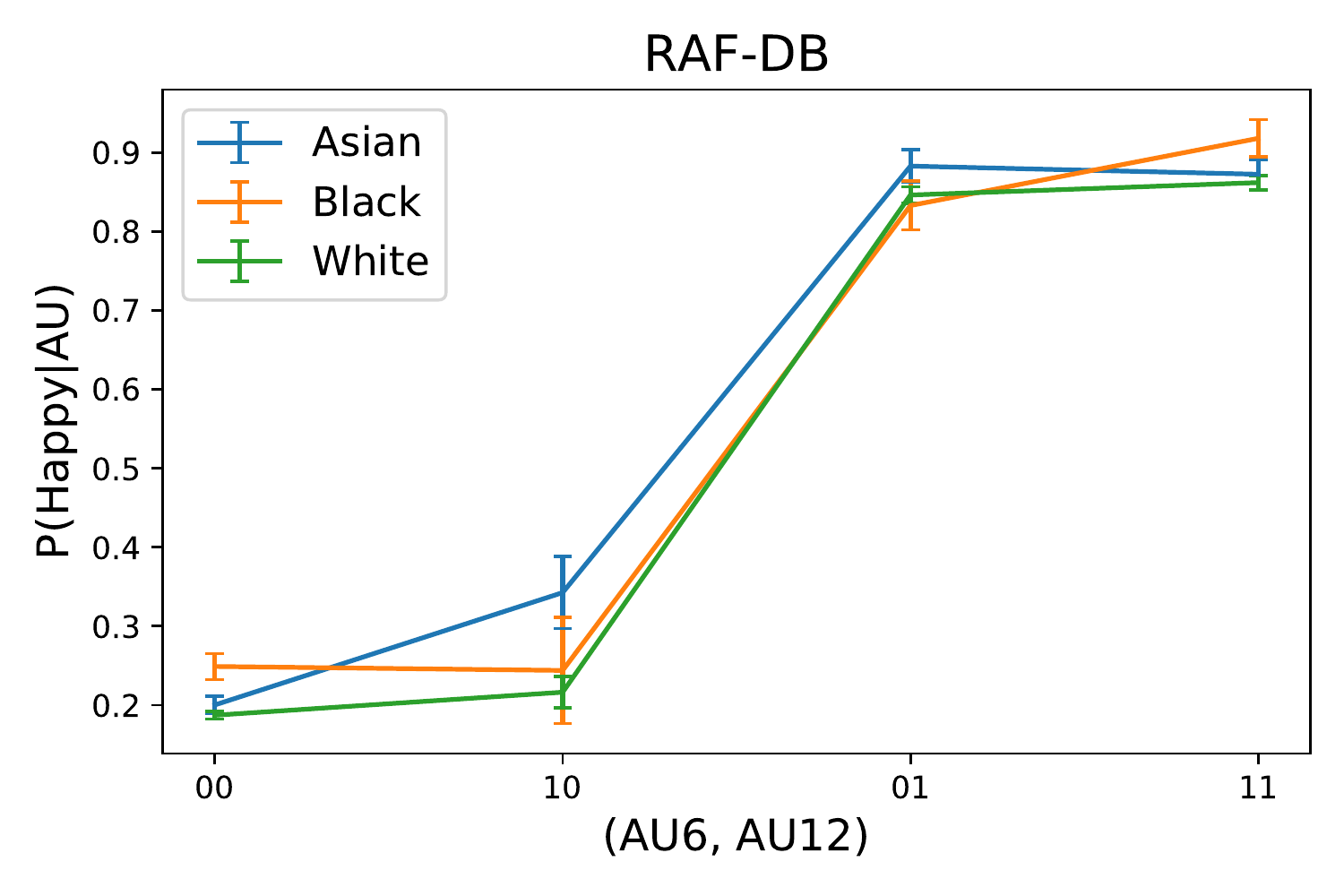}
\includegraphics[width=0.24\textwidth]{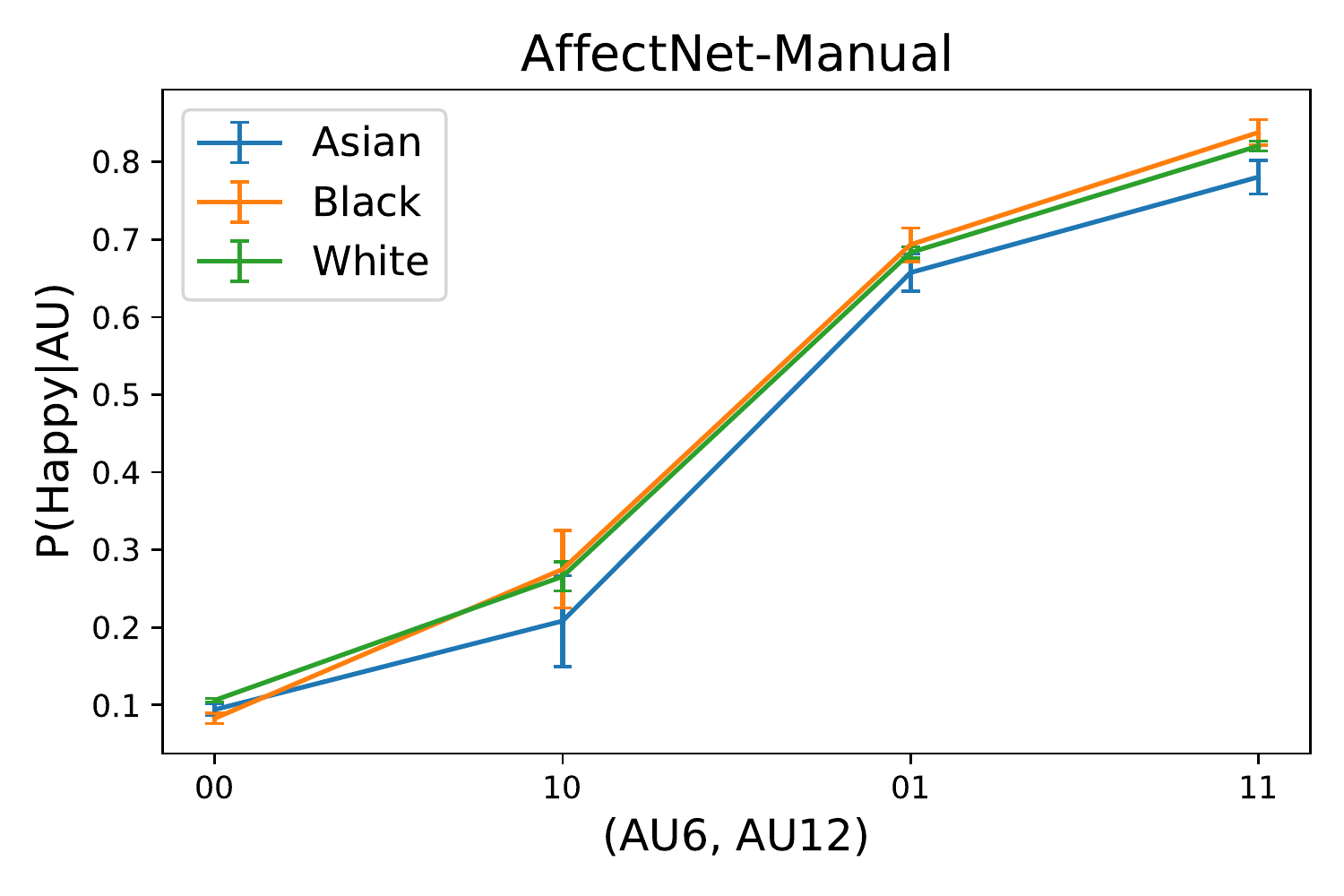}
\includegraphics[width=0.24\textwidth]{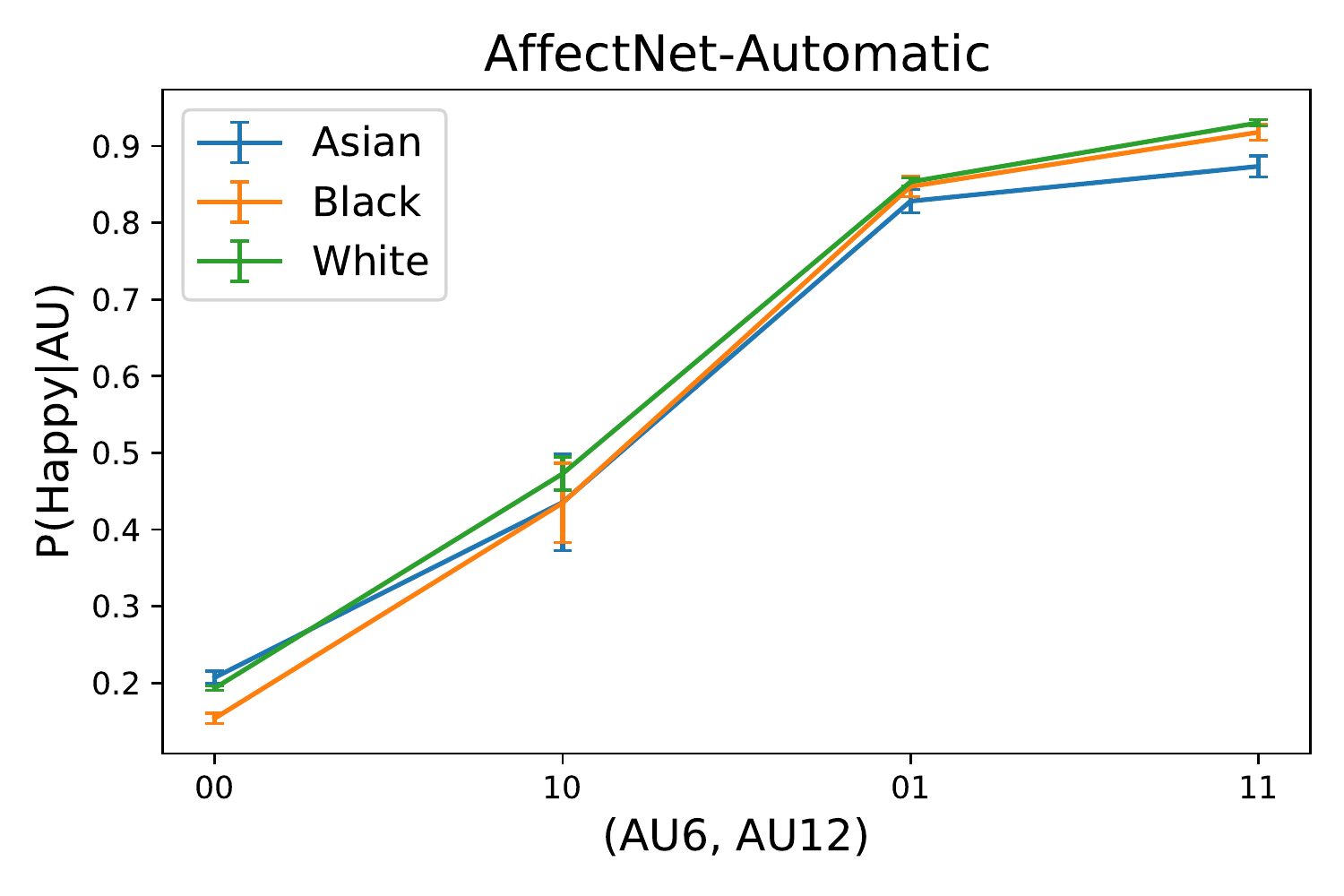}
\includegraphics[width=0.24\textwidth]{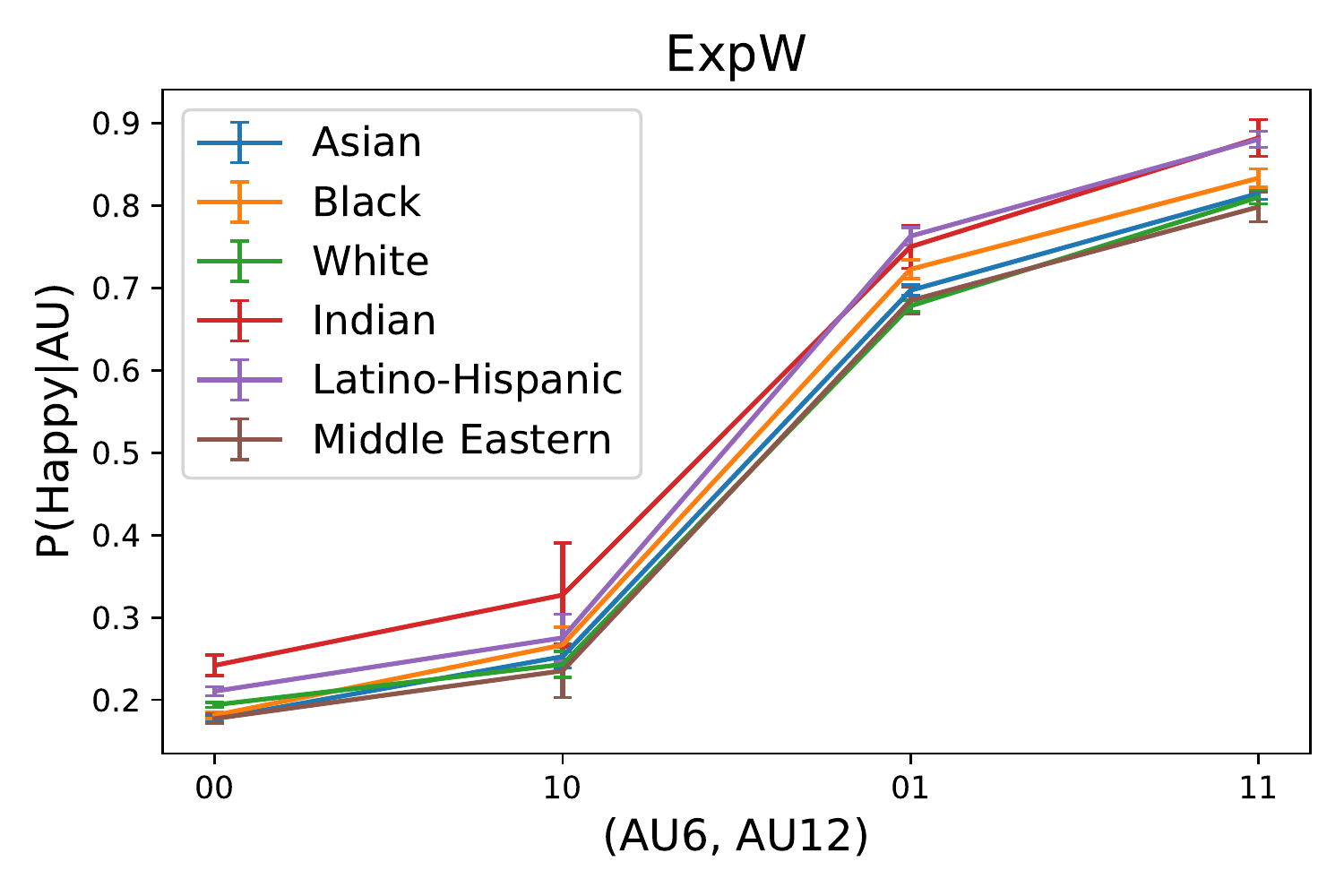}
\includegraphics[width=0.24\textwidth]{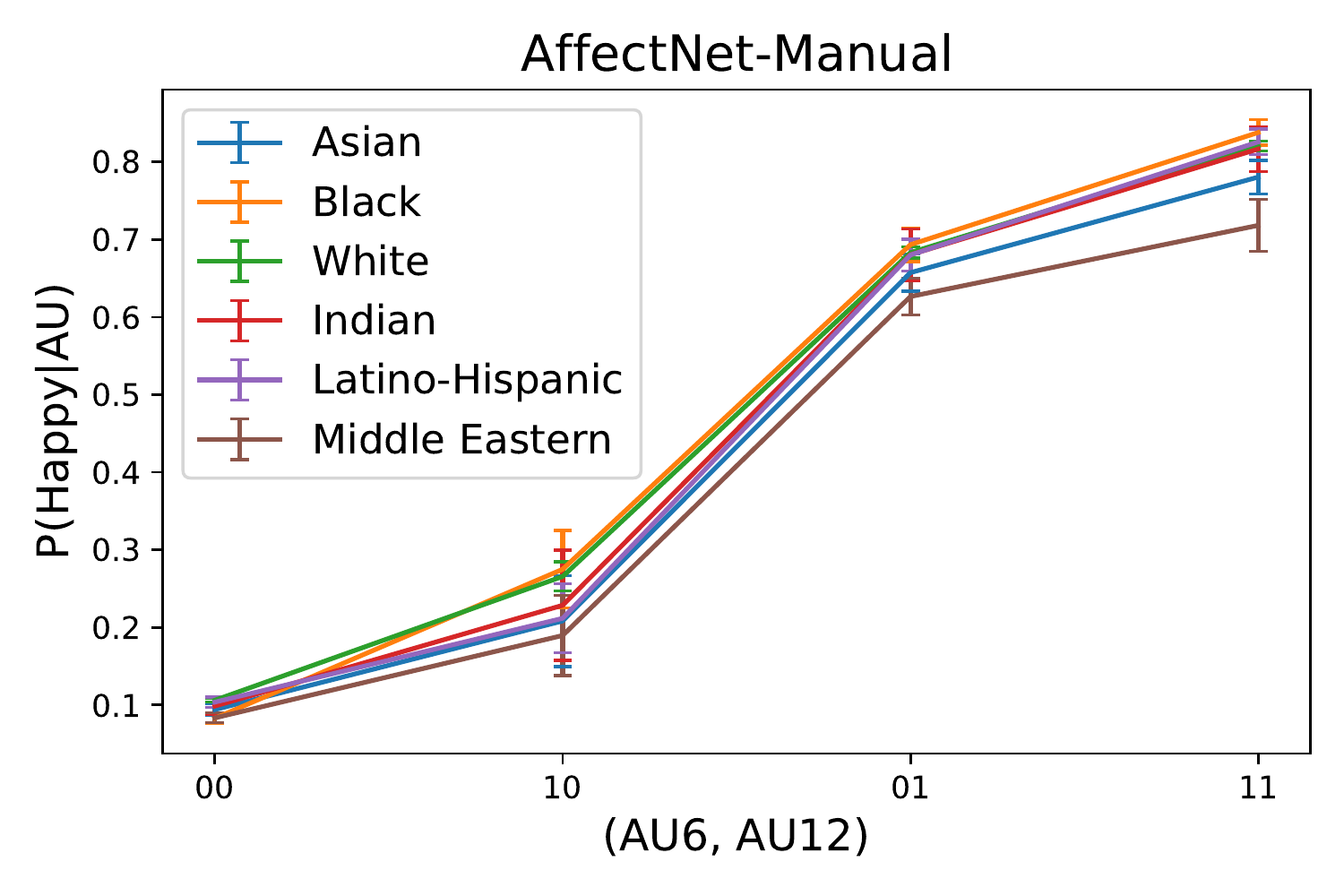}
\includegraphics[width=0.24\textwidth]{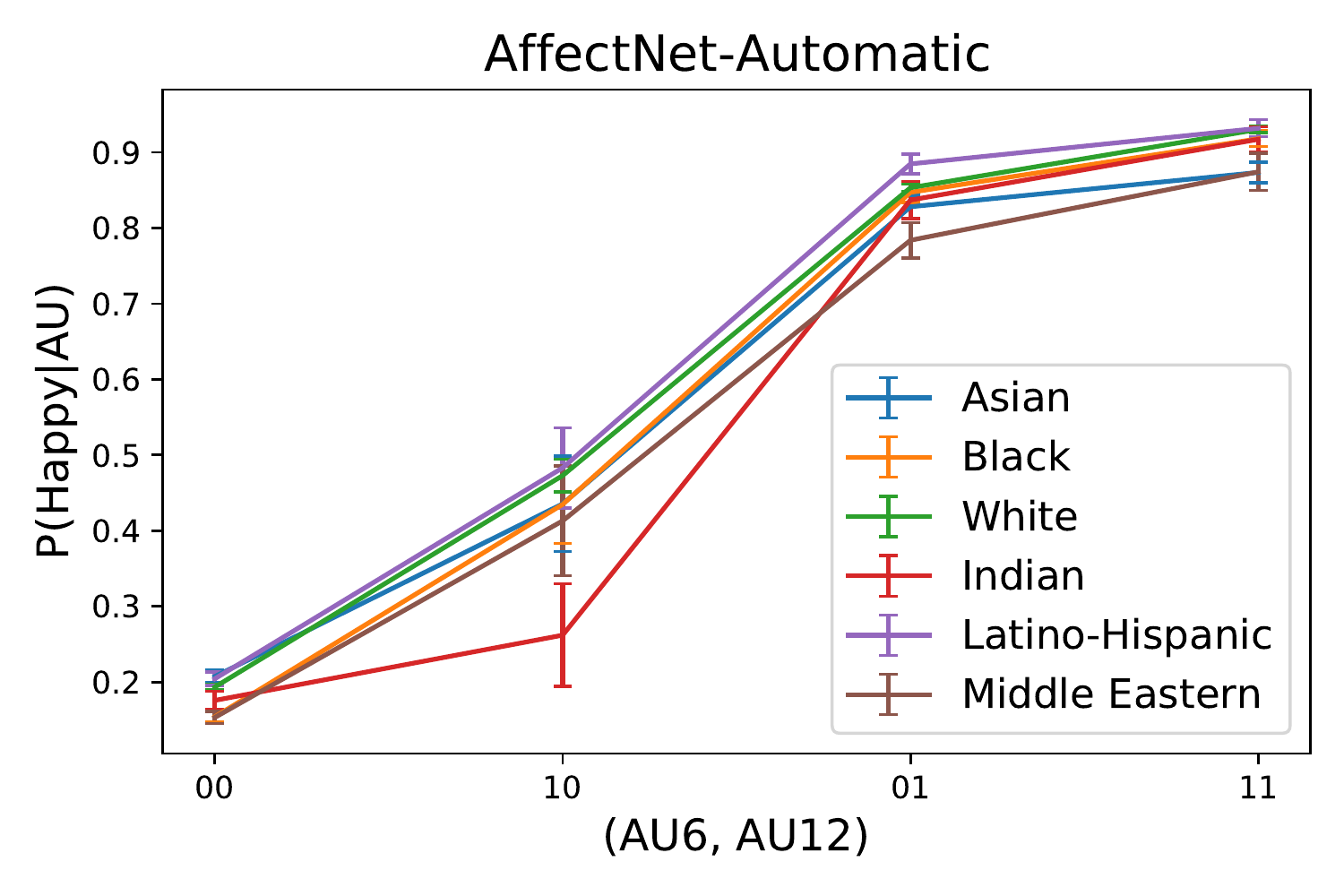}

\includegraphics[width=0.24\textwidth]{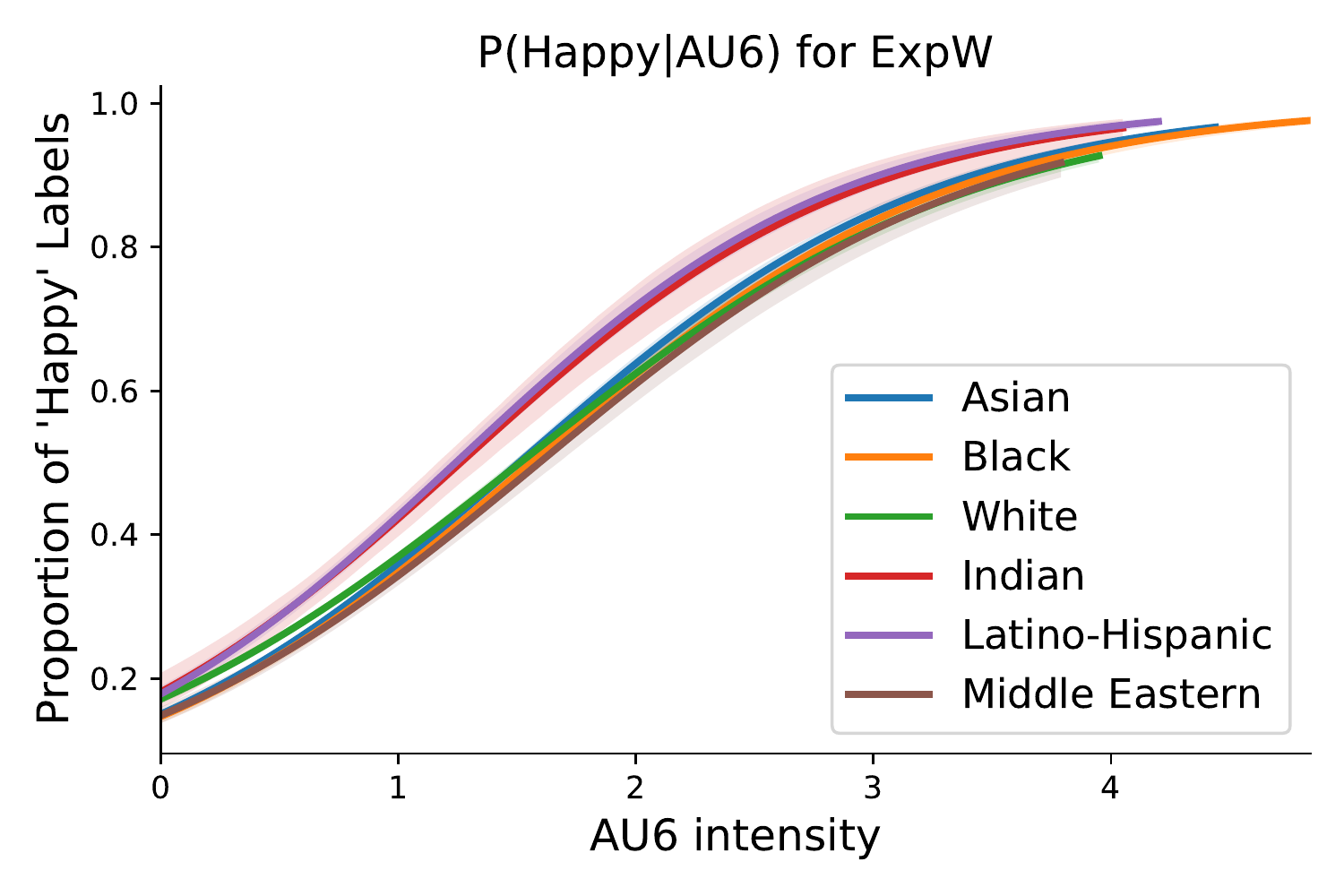}
\includegraphics[width=0.24\textwidth]{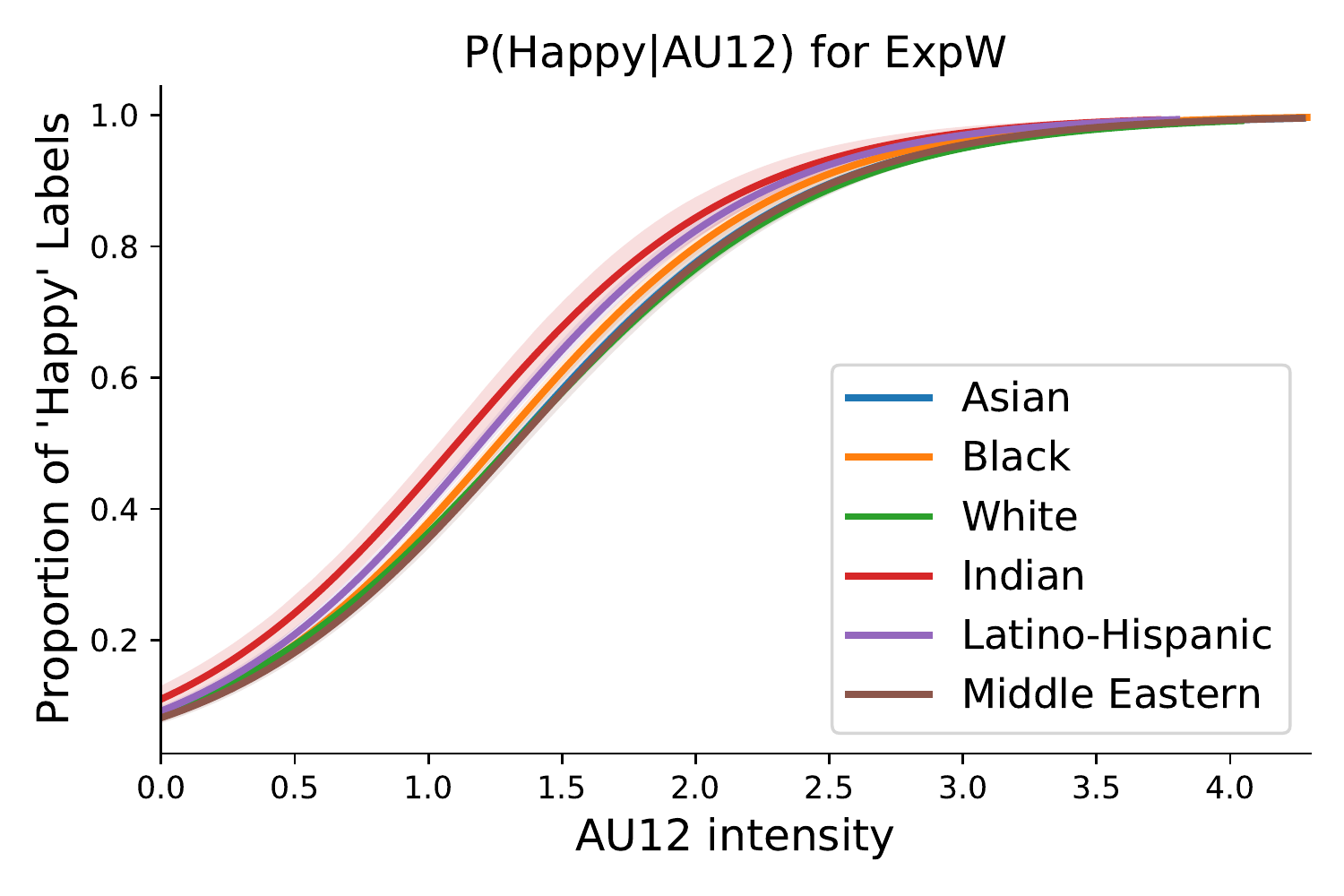}
\includegraphics[width=0.24\textwidth]{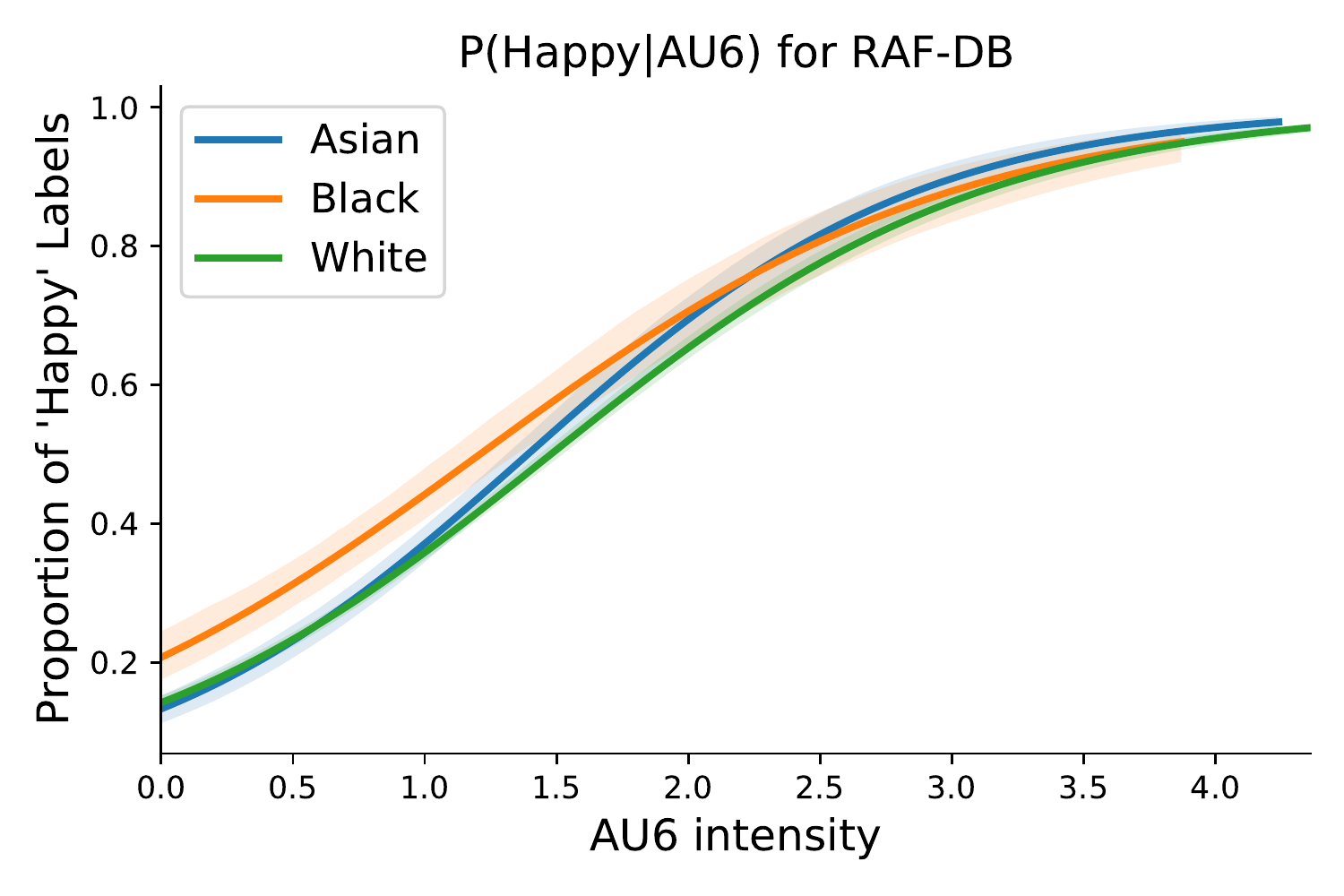}
\includegraphics[width=0.24\textwidth]{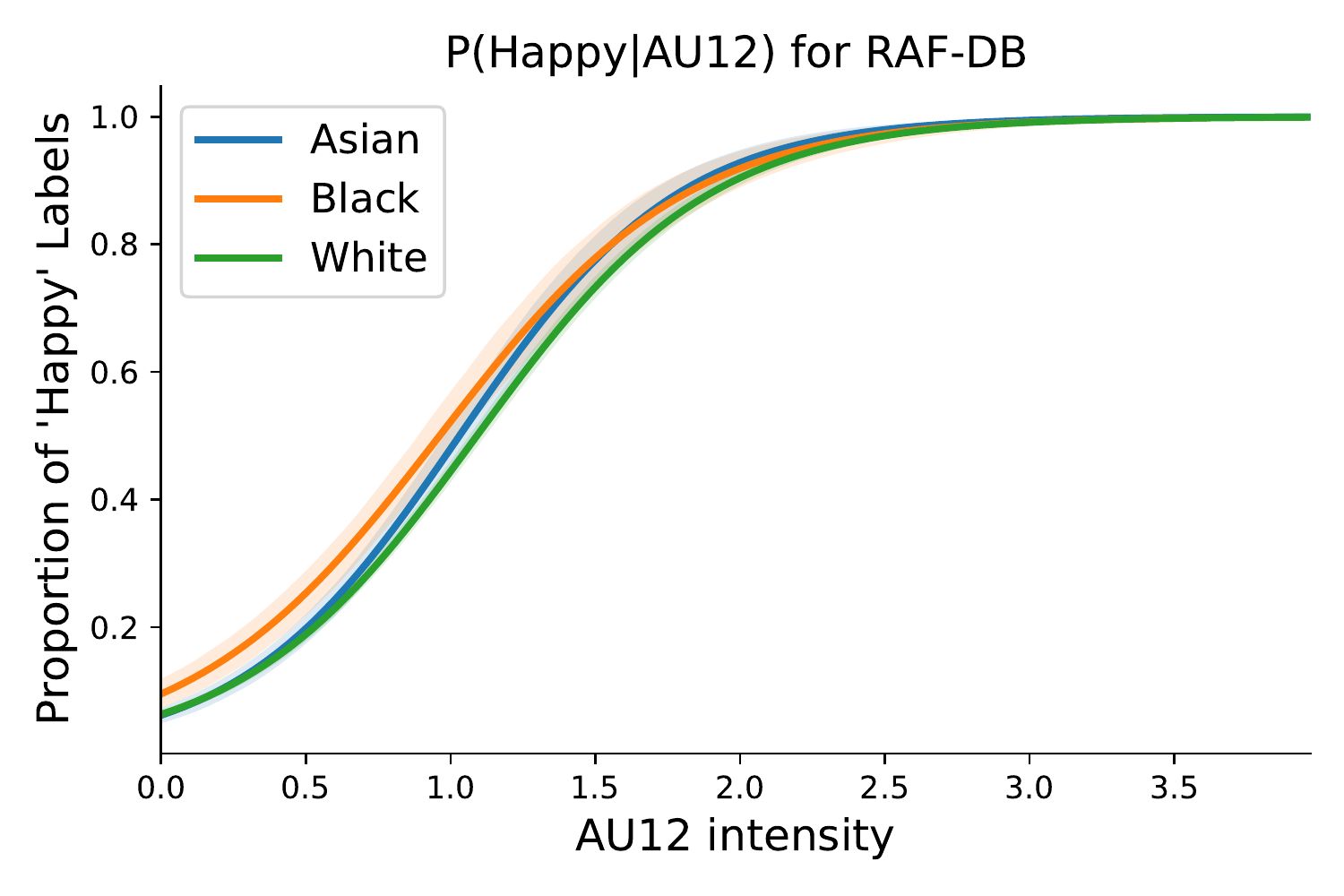}
\includegraphics[width=0.24\textwidth]{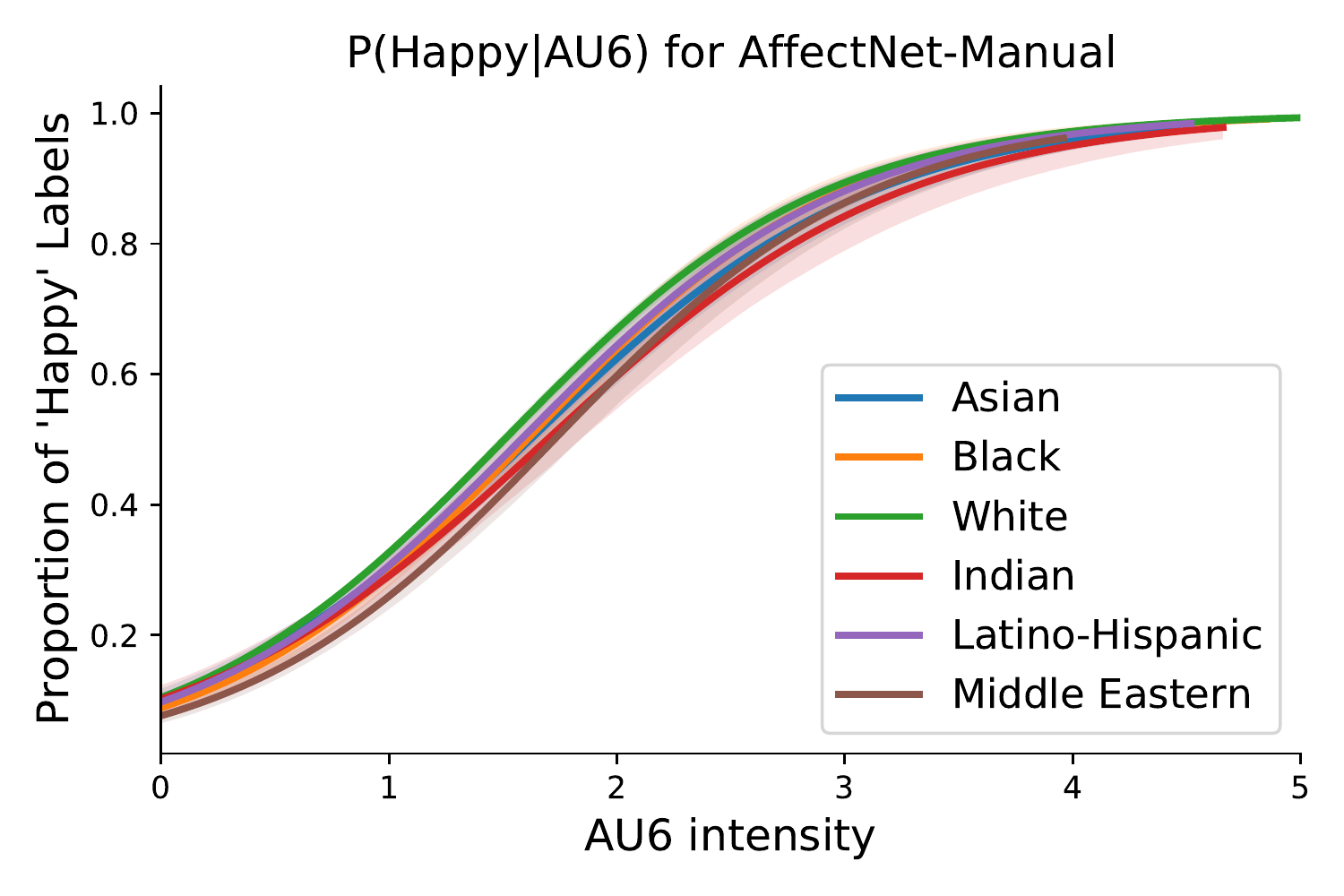}
\includegraphics[width=0.24\textwidth]{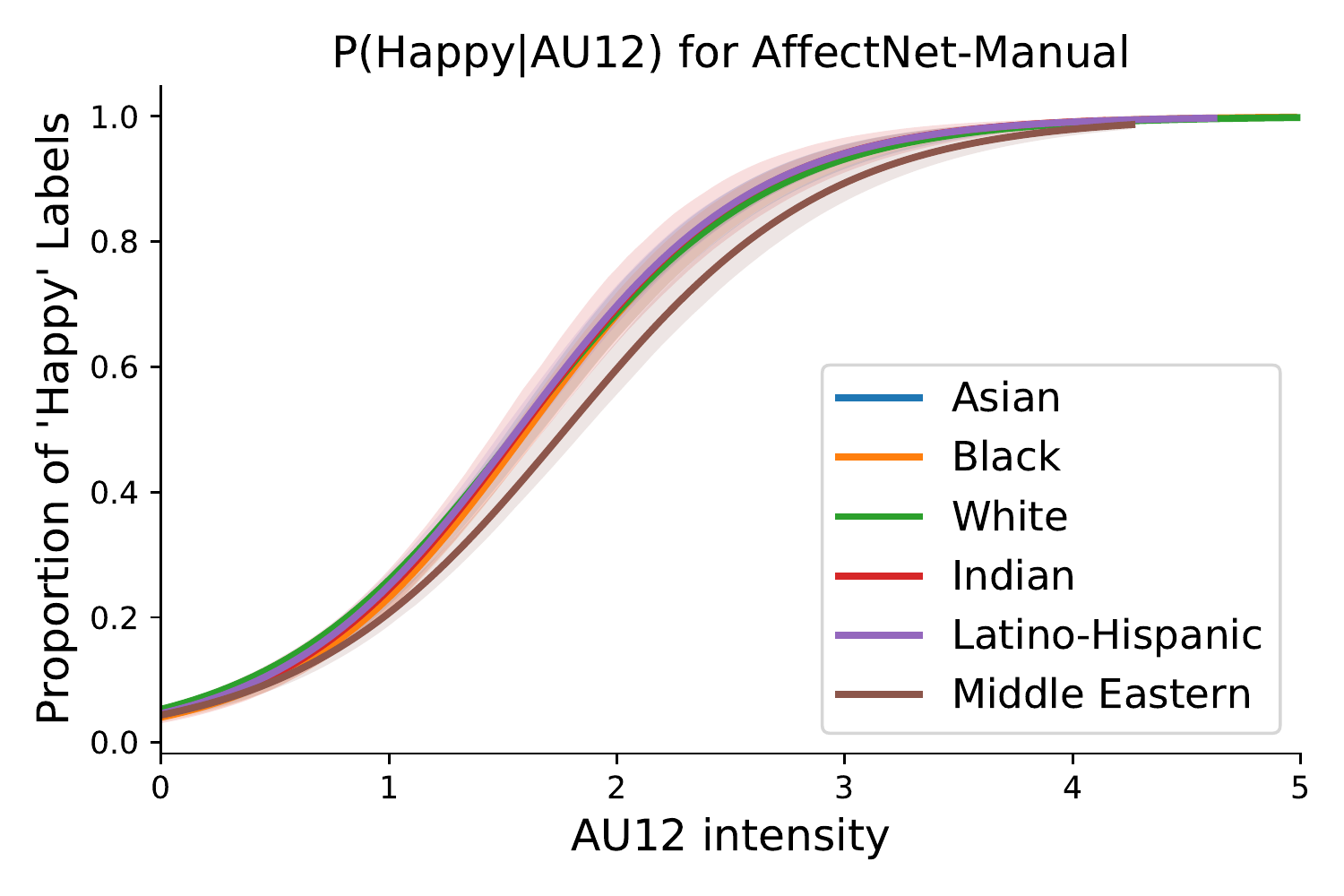}
\includegraphics[width=0.24\textwidth]{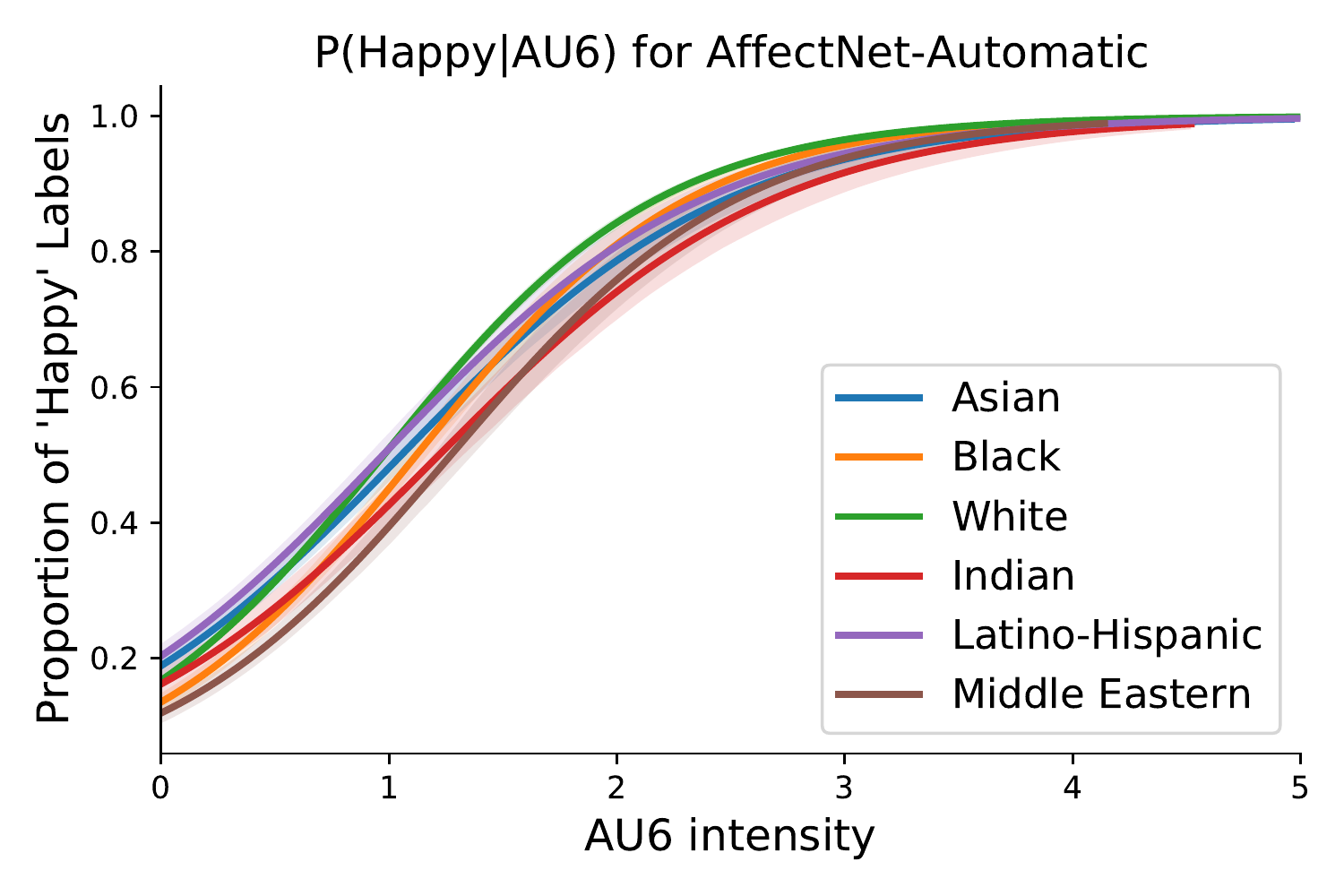}
\includegraphics[width=0.24\textwidth]{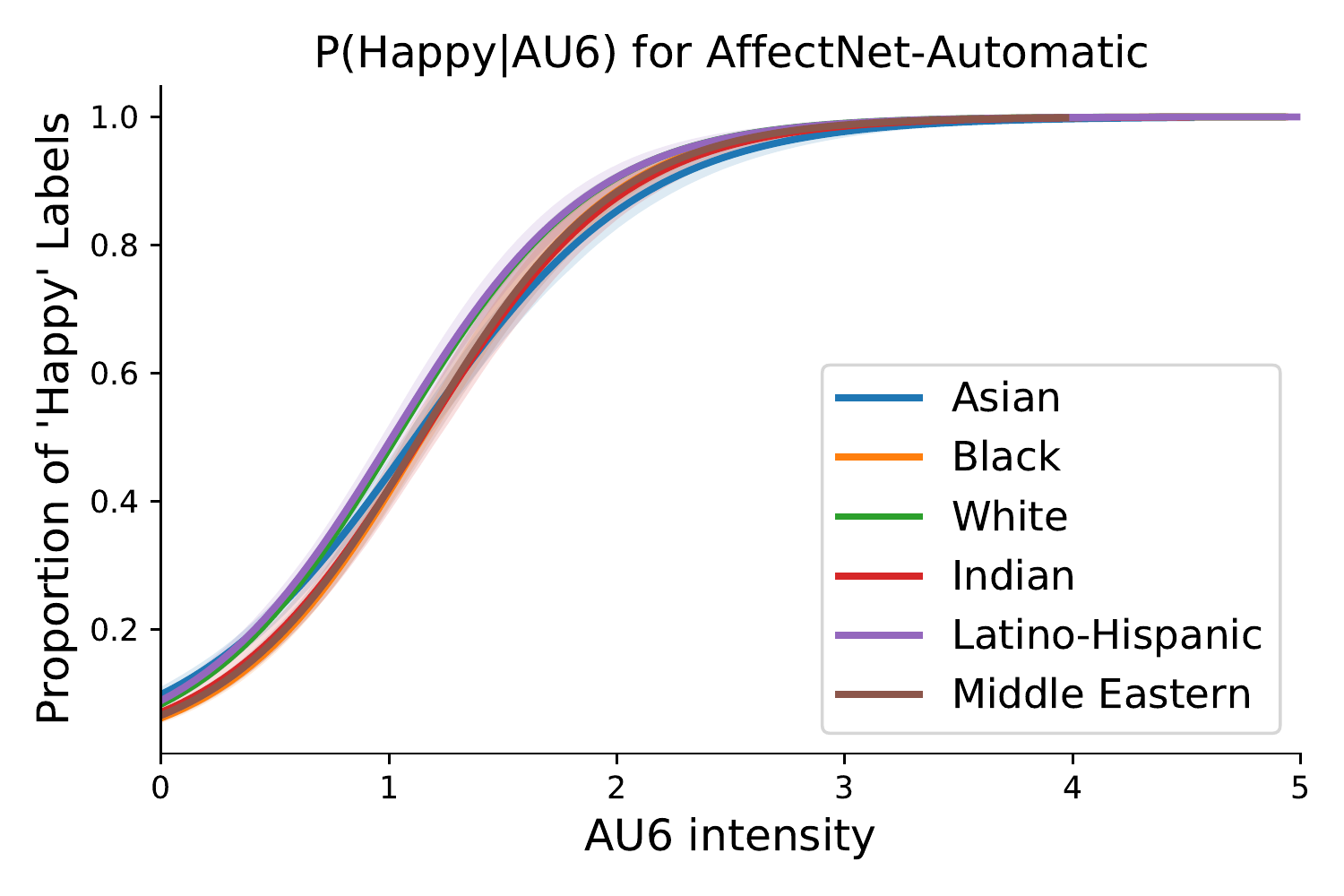}
\caption{Annotation bias of the ``happy'' expression across racial/ethnic groups for each in-the-wild expression dataset. The first row shows the proportions of ``happy'' labels for each of the three major racial groups only while the second row plots all six groups. The error bars indicate one standard error of the proportion. The last two rows show the fitted logistic regression curves as a function of AU intensities. 95\% confidence intervals are indicated by shaded regions.}
\label{fig:happy_annotation_bias_race}
\end{figure*}

Table \ref{tab:Anno_bias_happy_race_summary} shows the conditional and marginal distributions of ``happy'' labels along with the numbers of samples for each racial/ethnic group for each in-the-wild expression dataset. As mentioned in the previous section, the sum of the numbers for each racial group does not add up to those of the full datasets due to the fact that OpenFace fails to produce AU labels (possibly due to occlusion or blur) for a small fraction of the images. The p-values are the $\chi^2$ tests for independence of the "happy" labels and the racial/ethnic groups. When the p-values for the 3 racial groups are significant at the 0.05 level, the racial groups with the highest proportion of ``happy'' labels are highlighted. We can see that even though the differences in the proportion of ``happy'' labels are sometimes statistically significant, the bias is mostly idiosyncratic. In other words, there is no consistent pattern of systematic annotation bias that one group is more or less likely to be annotated ``happy'' than others.

Figure \ref{fig:happy_annotation_bias_race} plots the annotation bias of the ``happy'' expression across the racial and ethnic groups. The first row shows the proportions of ``happy'' labels for each of the three major racial groups only, while the second row plots all six groups. The last two rows show the fitted logistic regression curves as a function of AU intensities. Consistent with the results in Table \ref{tab:Anno_bias_happy_race_summary}, we see that the differences among the racial and ethnic groups are minor, and no consistent bias exists across all datasets. Further analysis on other expressions would ideally require more balanced datasets (\ie, datasets that have more minority races).
\end{revision}

{\small
\bibliographystyle{ieee_fullname}
\bibliography{bibliography}
}